\documentclass[runningheads]{llncs}
\usepackage[letterpaper,top=2cm,bottom=2cm,left=3cm,right=3cm,marginparwidth=1.75cm]{geometry}

\usepackage[T1]{fontenc}
\usepackage{graphicx}
\usepackage[nottoc]{tocbibind}
\usepackage[numbers,sort&compress,sectionbib]{natbib}
\makeatletter
\renewcommand\@biblabel[1]{#1.}
\makeatother

\usepackage{algorithm}
\usepackage{algorithmic}

\usepackage{amsmath}
\usepackage{amssymb}
\usepackage{mathtools}
\DeclareMathOperator*{\argmax}{arg\,max}

\usepackage{pgfplots}
\usepackage{subcaption}
\usepackage{tikz}
\usepackage{multirow}
\usepackage{adjustbox}
\usepackage{array}
\usepackage{booktabs}
\usepackage{nicematrix}

\begin{document}
\title{Exact Verification of Graph Neural Networks with Incremental Constraint Solving}
%
%
\author{Minghao Liu
\and
Chia-Hsuan Lu
\and
Marta Kwiatkowska
}
\authorrunning{M. Liu et al.}
%
\institute{University of Oxford, United Kingdom\\
\email{\{minghao.liu,chia-hsuan.lu,marta.kwiatkowska\}@cs.ox.ac.uk}}
\maketitle              
\begin{abstract}
Graph neural networks (GNNs) are increasingly often 
employed in high-stakes applications, such as fraud detection or healthcare, but are susceptible to adversarial attacks.
A number of techniques have been proposed to provide adversarial robustness guarantees, but support for commonly used aggregation functions in message-passing GNNs is lacking.
In this paper, we develop an exact (sound and complete) verification method for GNNs to compute guarantees against attribute and structural perturbations that involve edge addition or deletion, subject to budget constraints.
Our method employs constraint solving with bound tightening, and iteratively solves a sequence of relaxed constraint satisfaction problems while relying on incremental solving capabilities of solvers to improve efficiency.
We implement \textsc{GNNev}, a versatile exact verifier for message-passing neural networks, which supports three aggregation functions -- sum, max and mean -- with the latter two considered here for the first time.
Extensive experimental evaluation of \textsc{GNNev} on real-world fraud datasets (Amazon and Yelp) and biochemical datasets (MUTAG and ENZYMES) demonstrates its usability and effectiveness, as well as superior performance on node classification and competitiveness on graph classification compared to existing {exact verification} tools on sum-aggregated GNNs.

\keywords{Formal verification  \and Constraint solving \and Graph neural networks \and Adversarial robustness}
\end{abstract}
\section{Introduction}
\label{sec:intro}

Graph neural networks (GNNs) have been widely deployed in real-world applications, such as financial fraud detection \cite{WangQL0JWFYZY19,MotieR24}, autonomous driving \cite{CasasGLU20,CaiWS021}, healthcare treatment \cite{li2022grlhealth,GaoYWWDL24}, and scientific discovery \cite{jha2022proteingnn,reiser2022gnnscience}.
However, like all other neural network architectures, GNNs 
are vulnerable to adversarial attacks, 
where slight perturbations can cause a prediction change
\cite{ZugnerAG18,DaiLTHWZS18,TaoCSHWC21}. Therefore, it is desirable to ensure adversarial robustness of GNNs for high-stakes applications.

Formal verification is a rigorous methodology to mathematically prove that a system meets its specifications under all possible circumstances \cite{BarrettHS26}.
To evaluate the reliability and trustworthiness of neural networks, a number of approaches have been developed to verify the adversarial robustness of
fully connected neural networks \cite{KatzBDJK17,WengZCSHDBD18,WangPWYJ18,ZhangWCHD18,BotoevaKKLM20},
convolutional neural networks (CNNs) \cite{HuangKWW17,BoopathyWC0D19,TranBXJ20},
recurrent neural networks (RNNs) \cite{KoLWDWL19,AkintundeKLP19,DuJSZLSFYB021},
and transformers \cite{ShiZCHH20,BonaertDBV21}. 
Technically, these approaches can be classified into two categories.
The first category is \emph{exact} verification, also known as \emph{complete} verification, where the verification task is formulated as a constraint satisfaction problem (CSP) \cite{rossi2006handbook}
to deterministically prove whether the model is robust or not. The second category is \emph{approximate} verification, also known as {(sound but)} \emph{incomplete} verification, where the non-convex constraints are typically relaxed to convex ones, which admits efficient computation of a lower bound on the robust region.
However, when the lower bound cannot be accurately computed, the outcome of the verification may be inconclusive, which does not offer strong robustness guarantees.

Recently, the concept of adversarial robustness has been extended to GNNs. Graph inputs present additional challenges, in that the perturbations can pertain to both node attributes \cite{ZugnerG19,AnZZLSHYLQ24} and graph structure, i.e., the addition and deletion of edges \cite{BojchevskiG19,ZugnerG20,JinSPZ20,LadnerEA25}, or even node injection \cite{LaiZPZ24}. 
Since graph inputs are encoded using a combination of real-valued and discrete data, typical approaches to provide \emph{deterministic} robustness guarantees are based on Mixed-Integer Programming (MIP) \cite{ZhangCFWSMTM23,HojnyZCM24}. 
We also mention methods based on randomised smoothing \cite{BojchevskiKG20,OsselinK023,ScholtenSGBG22}, noting that their guarantees are \emph{probabilistic} and thus not directly comparable.

Despite recent advances, 
existing exact verification approaches are still limited to GNNs with \emph{sum} aggregation.
However, other common aggregations, including \emph{max} and \emph{mean}, have been shown to be necessary and effective in both theory \cite{CorsoCBLV20,RosenbluthTG23} and practice \cite{YingHCEHL18,DehmamyBY19}, and are supported as standard by the GraphSAGE library \cite{HamiltonYL17}. 
Existing methods also typically focus on graph classification problems in the context of structural perturbations,  
implemented only for edge deletion for tractability reasons. 
Node classification tasks have been under-explored, and yet are relied upon in high-stakes applications, for example, fraud detection in financial networks \cite{WangQL0JWFYZY19} and cyber attack localization in smart grids \cite{HaghshenasHN23}.
For such applications, in order to provide certified robustness guarantees, methods that yield conclusive outcomes are preferable.

We develop an exact verification method for GNNs to provide guarantees against both attribute and structural perturbations subject to local and global budget constraints.
We support GNNs for both node- and graph-classification tasks with three commonly used aggregation functions, {sum}, {max} and {mean}, where the latter two are non-linear.
We employ constraint solving {with bound tightening} and iteratively solve a sequence of relaxed CSPs while relying on incremental solving capabilities of solvers to improve efficiency. 
We implement \textsc{GNNev}, a versatile and efficient exact verifier for GNNs, and
conduct extensive experiments on two standard node-classification datasets,
Cora and CiteSeer, two real-world fraud datasets, Amazon and Yelp, {and two biochemical graph-classification datasets, MUTAG and ENZYMES.}
The results indicate that \textsc{GNNev} outperforms \textsc{SCIP-MPNN} \cite{HojnyZCM24}, the only MIP-based exact verifier known to us, on {sum} aggregation and edge deletion under different model sizes and perturbation budgets.
On a range of aggregation functions, we show that \textsc{GNNev} can provide useful insight into the susceptibility of
GNNs to adversarial attacks, which is important for deployment in high-stakes applications.

Our novel contributions can be summarised as follows.
\begin{itemize}
    \item We introduce the first exact verification method for GNNs with two common aggregation functions, {max} and {mean}, and design specialised tightened bound propagation strategies to reduce computational cost.
    \item We propose an algorithm that iteratively encodes GNN layers backwards and utilises incremental constraint solving to speed up performance while maintaining exact verification capability.
    \item We implement \textsc{GNNev}, a versatile and efficient exact verifier for GNNs that supports new aggregation functions and edge addition, and show its 
    usability on both structural and attribute perturbations 
    through extensive experiments on real-world datasets.
\end{itemize}
\section{Related Work}
\label{sec:related}

\subsubsection{Adversarial attacks against GNNs.}
Several works \cite{ZugnerAG18,DaiLTHWZS18,ZugnerG19attack,XuC0CWHL19} 
demonstrate that adversarial attacks can induce incorrect predictions by GNNs on node classification.
\citet{ChangRXHZC0H20,MuWL0XL21} propose black-box attack approaches without accessing any knowledge of the GNN classifiers.
\citet{ZouZDGKLT21,LiXCXHZ23} further improve the efficiency and scalability of adversarial attack algorithms.
Another direction is attacks on graph classification \cite{ZhangJWG21} and link prediction \cite{ZhouMWRV19,ChenZCDX23}.
We emphasise that a failure to find an attack does not imply robustness.
In contrast, we aim to verify that GNNs are not vulnerable to any admissible attack, which provides strong guarantees and can thus be used for robustness certification.

We further remark that, while most studies focus on undirected graphs and claim potential extensions to directed graphs, some works \cite{geisler2021robustness,ChenZCDX23}, along with our paper, consider both directed and undirected graphs.

\subsubsection{Robustness enhancement of GNNs.}
Adversarial training is a common technique applied to enhance the robustness of GNNs \cite{DaiSZLW19,XuC0CWHL19,FengHTC21,LiPCZLL23,GoschG0CZG23}.
\citet{Wu0TDLZ19,ZhangZ20} aim to enhance the robustness of GNNs by estimating potential adversarial perturbations and adjusting model weights to defend against these perturbations.
Moreover, generalised randomised smoothing \cite{BojchevskiKG20,WangJCG21,ZhangJWG21} and partition-ensemble \cite{XiaYWJ24} techniques have been developed for GNNs to provide defences against attacks with probabilistic guarantees.
Instead, we develop a formal verification algorithm, which is complementary to robustness enhancement and offers deterministic guarantees.

\subsubsection{Theoretical analysis of GNN verification.}
Expressiveness of sum-aggregated GNNs is studied in
\cite{Barcelo20, BenediktLMT24, NunnSST24} and mean-aggregated in \cite{Schonherr25}.
The adversarial robustness problem for
node classification is shown decidable in \citet{SalzerL23} under the assumption of bounded degree of input graphs.
\citet{BenediktLMT24} (and \citet{NunnSST24}) study verification of certain properties that quantify over all graph inputs 
proving PSPACE-completeness for GNNs with rational (resp. integer) coefficients, Boolean-attributed input graphs and truncated ReLU activation functions. 
We focus instead on a practical verification approach applicable to commonly used GNNs, with real-valued coefficients and attributes and ReLU activation functions. Our problem setting is clearly decidable because it only considers a finite and bounded input graph set.

\subsubsection{Verification of neural networks.}
Adversarial robustness verification methods can be classified into complete (exact) methods, such as constraint solving \cite{KatzBDJK17,HuangKWW17,TjengXT19}, sound though incomplete methods, e.g., convex relaxation \cite{SalmanY0HZ19,XuS0WCHKLH20}, which can be strengthened to completeness by employing branch-and-bound procedures \cite{BunelTTKM18,LuK20},
and probabilistic methods \cite{CohenRK19,WengCNSBOD19,MarzariCF25}. 
\citet{KatzBDJK17} prove that exact NN verification is generally NP-hard.
More recent works move beyond fully-connected and convolutional architectures, and include MIP-based exact verification of recurrent networks \cite{AkintundeKLP19}
as well as approximate verification of transformers utilising the zonotope abstract domain \cite{BonaertDBV21} and linear bound propagation \cite{huang2026parameterized}.
{While some works focus on properties such as differential verification between models \cite{MohammadinejadP21} and batched verification across inputs \cite{BanerjeeXS24}, we aim to verify the key property of adversarial robustness.}
Regarding GNN verification, existing approaches concern graph convolutional networks with respect to attribute \cite{ZugnerG19,AnZZLSHYLQ24} or structural \cite{BojchevskiG19,ZugnerG20,JinSPZ20,LadnerEA25} perturbations. Most are approximate, relying on optimisation \cite{BojchevskiG19,ZugnerG20,JinSPZ20,LadnerEA25} or zonotope-based relaxation \cite{LadnerEA25}.
Our approach is most similar to the exact MIP-based approach \textsc{SCIP-MPNN} \cite{HojnyZCM24} for {sum}-aggregated message-passing GNNs and edge deletion only. Their method supports static and dynamic (aggressive) bound tightening over the complete network encoding. In contrast, our method also supports {max} and {mean}, as well as edge addition, and proceeds by building the encoding incrementally layer by layer while statically tightening the bounds. For node classification tasks involving large numbers of neighbouring nodes, our exact method can greatly reduce the cost of encoding, thus enhancing performance.
We also note a recent verifier \textsc{RobLight} \cite{lu2025roblight}, which supports structural perturbations only and references our work. Their method relies on a branch-and-bound approach with bound propagation, and is unable to generate certificates for verification that MIP-based methods permit \cite{CheungGS17}.

\section{Preliminaries}
\label{sec:prelim}

\subsection{Graph Neural Networks}
Let $G = \langle V,E,X \rangle$ be an \emph{attributed directed graph}, where $V = \{ v_1, v_2, \dots, v_n\}$ is a set of nodes, $E \subseteq V \times V$ is a set of edges with no self-loops,
and $X = \{\mathbf{x}_1, \mathbf{x}_2, \dots, \mathbf{x}_n\}$ is a set of real
\emph{attribute vectors} for the nodes with the same dimension. 
For a node $v \in V$ and a natural number $k$,
we denote the set of
\emph{$k$-hop incoming neighbours of $v$} by $\mathcal{N}_{k}(v)$.
For simplicity, we write $\mathcal{N}(v)$ when $k=1$.
\emph{Graph neural networks (GNNs)} are a class of neural network architectures designed to operate on graph data.
In this paper, we consider node- and graph-classification tasks. Given a class set $C = \{ c_1, c_2, \dots, c_m \}$, for node classification, the goal is to classify a node $v \in V$ into a class within $C$; for graph classification, the goal is to classify a graph $G$ into a class within $C$.
More precisely, we view a GNN as a function $f$ such that,
given an attributed directed graph $G = \langle V,E,X \rangle$ and a node $v \in V$,
$f(G, v) \in C$ for node classification, and $f(G) \in C$ for graph classification.
The remainder of the formalisation focuses on the node classification case for simplicity. The variant for graph classification is similar, as discussed in Appendix~\ref{appx:defgraphclass}.

There are a variety of GNN architectures.
Existing formal verification works mostly focus on 
graph convolutional networks (GCNs) \cite{KipfW17},
which rely on a specific neighbourhood aggregation scheme and their expressiveness is restricted \cite{xu2019gin}.
In this work, we consider \emph{GraphSAGE} \cite{HamiltonYL17}, also known as \emph{Message-Passing Neural Networks (MPNNs)} \cite{GilmerSRVD17},
which are a more general GNN architecture because (i) they support the choice of multiple neighbourhood aggregation functions and (ii) the expressive power of MPNNs has been shown to encompass other popular variations such as GCNs and graph attention networks (GATs) \cite{BronsteinBCV21}.

A GNN consists of $K$ layers.
The dimension of a GNN is a $(K+1)$ tuple of positive integers $d_0, d_1, \dots, d_K$.
For an input attributed directed graph $G = \langle V, E, X \rangle$,
for each node $v \in V$, its real-valued
embedding vector $\mathbf{h}_v^{(0)}$ is a $d_0$-dimension vector initialised to its corresponding attribute vector $\mathbf{x}_v$.
For the $k$-th layer, GNNs compute the $d_k$-dimensional embedding $\mathbf{h}_v^{(k)}$ by
\begin{equation}
    \label{eq:gnn}
    \hspace{-3pt}
    \mathbf{h}_v^{(k)} = \sigma \left( 
    \mathbf{W}_1^{(k)} \cdot \mathbf{h}_v^{(k-1)} +
    \mathbf{W}_2^{(k)} \cdot \mathbf{aggr}\left( \left\{\!\!\!\left\{ \mathbf{h}_u^{(k-1)} \mid u \in \mathcal{N}(v) \right\}\!\!\!\right\} \right) + \mathbf{b}^{(k)}_1
    \right),
\end{equation}
where $\mathbf{W}_1^{(k)}, \mathbf{W}_2^{(k)}$ are learnable parameter matrices with dimension $d_k \times d_{k-1}$ for the $k$-th layer, {$\mathbf{b}_1^{(k)}$ is a learnable parameter vector}, $\sigma(x)=\max(0,x)$ is the ReLU activation function, $\mathbf{aggr} \in \left\{ \text{sum}, \text{max}, \text{mean} \right\}$ is the aggregation function, and $\left\{\!\!\left\{\right\}\!\!\right\}$ denotes a multiset.
Finally,
the last dimension $d_K$ is exactly $m$,
and the predicted class $\hat{c}_v$ of a node $v \in V$ is obtained by the Softmax function:
\begin{equation}
    \hat{c}_v = \argmax_{c \in C}{\,\left(\text{Softmax}\left( {\mathbf{h}_v^{(K)}} \right) \left[c\right]\right)}.
\end{equation}

\subsection{Adversarial Robustness of GNNs}
Node embeddings are directly affected by attribute perturbations and indirectly affected by structural perturbations through message passing, which may result in prediction instability.
Changes in predictions caused by admissible perturbations indicate a lack of \emph{adversarial robustness}.
We slightly adapt the definition of adversarial robustness for GNNs
in \cite{BojchevskiG19} by (i) allowing perturbations in both graph structure and node attributes, and (ii) removing the redundant fixed edge set.
Given an attributed directed graph $G = \langle V,E,X \rangle$, the attacker typically has limited capabilities to perturb it.
Let $F \subseteq V \times V$ be a set of \emph{fragile edges}.
Intuitively, an attacker can either remove edges in $F \cap E$ or insert edges in $F \backslash E$.
For structural perturbations, we assume a global budget $\Delta$ and a local budget $\delta_v$ for each $v \in V$, both of which are non-negative integers. 
For attribute perturbations, we work with {budgets $\epsilon_{v,i}^l,\epsilon_{v,i}^u$ ($\epsilon_{v,i}^l \le \mathbf{x}_v[i] \le \epsilon_{v,i}^u$)} for each $v \in V$ and $i \in \left\{1, 2, \dots, d_0\right\}$, which are real values representing the range of perturbation allowed for the $i$-th dimension of $\mathbf{x}_v \in X$.
The admissible perturbation space is defined as follows.
\begin{definition}[Admissible perturbation space of graph]
\label{def:perturbspace}
Given an attributed directed graph $G = \langle V, E, X \rangle$, 
a set of fragile edges $F \subseteq V \times V$,
and perturbation budgets $\Delta, \delta, \epsilon$,
the \emph{admissible perturbation space} $\mathcal{Q}(G)$ of $G$ with respect to $F$ and budgets is the set of attributed directed graphs $\tilde{G} = \langle V, \tilde{E}, \tilde{X} \rangle$ that satisfy
\begin{enumerate}v 
    \item $E \backslash F \subseteq \tilde{E} \subseteq E \cup F$, 
    \item $|E \backslash \tilde{E}| + |\tilde{E} \backslash E| \le \Delta$,
    \item For every $v \in V$,
        $|\mathcal{N}(v) \backslash \tilde{\mathcal{N}}(v)| + |\tilde{\mathcal{N}}(v) \backslash \mathcal{N}(v)| \le \delta_v$,
    \item For every $v \in V$ and $1 \le i \le d_0$,
    {$\epsilon_{v,i}^l \le \tilde{\mathbf{x}}_v[i] \le \epsilon_{v,i}^u$},
\end{enumerate}
\end{definition}
\noindent where we denote the incoming neighbours of $v$ in the graph $\tilde{G}$ by $\tilde{\mathcal{N}}(v)$.
Note that this is a general formulation as $F$ can be defined arbitrarily for different purposes, and allows us to consider edge addition when the size of $F$ is moderate\footnote{The definition in \citet{HojnyZCM24}, Eq.~(14) is a special case of ours when $F=E$.}.
For example, if $F=E$, only edge deletion is allowed, and 
any existing edge can be deleted. 
If $F=V \times V$, edges can be added or deleted between any pair of nodes.

Finally, we introduce the formal definition of adversarial robustness of GNNs.
\begin{definition}[Adversarial robustness of GNNs]
\label{def:robustnode}
Given a node-classifica\-tion GNN $f$, an attributed directed graph $G$ with admissible perturbation space $\mathcal{Q}(G)$,
a target node $t \in V$, and its predicted class $\hat{c}_t \in C$,
we say that $f$ is \emph{adversarially robust} for $t$ with class $\hat{c}_t$ if and only if, 
for every perturbed graph $\tilde{G} \in \mathcal{Q}(G)$,
it holds that $f(\tilde{G}, t) = \hat{c}_t$.
\end{definition}
Adversarial robustness verification aims to guarantee that the prediction for a given node will not change under any admissible perturbations, subject to budget constraints, and can be reduced to computing the worst-case margin between the target class and other classes.  

\section{Exact Verification of GNNs}
\label{sec:method}

In this section, we introduce our method, which encodes the verification task 
as a constraint satisfaction problem (CSP).
Next, the bound tightening strategies for {max-} and {mean}-aggregated GNNs are proposed.
Finally, we illustrate an efficient exact verification algorithm based on incremental solving.

\subsection{Encoding}
\label{sec:encoding}
A CSP aims to determine whether there exists an assignment of variables that satisfies a given set of constraints.
We encode the exact verification tasks as CSPs, which consist of three parts: input perturbation, GNN architecture, and verification objective.
Following existing MIP encodings for GNNs \cite{ZhangCFWSMTM23,HojnyZCM24}, we extend them to accommodate {max} and {mean} aggregations,
{where for mean we use big-M encoding and for max we rely on the solver.}

\subsubsection{Input perturbation.}
Consider an input {attributed directed} graph $G = \langle V,E,\\X \rangle$, where our goal is to verify the adversarial robustness for the target node $t \in V$.
Note that, for $K$-layer GNNs, only perturbations to nodes in $\mathcal{N}_{K}(t)$ can affect the prediction of $t$.
Two kinds of perturbations are allowed.
The first is attribute perturbation, where the node attributes can be modified within a specified range. For each node $v \in \mathcal{N}_K(t)$,
we set $d_0$ real variables $attr_{v,1}, attr_{v,2}, \dots, attr_{v,d_0}$ as the attribute values after perturbations.
Given the perturbation budgets {$\epsilon_{v,i}^l,\epsilon_{v,i}^u$} for the $i$-th dimension of $\mathbf{x}_v$, the following constraint is applied:
\begin{equation}
    \label{eq:attr}
    {\epsilon_{v,i}^l\ \le\ attr_{v,i}\ \le\ \epsilon_{v,i}^u.}
\end{equation}
The second type is structural perturbation, which can impact the message-passing mechanism. For each edge $(u,v) \in F$, we set a Boolean variable $pe_{u,v}$ to represent whether $(u,v)$ is perturbed.
Given the global budget $\Delta$ and the local budget $\delta_v$ for each $v \in \mathcal{N}_{K-1}(t)$, we have the constraints
\begin{equation}
    \sum_{(u,v) \in F}{pe_{u,v}} \le \Delta,\ \ \ \ \ \ \ \ 
    \sum_{(u,v) \in F}{pe_{u,v}} \le \delta_v.
\end{equation}

\subsubsection{GNN architecture.}
We encode the architecture of GNNs as follows.
For the $k$-th layer as shown in Eq.~(\ref{eq:gnn}),
the output embedding of node $v \in \mathcal{N}_{K-k}(t)$ is represented by $d_k$
real variables $h^{(k)}_{v,1}, h^{(k)}_{v,2}, \dots, h^{(k)}_{v,d_k}$.
Let $h^{(0)}_{v,i}=attr_{v,i}$.
The encoding for each layer consists of three parts.

First, the neighbours of node $v$ are aggregated to produce the message. We set $d_{k-1}$ real variables $msg^{(k)}_{v,1}, msg^{(k)}_{v,2}, \dots, msg^{(k)}_{v,d_{k-1}}$ as the message vector.
We support three common aggregation functions: {sum}, {max}, and {mean}. Note that the last two functions are non-linear, which accounts for higher computational complexity for verification.
Before we give the encoding for the aggregation functions, we set the auxiliary variables $a^{(k)}_{v,i,u}$, which indicate the contribution of node $u$ to $msg^{(k)}_{v,i}$ through the edge $(u,v)$.
The following constraints restrict the values of $a^{(k)}_{v,i,u}$:
for every $(u, v) \in E \backslash F$,
\begin{equation}
    a^{(k)}_{v,i,u} = h^{(k-1)}_{v,i};
\end{equation}
for every $(u, v) \in E \cap F$,
\begin{equation}
    pe_{u,v} \rightarrow a^{(k)}_{v,i,u} = 0
    \ \ \ \text{and} \ \ \
    \neg pe_{u,v} \rightarrow a^{(k)}_{v,i,u} = h^{(k-1)}_{v,i};
\end{equation}
for every $(u, v) \in F \backslash E$,
\begin{equation}
    pe_{u,v} \rightarrow a^{(k)}_{v,i,u} = h^{(k-1)}_{v,i}
    \ \ \ \text{and} \ \ \ 
    \neg pe_{u,v} \rightarrow a^{(k)}_{v,i,u} = 0.
\end{equation}
For {sum} and {max} aggregation, the constraints are shown in Eq.~(\ref{eq:sum_aggr}) and (\ref{eq:max_aggr}), respectively.
\begin{equation}
    \label{eq:sum_aggr}
    msg^{(k)}_{v,i} = \sum_{(u,v) \in F \cup E}{a^{(k)}_{v,i,u}},
\end{equation}
\begin{equation}
    \label{eq:max_aggr}
    msg^{(k)}_{v,i} = \max_{(u,v) \in F \cup E}{a^{(k)}_{v,i,u}}.
\end{equation}
For {mean} aggregation, the constraints are set as follows:
\begin{equation}
    \label{eq:mean_aggr}
    \begin{split}
        & deg_v = \sum_{(u,v) \in F \cap E}{(1-pe_{u,v})} + \sum_{(u,v) \in F \backslash E}{pe_{u,v}} + |E \backslash F|, \\
        & deg_v \cdot msg^{(k)}_{v,i} = \sum_{(u,v) \in F \cup E}{a^{(k)}_{v,i,u}},\\
        & -M \cdot deg_v \le msg^{(k)}_{v,i} \le M \cdot deg_v,
    \end{split}
\end{equation}
where $M$ is a big real number\footnote{$M$ is calculated by finding the maximum of the absolute values of the upper and lower bounds of all $a^{(k)}_{v,i,u}$ in Eq.~(\ref{eq:mean_aggr}).}.
Next, we introduce a constraint that represents the pre-ReLU embedding of node $v$ as follows:
\begin{equation}
    \label{eq:lin_trans}
    {y}^{(k)}_{v,i} = \sum_{j \in [1,d_{k-1}]}{ \mathbf{W}^{(k)}_1[i,j] \cdot h^{(k-1)}_{v,j} + \mathbf{W}^{(k)}_2[i,j] \cdot msg^{(k)}_{v,j} + \mathbf{b}^{(k)}_1}.
\end{equation}
Finally, we define the post-ReLU embedding, except the last GNN layer, as follows:
\begin{equation}
    \label{eq:relu}
    {h}^{(k)}_{v,i} = \max\left( {y}^{(k)}_{v,i}, 0 \right).
\end{equation}

\subsubsection{Verification objective.}
According to Definition~\ref{def:robustnode}, our objective is to verify that no admissible attribute or structural perturbation can make the prediction of the GNN inconsistent with the input {attributed directed} graph for a target node. 
Let the predicted class of target node $t$ be $\hat{c}_t$, where the set of classes is denoted by $C$. Then we set the following constraint:
\begin{equation}
    \label{eq:obj}
    \left(\max_{c\in C\backslash\{\hat{c}_t\}}{{y}^{(K)}_{t,c}}\right)
    \geq {y}^{(K)}_{t,\hat{c}_t},
\end{equation}
which expresses the worst-case margin in the input of Softmax.

Let $\mathbf{y}^{(K)}_t = \left[y^{(K)}_{t, 1}, y^{(K)}_{t, 2}, \dots, y^{(K)}_{t, |C|}\right]$.
Since the Softmax function is monotonically increasing,
Eq.~(\ref{eq:obj}) implies that there exists $c \in C$ with $c \neq \hat{c}_t$ such that
\begin{equation*}
    \text{Softmax}\left(\mathbf{y}^{(K)}_{t}\right)[c]
    \ge
    \text{Softmax}\left(\mathbf{y}^{(K)}_{t}\right)[\hat{c}_t].
\end{equation*}
If Eq.~(\ref{eq:obj}) is feasible, this means that a perturbed graph has been found that makes the GNN misclassify node $t$.
Therefore, the robustness verification task is translated to deciding whether the above CSP is unsatisfiable.

Note that the encoding involves at most $5N^2D$ real variables, $N^2$ Boolean variables, and $8ND + 2$ constraints,
where $N$ is the number of nodes in $\mathcal{N}_K(t)$ and $D := \sum_{0 \le i \le K} d_i$.
Furthermore, the encoding can be computed in time polynomial in the size of the GNN.

\subsection{Bound Tightening for Aggregations}
\label{sec:bound-tight}
Tightening of variable bounds has been found to be critical to enhancing the efficiency of MIP solvers. Simultaneously, high-quality bounds form the basis of our incremental solving algorithm {introduced in Section~\ref{sec:inc-solve}}.
Starting from the input variables (i.e., $attr$), whose bounds have been determined by Eq.~(\ref{eq:attr}), we propagate these bounds layer by layer to obtain tight bounds for other variables.

Since tightened bound propagation is straightforward for linear transformations and ReLU functions, our main focus is on tightening the bounds for aggregation functions.
We formalise the problem as follows:
given three finite sets of variables $X_1$, $X_2$, and $X_3$,
the upper and lower bounds for each variable,
and a non-negative integer $s$,
compute the upper and lower bounds for the variable
\begin{equation*}
    z := \mathbf{aggr}(X_1 \cup X'_2 \cup X'_3),
\end{equation*}
where 
$\mathbf{aggr} \in \left\{ \text{sum}, \text{max}, \text{mean} \right\}$,
$X_2' \subseteq X_2$, $X_3' \subseteq X_3$, and $|X_2 \backslash X_2'| + |X_3'| \le s$.
Let $N$ be the total size of the sets $X_1$, $X_2$, and $X_3$.
The intuition behind the formalisation is as follows: $X_1$, $X_2$, and $X_3$ represent the embedding {vectors} of nodes connected to the target node via non-fragile edges ($E/F$), fragile edges ($E \cap F$), and fragile non-edges ($F/E$), respectively.
The attacker can either delete fragile edges, that is, remove elements from $X_2$, or insert fragile non-edges, that is, select elements from $X_3$.
The number of deletion and insertions is constrained by the budget $s$.

For a set of variables $X$ and an integer $k$,
let $hi(X, k)$ denote the $k$-th \emph{largest upper bound} among the variables in $X$.
If $k > |X|$, we define
$hi(X, k) = -\infty$.
Similarly, let $lo(X,k)$ denote the $k$-th \emph{smallest lower bound} among the variables in $X$,
with $lo(X,k)=\infty$ {when} $k \le 0$.
Note that $hi(X, k)$ and $lo(X, k)$ can be computed in $O(|X| \log k)$ time by maintaining a max- or min-heap.
For multiple values $k_1, k_2, \dots, k_\ell$,
we can reuse the heap, resulting in an overall complexity of $O(|X| \log k)$, where $k$ is the maximum value among $k_1, k_2, \dots, k_\ell$.

For a variable $x$, its upper and lower bounds are denoted by $\overline{x}$ and $\underline{x}$ respectively.
The case for $\text{sum}$ aggregation has been shown by \citet{HojnyZCM24}.
We restate their bounds in the context of our formalisation for completeness:
\begin{equation*}
    \begin{aligned}
        \overline{z} =&
        \sum_{x \in X_1 \cup X_2} \overline{x} +
        \sum_{1 \le i \le s} \max(hi(Y, i), 0), \\
        \underline{z} =&
        \sum_{x \in X_1 \cup X_2} \underline{x} +
        \sum_{1 \le i \le s} \min(lo(Y, i), 0),
    \end{aligned}
\end{equation*}
where $Y=\left\{-x\mid x \in X_2 \right\}\cup X_3$.
The upper and lower bounds of $z$ can be computed in time $O(N \log s)$.

\subsubsection{Max aggregation.}
The upper and lower bounds of $z$ for $\text{max}$ aggregation can be obtained by case analysis.
For the corner case $s = 0$, $z = \max(X_1 \cup X_2)$.
Note that, if $X_1 \cup X_2 = \emptyset$,
then $\overline{z} = \underline{z} = 0$.
We now focus on the case $s > 0$.
For the upper bound:
\begin{equation*}
    \overline{z} =
    \begin{cases}
        \max(0, hi(X_2, 1), hi(X_3, 1)), \hspace{50pt} \text{if } X_1 = \emptyset, s \ge |X_2|, \\
        \max(hi(X_1, 1), hi(X_2, 1), hi(X_3, 1)), \hspace{15pt} \text{otherwise}.
    \end{cases}
\end{equation*}
On the other hand, for the lower bound:
\begin{equation*}
    \underline{z} =
    \begin{cases}
        \min(0, lo(X_2, 1), lo(X_3, 1)),\hspace{56pt} \text{if } X_1 = \emptyset, s > |X_2|, \\
        \min(0, lo(X_2, 1)),\hspace{98.5pt} \text{if } X_1 = \emptyset, s = |X_2|, \\
        lo(X_2, |X_2| - s),\hspace{102pt} \text{if } X_1 = \emptyset, s < |X_2|, \\
        lo(X_1, |X_1|),\hspace{119pt} \text{if } X_1 \neq \emptyset, s \ge |X_2|, \\
        \max(lo(X_1, |X_1|), lo(X_2, |X_2| - s)),\hspace{20pt} \text{if } X_1 \neq \emptyset, s < |X_2|.
    \end{cases}
\end{equation*}
Note that the upper and lower bounds for $z$ can be computed in time $O(N \log s)$.

\subsubsection{Mean aggregation.}
In this case,
we reduce the original problem to a combination of subproblems.
For non-negative integers $s_2$ and $s_3$, define the variable
\begin{equation*}
    z_{s_2, s_3} := \mathbf{mean}(X_1 \cup X'_2 \cup X'_3),
\end{equation*}
where 
$X_2' \subseteq X_2$,
$X_3' \subseteq X_3$,
$|X_2 \backslash X_2'| = s_2$,
and $|X_3'| = s_3$.
The main difference is that we now explicitly fix separate budgets for deletions and insertions.
The upper and lower bounds of the original problem $z$ are given by
\begin{equation*}
    \begin{aligned}
        \overline{z}  =
        &\max_{\substack{s_2 + s_3 \le s\\ 0 \le s_2 \le |X_2|\\ 0 \le s_3 \le |X_3|}}
        \overline{z}_{s_2, s_3}
        \quad \text{and}\quad
        \underline{z} =
        &\min_{\substack{s_2 + s_3 \le s\\ 0 \le s_2 \le |X_2|\\ 0 \le s_3 \le |X_3|}}
        \underline{z}_{s_2, s_3}.        
    \end{aligned}
\end{equation*}
For the subproblem $z_{s_2, s_3}$, since the budgets for deletion and insertion are
separate,
the upper and lower bounds can be obtained through greedy choices.
In the corner case where $X_1$ is empty, $s_2 = |X_2|$, and $s_3 = 0$,
$\overline{z}_{s_2, s_3} = \underline{z}_{s_2, s_3} = 0$.
In all other cases, we have 
$\overline{z}_{s_2, s_3} = \overline{w}_{s_2, s_3} / n_{s_2, s_3}$ and
$\underline{z}_{s_2, s_3} = \underline{w}_{s_2, s_3} / n_{s_2, s_3}$,
where $n_{s_2, s_3} = |X_1| + |X_2| - s_2 + s_3$ and
\begin{equation*}
    \begin{aligned}
        \overline{w}_{s_2, s_3} =
        &\sum_{x \in X_1} \overline{x} + 
            \sum_{1 \le i \le |X_2| - s_2} hi(X_2, i) +
            \sum_{1 \le i \le s_3} hi(X_3, i),
        \\
        \underline{w}_{s_2, s_3} =
        &\sum\limits_{x \in X_1} \underline{x} + 
            \sum\limits_{1 \le i \le |X_2| - s_2} lo(X_2, i) +
            \sum\limits_{1 \le i \le s_3} lo(X_3, i).
    \end{aligned}
\end{equation*}
Note that the summations over $lo$ and $hi$ can be computed incrementally among subproblems and, as there are at most $(s+1)^2$ subproblems,
we can compute the upper and lower bounds for $z$ in time $O(s^2 + N \log s)$ using a nai\"ve approach.
In fact, we can further improve this result by exploiting the unimodality of the sequence,
reducing the computation time to $O((s + N) \log s)$.
The detailed algorithm is provided in Appendix~\ref{appx:greedy}.

\begin{theorem}
The upper and lower bounds provided in this section for max and mean aggregations are tight. That is, there exist sets $X_2'\subseteq X_2$ and $X_3'\subseteq X_3$ with $|X_2 \backslash X_2'| + |X_3'| \le s$ that achieve these bounds exactly.
\end{theorem}
\noindent
The proof is included in Appendix~\ref{appx:tight_bounds}.

\begin{example}
    Consider sets of variables
    $X_1 = \{x_{1, 1}\}$,
    $X_2 = \{x_{2, 1}, x_{2, 2}\}$, and
    $X_3 = \{x_{3, 1}, x_{3, 2}\}$.
    Each variable has upper and lower bounds as shown in \figurename~\ref{fig:tb}.
    Let the budget $s = 1$.
    We now consider tightened bound propagation:
    \begin{itemize}
        \item For $\text{max}$ aggregation,
        the upper bound is $5$ by no operation, the same as the plain method\footnote{The plain method computes the upper (lower) bounds for max aggregation as the maximum (minimum) bound of all the variables, and for mean aggregation by sorting variable bounds from $X_2 \cup X_3$ in ascending order and finding the maximum (minimum) mean value among all postfix (prefix) sequences.};
        the lower bound is $2$ by deleting $x_{2, 2}$, improved by 5 over the plain method.
        \item For $\text{mean}$ aggregation,
        the upper bound is $3.25$ by inserting $x_{3, 1}$, improved by 0.08;
        the lower bound is $0.5$ by inserting $x_{3, 1}$, improved by 2.17.
    \end{itemize}
    For both aggregations, tightened bound propagation reduces the domain of the variable $z$.
\end{example}

\begin{figure}[t]
    \centering
    
    \begin{minipage}{0.45\textwidth}
        \resizebox{\linewidth}{!}{
        \begin{tikzpicture}
            \node[circle,draw=black,fill=white,inner sep=0pt,minimum size=18pt,label=below:{}] (vt) at (0, -1.5) {$z$};
            
            \node[circle,draw=black,fill=white,inner sep=0pt,minimum size=20pt,label=above:{\scriptsize $[0, 1]$},label=left:{}] (v11) at (-3, 0) {$x_{1,1}$};
            
            \node[circle,draw=black,fill=white,inner sep=0pt,minimum size=20pt,label=above:{\scriptsize $[2, 3]$},label=left:{}] (v21) at (-1.5, 0) {$x_{2,1}$};
            \node[circle,draw=black,fill=white,inner sep=0pt,minimum size=20pt,label=above:{\scriptsize $[3, 5]$},label=left:{}] (v22) at (0, 0) {$x_{2,2}$};
            
            \node[circle,draw=black,fill=white,inner sep=0pt,minimum size=20pt,label=above:{\scriptsize $[-3, 4]$},label=right:{}] (v31) at (1.5, 0) {$x_{3,1}$};
            \node[circle,draw=black,fill=white,inner sep=0pt,minimum size=20pt,label=above:{\scriptsize $[-2, 2]$},label=right:{}] (v32) at (3, 0) {$x_{3,2}$};

            \draw[black,<-] (vt) --  node[color=black,above=-2,sloped] {} (v11);
            
            \draw[dashed,black,<-] (vt) --  node[color=black,above=-2,sloped] {} (v21);
            \draw[dashed,black,<-] (vt) --  node[color=black,above=-2,sloped] {} (v22);

            \draw[dashed,gray!60,<-] (vt) --  node[color=black,above=-2,sloped] {} (v31);
            \draw[dashed,gray!60,<-] (vt) --  node[color=black,above=-2,sloped] {} (v32);
        \end{tikzpicture}
        }
    \end{minipage}
    \hfill
    \begin{minipage}{0.51\textwidth}
        \resizebox{\linewidth}{!}{
        \renewcommand{\arraystretch}{1.2}
        \begin{tabular}{lcc}
          & Plain & Tightened \\
          \hline
          max (\emph{ub}) & $\max(1,3,5,4,2)=\mathbf{5}$ & $\max(1,5,4)=\mathbf{5}$ \\
          max (\emph{lb}) & $\min(0,2,3,-3,-2)=-3$ & $\max(0,2)=\mathbf{2}$ \\
          mean (\emph{ub}) & $\frac{1+5+4}{3}=3.33$ & $\frac{1+5+3+4}{4}=\mathbf{3.25}$ \\
          mean (\emph{lb}) & $\frac{0-3-2}{3}=-1.67$ & $\frac{0+2+3-3}{4}=\mathbf{0.5}$ 
        \end{tabular}
        }
    \end{minipage}
    
    \caption{An illustration of tightened bound propagation.
    Solid lines denote the non-fragile edges,
    black dashed lines denote fragile edges, and
    gray dashed lines denote fragile non-edges.
    The tighter bounds are shown in \textbf{bold}.}
    \label{fig:tb}
\end{figure}

\vspace{-15pt}
\subsection{Verification with Incremental Solving}
\label{sec:inc-solve}
With tightened bound propagation, we can verify adversarial robustness of GNNs more efficiently. 
However, in node classification tasks, graphs are typically large in size, and for a node $t$ we typically have that $|\mathcal{N}_{k+1}(t)| \gg |\mathcal{N}_{k}(t)|$ for the $k$-th layer, leading to rapid expansion of the encoding and a decrease in efficiency.
To improve performance while maintaining exactness of verification, 
we utilise \emph{incremental solving}, an effective mechanism in existing CSP solvers, to iteratively solve a series of simplified relaxation problems compared to the original problem.

\begin{algorithm}[t]
\caption{Exact Verification of GNNs with Incremental Solving}
\label{alg:incgnnverify}
\textbf{Input}: Trained GNN $f$ with $K$ layers, attributed {directed} graph $G=\langle V,E,X \rangle$, target node $t$ (for node classification), fragile edge set $F$, perturbation budgets $\Delta,\delta,\epsilon$. \\
\textbf{Output}: Verification result.
\begin{algorithmic}[1] 
\STATE $\mathit{Bounds} \leftarrow \{\underline{attr},\overline{attr}\}$;
\FOR{$k$ from $1$ to $K$}
    \STATE $\varphi_k \leftarrow \text{EncodeGNNLayer}(f,k,G,F,\Delta,\delta,\epsilon)$;
    \STATE $\mathit{Bounds} \leftarrow \text{BoundPropagation}(\varphi_k,\mathit{Bounds})$;
\ENDFOR
\STATE $\varphi_{obj} \leftarrow \text{EncodeObjective}(\varphi_K,t)$;  \COMMENT Also encode final layer for graph classification.
\STATE $S_{\Theta} \leftarrow \text{InitIncSolver()}$;
\STATE $\Phi \leftarrow \varphi_{obj}$;
\FOR{$k$ from $K$ to $1$}
    \STATE $\Phi \leftarrow \Phi \wedge \varphi_{k}$;
    \STATE $\left(\mathit{result},\Theta\right) \leftarrow \text{IncSolve}(S_{\Theta},\Phi,\mathit{Bounds})$;
    \IF{$\mathit{result} =$ \texttt{unsat}}
        \RETURN \texttt{robust};
    \ENDIF
\ENDFOR
\RETURN \texttt{non-robust};
\end{algorithmic}
\end{algorithm}

Algorithm~\ref{alg:incgnnverify} describes our 
verification method for GNNs {that leverages incremental solving}.
First, the CSP is formulated and the bounds of variables are derived based on the rules of bound tightening {of Section~\ref{sec:bound-tight}} and layer-by-layer propagation (lines 1--6).
Next, we solve the formula $\Phi$ iteratively using an incremental solving mechanism (lines 7--15).
Initially, $\Phi=\varphi_{obj}$.
For node classification, $\varphi_{obj}$ only contains the variables representing the final embeddings of the target node $t$. For graph classification, $\varphi_{obj}$ contains the variables representing the sum of final embeddings of all nodes, along with the final summation and linear transformation layer.
In each iteration, new variables and the corresponding constraints are added to $\Phi$, and a MIP solver $S_{\Theta}$ is then called (line 11),
where $\Theta$ represents the generated cuts, which are redundant constraints produced by the solver to prune the search space.
If $\Phi$ is found to be unsatisfiable, then the GNN has been verified to be \texttt{robust}. Otherwise, the encoding of the previous layer is added to $\Phi$.
Finally, if the result of the last iteration 
remains satisfiable, then the GNN is verified as \texttt{non-robust}.
Note that, when each additional layer is added in the graph, the number of neighbouring nodes can increase significantly. Therefore, our algorithm effectively reduces the size of the encoding in early iterations, which leads to a more efficient exact verification process.

Finally, we show the correctness of Algorithm~\ref{alg:incgnnverify} with the proof in Appendix~\ref{appx:incremental}.

\begin{theorem}
\label{thm:correctness}
    Given a GNN $f$,
    an attributed directed graph $G$, and a perturbation space $\mathcal{Q}(G)$,
    $f$ is adversarially robust {(for a node $t$, if it is a node-classification GNN)} if and only if Algorithm~\ref{alg:incgnnverify} returns \texttt{robust}.
\end{theorem}
\noindent

\section{Experimental Evaluation}
\label{sec:exp}

\subsection{Implementation and Setup}
We implement our method as \textsc{GNNev}, a versatile and efficient exact\footnote{For edge addition/deletion, \textsc{GNNev} is a conditional exact verifier with respect to a given fragile edge set $F$.} verifier specifically designed for {message-passing} GNNs\footnote{\url{https://github.com/minghao-liu/GNNev}}.
\textsc{GNNev} calls the underlying Gurobi 11.0.3 solver\footnote{\url{https://www.gurobi.com}} for MIP solving.
Compared to available exact GNN verifiers, \textsc{GNNev} has three advantages: (i) it accepts three commonly used aggregation functions, {sum}, {max}, and {mean}, which broadens usability, (ii) it allows for both node attribute perturbation and edge addition/deletion, making it applicable to a wider range of {attack} scenarios, and (iii) it is designed as a Python library that accepts trained models built by the {SAGEConv} module in {PyTorch Geometric} \cite{FeyL19PyG}, and thus can be used directly in the same environment as model implementation and training.

To evaluate performance, we trained a batch of GraphSAGE models on two citation network datasets, Cora and CiteSeer \cite{SenNBGGE08}, two real-world fraud datasets, Amazon \cite{McAuleyL13} and Yelp \cite{RayanaA15}, {and two biochemical datasets, MUTAG and ENZYMES \cite{morris2020tudataset}. The first four datasets are for node classification and the last two are for graph classification.} We set the number of GNN layers {$K =3$} and aggregation functions $\mathbf{aggr} \in \left\{\text{mean}, \text{max}, \text{sum}\right\}$.
We regard robustness verification for each node/graph as a task, which checks prediction stability for any set of admissible perturbations up to a given {budget}.
For evaluation purposes, we run experiments with attribute and structural perturbation separately, but \textsc{GNNev} supports their simultaneous use.
All experiments were conducted on a server with Intel Xeon Gold 6252 @ 2.10GHz CPU, 252GB RAM, and Ubuntu 18.04 OS.
A verifier, including both \textsc{GNNev} and the baselines, ran each task on a single CPU core for fair comparison.
The time limit for each task was set to 300s.
More details of the datasets and experimental setups can be found in Appendix~\ref{appx:training}.

\begin{figure}[t]
    \centering
    
    \begin{subfigure}[t]{0.243\linewidth}
        \centering
        \includegraphics[trim={0cm 0cm 0cm 0cm}, clip, width=\textwidth]{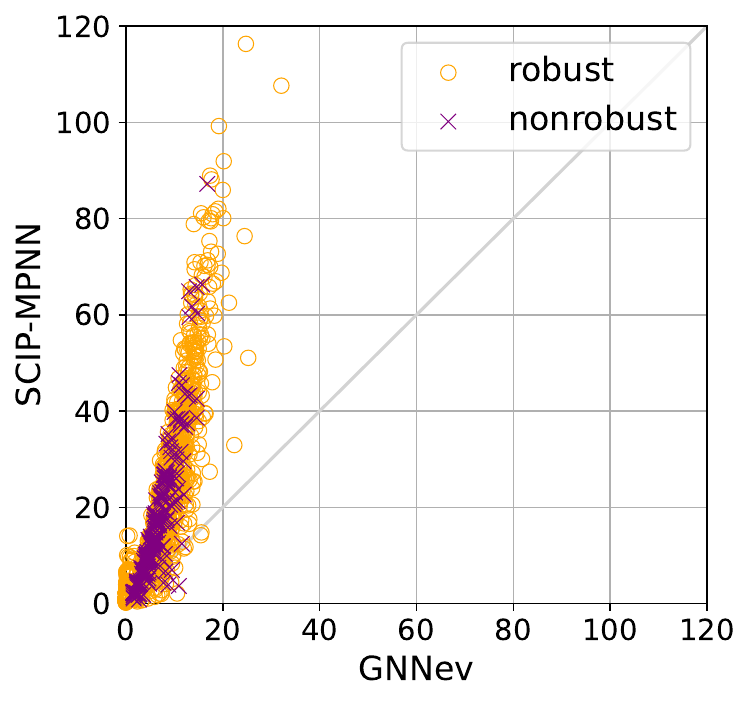}
        \caption{Cora}
        \label{fig:compare-cora-3lyr}
    \end{subfigure}
    \hfill
    \begin{subfigure}[t]{0.243\linewidth}
        \centering
        \includegraphics[trim={0cm 0cm 0cm 0.7cm}, clip, width=\textwidth]{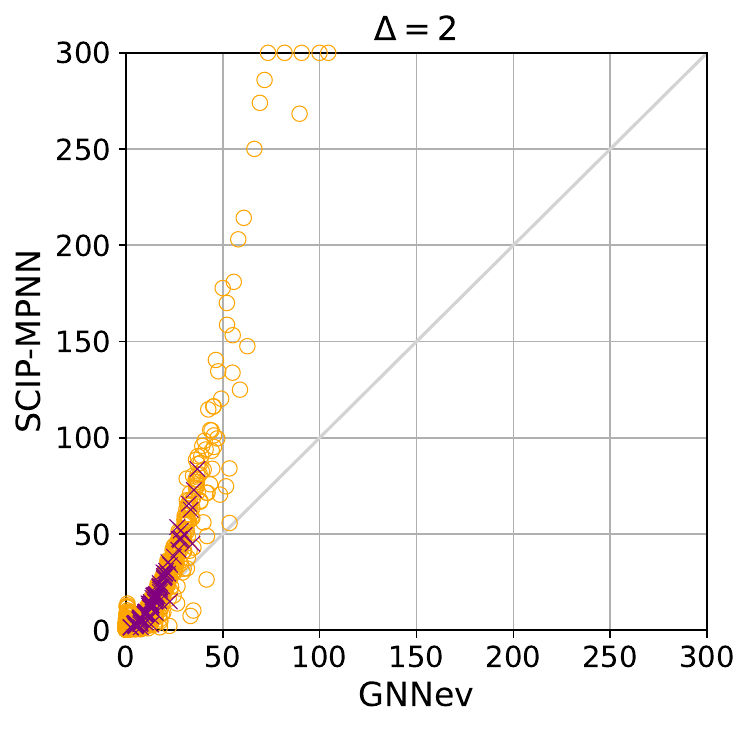}
        \caption{CiteSeer}
        \label{fig:compare-citeseer-3lyr}
    \end{subfigure}
    \hfill
    \begin{subfigure}[t]{0.243\linewidth}
        \centering
        \includegraphics[trim={0cm 0cm 0cm 0cm}, clip, width=\textwidth]{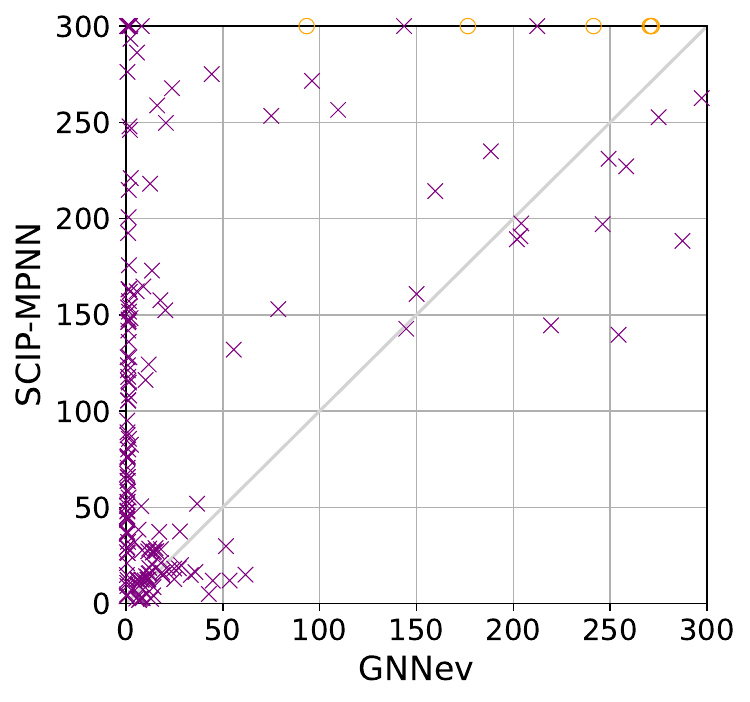}
        \caption{MUTAG}
        \label{fig:compare-mutag-3lyr}
    \end{subfigure}
    \hfill
    \begin{subfigure}[t]{0.243\linewidth}
        \centering
        \includegraphics[trim={0cm 0cm 0cm 0cm}, clip, width=\textwidth]{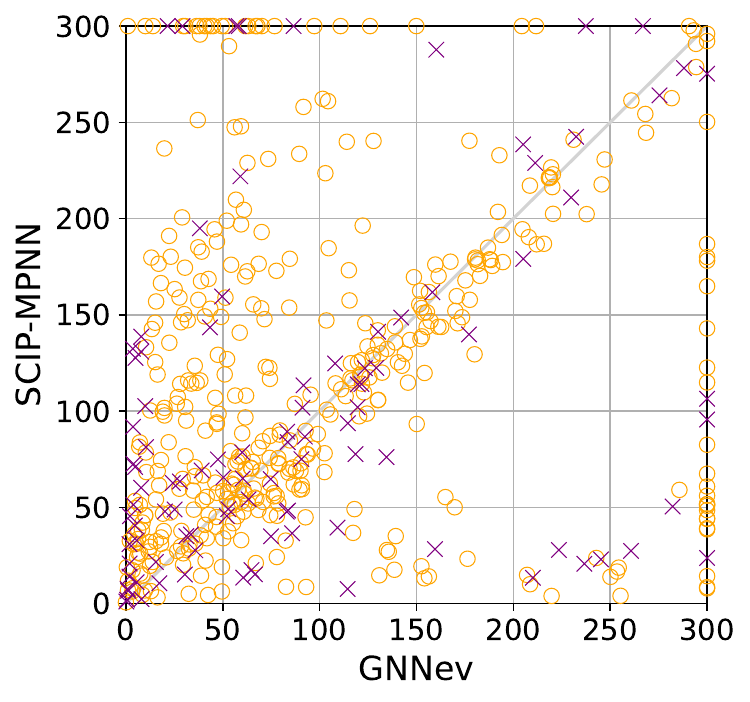}
        \caption{ENZYMES}
        \label{fig:compare-enzymes-3lyr}
    \end{subfigure}
    \caption{Comparison of runtime (in seconds) for {sum}-aggregated GNNs {on node classification (a-b) and graph classification (c-d)}. Each data point represents a robustness verification task {with $\Delta=2$}. A point above the diagonal line indicates that \textsc{GNNev} outperforms \textsc{SCIP-MPNN} on that task.
    }
    \label{fig:compare}
\end{figure}

\subsection{Comparison with Baselines on Sum-aggregation} 
We compare \textsc{GNNev} to \textsc{SCIP-MPNN} \cite{HojnyZCM24} on edge deletion\footnote{We note that \textsc{SCIP-MPNN} does not implement edge addition and
its interface does not support attribute perturbations.},  the only {MIP-based} exact verifier immediately applicable to adversarial robustness of GNNs\footnote{We remark that \textsc{RobLight} \cite{lu2025roblight} outperforms our tool but it is not MIP-based, limited to structural perturbations only and does not allow certificate generation in the sense of \cite{BarrettHS26}.}.
\textsc{SCIP-MPNN} includes five variants with different underlying solvers and bound tightening strategies.
To make fair comparison, we use the two versions that also call Gurobi:
{GRBbasic} and {GRBsbt}\footnote{We excluded the three variants based on SCIP,
noting that the experimental results in \cite{HojnyZCM24} showed that {GRBbasic} and {GRBsbt} have notably shorter runtime.}.
For each task, we ran these two versions separately and reported
the shorter runtime 
as the result of \textsc{SCIP-MPNN}.
To align with the restrictions of \textsc{SCIP-MPNN},
we utilised GNNs with {sum} aggregation, only allowed edge deletions (i.e., $F=E$), and adapted our verification objective\footnote{According to \citet{HojnyZCM24}, Eq.~(11) and Section 4.2, given the predicted class $\hat{c}_t$ for target node $t$, \textsc{SCIP-MPNN} aims to verify that $\mathbf{h}^{(K)}_t[\hat{c}_t] > \mathbf{h}^{(K)}_t[(\hat{c}_t+1)\%|C|]$ holds for all perturbed graphs. 
This simpler objective, which provides weaker robustness guarantees than ours stated in Eq.~(14), is used for tractability reasons.}.

\subsubsection{Results.}

\figurename~\ref{fig:compare} demonstrates the comparison of runtime between \textsc{GNNev} and \textsc{SCIP-MPNN} with edge deletions.
For node-classification datasets, Cora and CiteSeer, \textsc{GNNev} solved all tasks within 40s and 120s, respectively, outperforming \textsc{SCIP-MPNN} on the vast majority of tasks.
Note that \textsc{GNNev} also solved 5 challenging tasks on which \textsc{SCIP-MPNN} failed within the time limit on CiteSeer.
For graph-classification datasets, \textsc{GNNev} demonstrated competitive performance.
On MUTAG, \textsc{GNNev} solved most tasks within seconds,
likely due to the simpler and more efficient encoding produced through incremental solving.
On ENZYMES, a highly challenging dataset, \textsc{GNNev} also performed well 
compared to
\textsc{SCIP-MPNN} by completing {6.1\%} of the solvable tasks that \textsc{SCIP-MPNN} could not complete within the time limit, and reducing runtime on {54.8\%} of the solvable tasks.
Our experiments thus indicate that \textsc{GNNev} is a highly promising approach for exact verification.
See Appendix~\ref{appx:fullexp} for detailed comparison results on different perturbation budgets.

\subsection{Performance on All Aggregation Functions}
Next, we systematically evaluated the performance of \textsc{GNNev} on {sum}-, {max}- and {mean}-aggregated GNNs {with structural perturbations}.
For this experiment we set the {strong} verification objective as Eq.~(\ref{eq:obj}),
which is more challenging to compute than \citet{HojnyZCM24}, Eq.~(11).
Due to the large number of nodes in Amazon and Yelp, we randomly selected 1,000 nodes for verification.
Two fragile edge sets $F$ were used: (i) \emph{edge deletions} are allowed for all edges, i.e., $F = E$, and (ii) \emph{edge additions} are allowed for selected set, i.e., for each node $v$, several sampled non-edges $(u, v) \notin E$ are added to $F$.
See Appendix~\ref{appx:fullexp} for more details.

\subsubsection{Results.}

\begin{figure}[t]
    \centering
    
    \begin{subfigure}[b]{0.47\textwidth}
        \centering
        \includegraphics[trim={0.3cm 0cm 0.24cm 0cm}, clip, width=\linewidth]{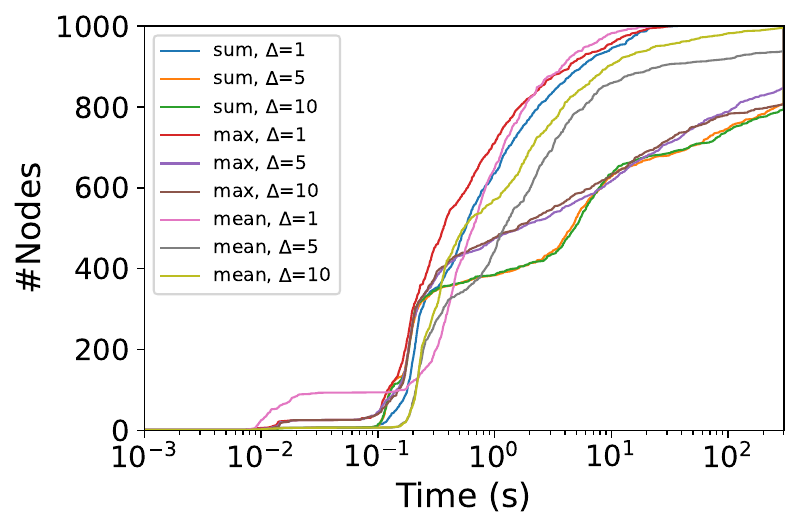}
        \caption{Amazon}
        \label{fig:time_amazon_delete_main}
    \end{subfigure}
    \hfill
    \begin{subfigure}[b]{0.47\textwidth}
        \centering
        \includegraphics[trim={0.3cm 0cm 0.24cm 0cm}, clip, width=\linewidth]{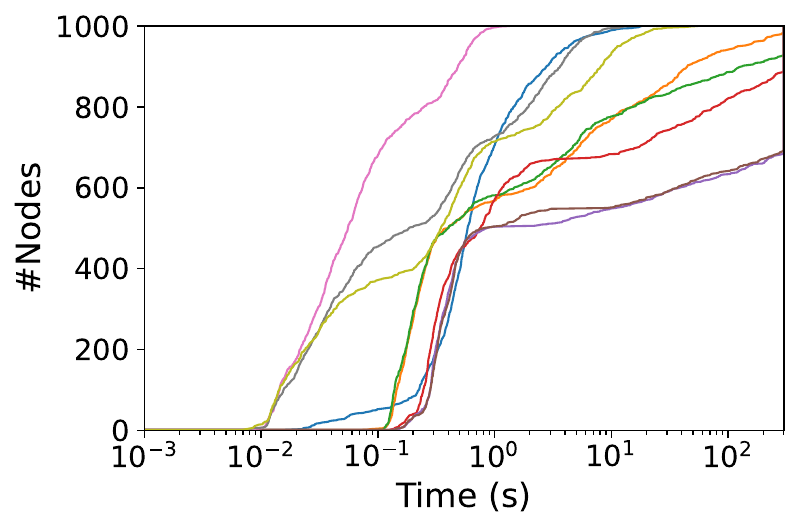}
        \caption{Yelp}
        \label{fig:time_yelp_delete_main}
    \end{subfigure}

    \begin{subfigure}[b]{0.47\textwidth}
        \centering
        \includegraphics[trim={0.26cm 0cm 0.24cm 0cm}, clip, width=\linewidth]{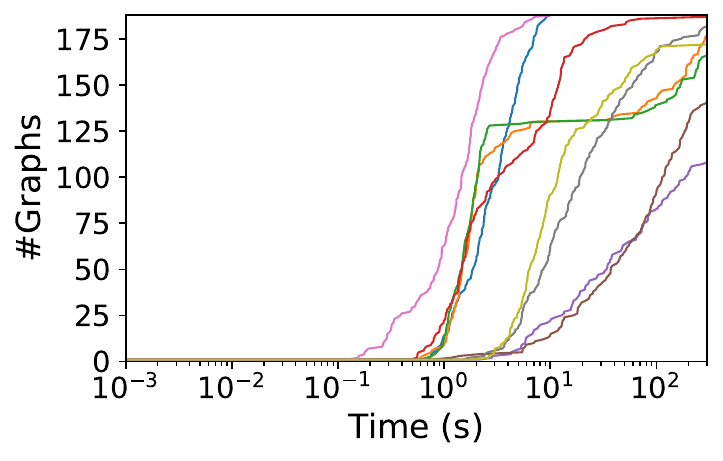}
        \caption{MUTAG}
        \label{fig:time_mutag_delete_main}
    \end{subfigure}
    \hfill
    \begin{subfigure}[b]{0.47\textwidth}
        \centering
        \includegraphics[trim={0.26cm 0cm 0.24cm 0cm}, clip, width=\linewidth]{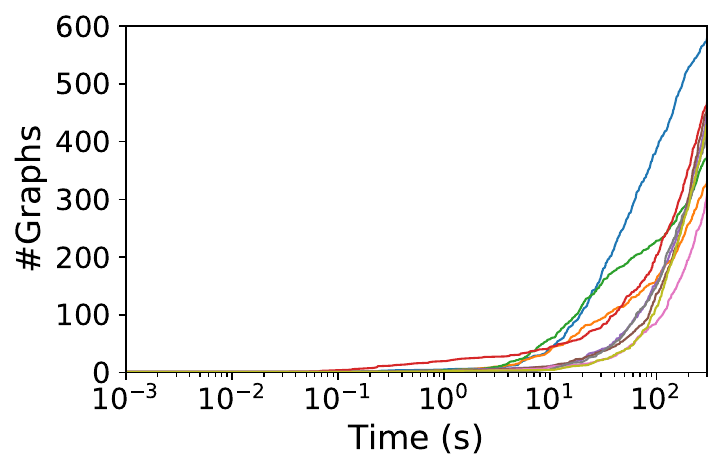}
        \caption{ENZYMES}
        \label{fig:time_enzymes_delete_main}
    \end{subfigure}
    
    \caption{The number of tasks solved by \textsc{GNNev} on real-world node-classification (a-b) and graph-classification (c-d) datasets plotted against runtime under different aggregations and budgets. Only edge deletions are allowed ($F=E$).
    The verification objective is as defined in Section~\ref{sec:encoding}, Eq.~(\ref{eq:obj}).
    Full results are shown in \figurename~\ref{fig:time} in Appendix~\ref{appx:fullexp}.}
    \label{fig:time_main}
\end{figure}

\figurename~\ref{fig:time_main} illustrates the runtime of \textsc{GNNev} on real-world node- and graph-classification datasets under different aggregations and budgets $\Delta$ for edge deletion; results on additional datasets and for edge addition are in  \figurename~\ref{fig:time}, \ref{fig:time_add} in Appendix~\ref{appx:fullexp}.
First, MUTAG and ENZYMES are more challenging, as incremental solving can reduce the number of variables representing both nodes and layers for node classification, whereas it can only reduce the number of variables representing layers for graph classification.
Second, across different aggregations, there was a performance drop for {max} aggregation when $\Delta \ge 5$, likely due to
Gurobi struggling to efficiently process max constraints using internal big-M encoding.
Moreover, for {sum} aggregation, the performance on Amazon and Yelp dropped
compared to Cora and CiteSeer.
This can be attributed to the much larger gap between upper and lower bounds of variables,
which is most pronounced for sum aggregation in \tablename~\ref{tab:bound_stats} in  Appendix~\ref{appx:fullexp}, where we see a significant increase in the gap after each layer.
Finally, we conducted ablation experiments to show the effectiveness of bound tightening and incremental solving, with results detailed in Appendix~\ref{appx:fullexp}.

\subsection{Adversarial Robustness Case Study}
We further use \textsc{GNNev} to analyse the robustness of GNNs for both structural and attribute perturbations, especially in high-stakes domains such as fraud detection and scientific discovery.
First, we performed evaluation of the number of (non-)robust tasks verified plotted against an increasing global structural perturbation budget, shown in \figurename~\ref{fig:evolution_delete}, \ref{fig:evolution_add} in Appendix~\ref{appx:fullexp}.
In most cases, instances were attacked successfully with a small budget.
Some GNNs exhibited quite weak robustness.
Specifically, we found that {mean}-aggregated GNNs were more vulnerable to adversarial perturbations than other types.
Moreover, \textsc{GNNev} was executed on Amazon and Yelp to verify mean-aggregated GNNs with attribute perturbation enabled. From the results shown in \figurename~\ref{fig:attrperturb} in Appendix~\ref{appx:fullexp}, even perturbing a single dimension of attributes can cause models to make wrong predictions in up to 75.8\% of tasks.

\begin{figure}[t]
    \centering
    \includegraphics[width=0.75\linewidth]{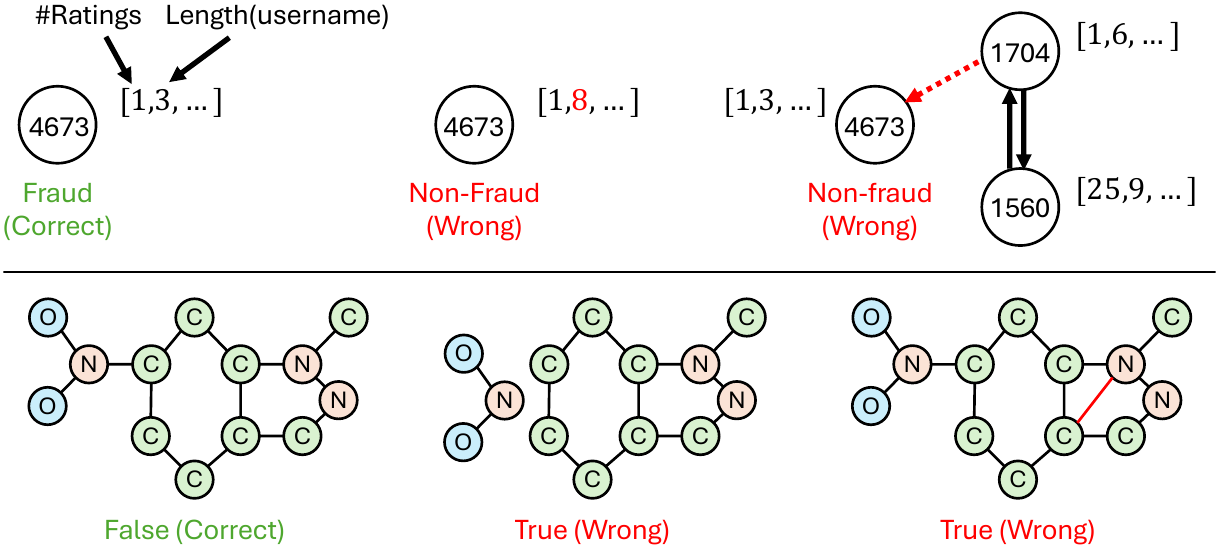}
    \caption{Adversarial examples found by \textsc{GNNev}. 
    Top: Either modifying an attribute {(here length of username)} or adding an edge caused a node in the Amazon dataset to be misclassified as non-fraudulent.
    Bottom: Adding or deleting an edge on a graph in the MUTAG dataset leads to wrong prediction of having the given property even if the perturbed graphs are invalid.}
    \label{fig:attack-cases}
\end{figure}

Within these results, we identified a number of adversarial attacks on non-robust instances, which demonstrate the vulnerability of GNNs.
\figurename~\ref{fig:attack-cases}
illustrates two cases from Amazon and MUTAG, respectively.
In Amazon, one of the attributes is the length of username, which is easily manipulated in practice. We configured \textsc{GNNev} to only allow perturbations of this attribute within the range $[1,10]$. The results showed that 29.1\% of nodes were successfully attacked for the mean-aggregated GNN.
In MUTAG, edge addition and deletion can easily lead to invalid structures, but \textsc{GNNev} found that mean-aggregated GNNs reported 12.7\% false positive results when $\Delta=1$.
These demonstrate that \textsc{GNNev} is able to gain useful insight into the susceptibility of GNN models to adversarial attacks, which is important for pre-deployment analysis for high-stakes applications.

\section{Conclusion and Future Work}
\label{sec:conclu}

We have developed an exact robustness verification method for message-passing GNNs supporting a range of aggregation functions, which is implemented in \textsc{GNNev}, a versatile and efficient verifier.
Our approach is based on a reduction to a constraint satisfaction problem with bound tightening, solved incrementally.
The support for two common aggregation functions, max and mean, is formulated here for the first time.
Experiments on a range of datasets
show that \textsc{GNNev} {can verify adversarial robustness of GNNs to attribute and structural perturbations. It}  outperforms state-of-the-art MIP-based baselines on node classification in both efficiency and functionality while remaining competitive on graph classification, and delivers strong performance on real-world fraud and biochemical datasets.
One practical limitation of our approach is that the computational complexity grows quickly with the size of the fragile edge set. 
Future work will include mitigating the impact of the fragile edge sets on performance by combining heuristic search and distributed frameworks, and developing hybrid approaches to integrate probabilistic verification for improving scalability.

\begin{credits}
\subsubsection{\ackname}
The authors would like to thank Michael Benedikt for the insightful comments and suggestions.
This work was supported by
the EPSRC Prosperity Partnership FAIR (grant number EP/V056883/1)
and
ELSA: European Lighthouse on Secure and Safe AI project (grant agreement No. 101070617 under UK guarantee).
MK receives funding from the ERC under the European Union’s Horizon 2020 research and innovation programme (FUN2MODEL, grant agreement No.~834115).
\end{credits}

%
%
%
\bibliography{main}

@inproceedings{BojchevskiG19,
  author       = {Aleksandar Bojchevski and
                  Stephan G{\"{u}}nnemann},
  //editor       = {Hanna M. Wallach and
                  Hugo Larochelle and
                  Alina Beygelzimer and
                  Florence d'Alch{\'{e}}{-}Buc and
                  Emily B. Fox and
                  Roman Garnett},
  title        = {Certifiable Robustness to Graph Perturbations},
  booktitle    = {NeurIPS},
  pages        = {8317--8328},
  year         = {2019},
  //url          = {https://proceedings.neurips.cc/paper/2019/hash/e2f374c3418c50bc30d67d5f7454a5b4-Abstract.html},
  timestamp    = {Mon, 16 May 2022 15:41:51 +0200},
  biburl       = {https://dblp.org/rec/conf/nips/BojchevskiG19.bib},
  bibsource    = {dblp computer science bibliography, https://dblp.org}
}

@inproceedings{GilmerSRVD17,
  author       = {Justin Gilmer and
                  Samuel S. Schoenholz and
                  Patrick F. Riley and
                  Oriol Vinyals and
                  George E. Dahl},
  //editor       = {Doina Precup and
                  Yee Whye Teh},
  title        = {Neural Message Passing for Quantum Chemistry},
  booktitle    = {{ICML}},
  //series       = {Proceedings of Machine Learning Research},
  //volume       = {70},
  pages        = {1263--1272},
  //publisher    = {{PMLR}},
  year         = {2017},
  //url          = {http://proceedings.mlr.press/v70/gilmer17a.html},
  timestamp    = {Wed, 29 May 2019 08:41:45 +0200},
  biburl       = {https://dblp.org/rec/conf/icml/GilmerSRVD17.bib},
  bibsource    = {dblp computer science bibliography, https://dblp.org}
}

@inproceedings{WangQL0JWFYZY19,
  author       = {Daixin Wang and
                  Yuan Qi and
                  Jianbin Lin and
                  Peng Cui and
                  Quanhui Jia and
                  Zhen Wang and
                  Yanming Fang and
                  Quan Yu and
                  Jun Zhou and
                  Shuang Yang},
  //editor       = {Jianyong Wang and
                  Kyuseok Shim and
                  Xindong Wu},
  title        = {A Semi-Supervised Graph Attentive Network for Financial Fraud Detection},
  booktitle    = {{ICDM}},
  pages        = {598--607},
  //publisher    = {{IEEE}},
  year         = {2019},
  //url          = {https://doi.org/10.1109/ICDM.2019.00070},
  //doi          = {10.1109/ICDM.2019.00070},
  timestamp    = {Tue, 26 Jan 2021 13:51:21 +0100},
  biburl       = {https://dblp.org/rec/conf/icdm/WangQL0JWFYZY19.bib},
  bibsource    = {dblp computer science bibliography, https://dblp.org}
}

@article{MotieR24,
  author       = {Soroor Motie and
                  Bijan Raahemi},
  title        = {Financial fraud detection using graph neural networks: {A} systematic review},
  journal      = {Expert Syst. Appl.},
  volume       = {240},
  pages        = {122156},
  year         = {2024},
  //url          = {https://doi.org/10.1016/j.eswa.2023.122156},
  //doi          = {10.1016/J.ESWA.2023.122156},
  timestamp    = {Fri, 31 May 2024 21:07:07 +0200},
  biburl       = {https://dblp.org/rec/journals/eswa/MotieR24.bib},
  bibsource    = {dblp computer science bibliography, https://dblp.org}
}

@inproceedings{CasasGLU20,
  author       = {Sergio Casas and
                  Cole Gulino and
                  Renjie Liao and
                  Raquel Urtasun},
  title        = {{SpAGNN}: Spatially-Aware Graph Neural Networks for Relational Behavior
                  Forecasting from Sensor Data},
  booktitle    = {{ICRA}},
  pages        = {9491--9497},
  //publisher    = {{IEEE}},
  year         = {2020},
  //url          = {https://doi.org/10.1109/ICRA40945.2020.9196697},
  //doi          = {10.1109/ICRA40945.2020.9196697},
  timestamp    = {Mon, 28 Sep 2020 12:19:28 +0200},
  biburl       = {https://dblp.org/rec/conf/icra/0002GLU20.bib},
  bibsource    = {dblp computer science bibliography, https://dblp.org}
}

@inproceedings{CaiWS021,
  author       = {Peide Cai and
                  Hengli Wang and
                  Yuxiang Sun and
                  Ming Liu},
  title        = {{DiGNet}: Learning Scalable Self-Driving Policies for Generic Traffic Scenarios with Graph Neural Networks},
  booktitle    = {{IROS}},
  pages        = {8979--8984},
  //publisher    = {{IEEE}},
  year         = {2021},
  //url          = {https://doi.org/10.1109/IROS51168.2021.9636376},
  //doi          = {10.1109/IROS51168.2021.9636376},
  timestamp    = {Mon, 05 Feb 2024 20:27:21 +0100},
  biburl       = {https://dblp.org/rec/conf/iros/CaiWS021.bib},
  bibsource    = {dblp computer science bibliography, https://dblp.org}
}

@article{GaoYWWDL24,
  author       = {Chao Gao and
                  Shu Yin and
                  Haiqiang Wang and
                  Zhen Wang and
                  Zhanwei Du and
                  Xuelong Li},
  title        = {Medical-Knowledge-Based Graph Neural Network for Medication Combination Prediction},
  journal      = {{IEEE} Trans. Neural Networks Learn. Syst.},
  volume       = {35},
  number       = {10},
  pages        = {13246--13257},
  year         = {2024},
  //url          = {https://doi.org/10.1109/TNNLS.2023.3266490},
  //doi          = {10.1109/TNNLS.2023.3266490},
  timestamp    = {Tue, 22 Oct 2024 21:09:31 +0200},
  biburl       = {https://dblp.org/rec/journals/tnn/GaoYWWDL24.bib},
  bibsource    = {dblp computer science bibliography, https://dblp.org}
}

@article{li2022grlhealth,
  title={Graph representation learning in biomedicine and healthcare},
  author={Li, Michelle M and
          Huang, Kexin and
          Zitnik, Marinka},
  journal={Nat. Biomed. Eng.},
  volume={6},
  number={12},
  pages={1353--1369},
  year={2022},
  //publisher={Nature Publishing Group UK London}
}

@article{jha2022proteingnn,
  title={Prediction of protein--protein interaction using graph neural networks},
  author={Jha, Kanchan and
          Saha, Sriparna and
          Singh, Hiteshi},
  journal={Scientific Reports},
  volume={12},
  number={1},
  pages={8360},
  year={2022},
  //publisher={Nature Publishing Group UK London}
}

@article{reiser2022gnnscience,
  title={Graph neural networks for materials science and chemistry},
  author={Reiser, Patrick and
          Neubert, Marlen and
          Eberhard, Andr{\'e} and
          Torresi, Luca and
          Zhou, Chen and
          Shao, Chen and
          Metni, Houssam and
          van Hoesel, Clint and
          Schopmans, Henrik and
          Sommer, Timo and
          Friederich, Pascal},
  journal={Commun. Mater.},
  volume={3},
  number={1},
  pages={93},
  year={2022},
  //publisher={Nature Publishing Group UK London}
}

@inproceedings{ZugnerAG18,
  author       = {Daniel Z{\"{u}}gner and
                  Amir Akbarnejad and
                  Stephan G{\"{u}}nnemann},
  //editor       = {Yike Guo and
                  Faisal Farooq},
  title        = {Adversarial Attacks on Neural Networks for Graph Data},
  booktitle    = {{KDD}},
  pages        = {2847--2856},
  //publisher    = {{ACM}},
  year         = {2018},
  //url          = {https://doi.org/10.1145/3219819.3220078},
  //doi          = {10.1145/3219819.3220078},
  timestamp    = {Wed, 21 Nov 2018 12:44:27 +0100},
  biburl       = {https://dblp.org/rec/conf/kdd/ZugnerAG18.bib},
  bibsource    = {dblp computer science bibliography, https://dblp.org}
}

@inproceedings{DaiLTHWZS18,
  author       = {Hanjun Dai and
                  Hui Li and
                  Tian Tian and
                  Xin Huang and
                  Lin Wang and
                  Jun Zhu and
                  Le Song},
  //editor       = {Jennifer G. Dy and
                  Andreas Krause},
  title        = {Adversarial Attack on Graph Structured Data},
  booktitle    = {{ICML}},
  //series       = {Proceedings of Machine Learning Research},
  //volume       = {80},
  pages        = {1123--1132},
  //publisher    = {{PMLR}},
  year         = {2018},
  //url          = {http://proceedings.mlr.press/v80/dai18b.html},
  timestamp    = {Wed, 03 Apr 2019 18:17:30 +0200},
  biburl       = {https://dblp.org/rec/conf/icml/DaiLTHWZS18.bib},
  bibsource    = {dblp computer science bibliography, https://dblp.org}
}

@inproceedings{TaoCSHWC21,
  author       = {Shuchang Tao and
                  Qi Cao and
                  Huawei Shen and
                  Junjie Huang and
                  Yunfan Wu and
                  Xueqi Cheng},
  //editor       = {Gianluca Demartini and
                  Guido Zuccon and
                  J. Shane Culpepper and
                  Zi Huang and
                  Hanghang Tong},
  title        = {Single Node Injection Attack against Graph Neural Networks},
  booktitle    = {{CIKM}},
  pages        = {1794--1803},
  //publisher    = {{ACM}},
  year         = {2021},
  //url          = {https://doi.org/10.1145/3459637.3482393},
  //doi          = {10.1145/3459637.3482393},
  timestamp    = {Sun, 19 Jan 2025 13:12:22 +0100},
  biburl       = {https://dblp.org/rec/conf/cikm/TaoCSHWC21.bib},
  bibsource    = {dblp computer science bibliography, https://dblp.org}
}

@inproceedings{HuangKWW17,
  author       = {Xiaowei Huang and
                  Marta Kwiatkowska and
                  Sen Wang and
                  Min Wu},
  //editor       = {Rupak Majumdar and
                  Viktor Kuncak},
  title        = {Safety Verification of Deep Neural Networks},
  booktitle    = {{CAV}},
  //series       = {Lecture Notes in Computer Science},
  //volume       = {10426},
  pages        = {3--29},
  //publisher    = {Springer},
  year         = {2017},
  //url          = {https://doi.org/10.1007/978-3-319-63387-9\_1},
  //doi          = {10.1007/978-3-319-63387-9\_1},
  timestamp    = {Wed, 22 Dec 2021 17:17:02 +0100},
  biburl       = {https://dblp.org/rec/conf/cav/HuangKWW17.bib},
  bibsource    = {dblp computer science bibliography, https://dblp.org}
}

@inproceedings{BoopathyWC0D19,
  author       = {Akhilan Boopathy and
                  Tsui{-}Wei Weng and
                  Pin{-}Yu Chen and
                  Sijia Liu and
                  Luca Daniel},
  title        = {{CNN-Cert}: An Efficient Framework for Certifying Robustness of Convolutional Neural Networks},
  booktitle    = {{AAAI}},
  pages        = {3240--3247},
  //publisher    = {{AAAI} Press},
  year         = {2019},
  //url          = {https://doi.org/10.1609/aaai.v33i01.33013240},
  //doi          = {10.1609/AAAI.V33I01.33013240},
  timestamp    = {Mon, 04 Sep 2023 12:29:24 +0200},
  biburl       = {https://dblp.org/rec/conf/aaai/BoopathyWC0D19.bib},
  bibsource    = {dblp computer science bibliography, https://dblp.org}
}

@inproceedings{TranBXJ20,
  author       = {Hoang{-}Dung Tran and
                  Stanley Bak and
                  Weiming Xiang and
                  Taylor T. Johnson},
  //editor       = {Shuvendu K. Lahiri and
                  Chao Wang},
  title        = {Verification of Deep Convolutional Neural Networks Using {ImageStars}},
  booktitle    = {{CAV}},
  //series       = {Lecture Notes in Computer Science},
  //volume       = {12224},
  pages        = {18--42},
  //publisher    = {Springer},
  year         = {2020},
  //url          = {https://doi.org/10.1007/978-3-030-53288-8\_2},
  //doi          = {10.1007/978-3-030-53288-8\_2},
  timestamp    = {Mon, 08 Apr 2024 20:42:03 +0200},
  biburl       = {https://dblp.org/rec/conf/cav/TranBXJ20.bib},
  bibsource    = {dblp computer science bibliography, https://dblp.org}
}

@inproceedings{WangPWYJ18,
  author       = {Shiqi Wang and
                  Kexin Pei and
                  Justin Whitehouse and
                  Junfeng Yang and
                  Suman Jana},
  //editor       = {Samy Bengio and
                  Hanna M. Wallach and
                  Hugo Larochelle and
                  Kristen Grauman and
                  Nicol{\`{o}} Cesa{-}Bianchi and
                  Roman Garnett},
  title        = {Efficient Formal Safety Analysis of Neural Networks},
  booktitle    = {NeurIPS},
  pages        = {6369--6379},
  year         = {2018},
  //url          = {https://proceedings.neurips.cc/paper/2018/hash/2ecd2bd94734e5dd392d8678bc64cdab-Abstract.html},
  timestamp    = {Mon, 16 May 2022 15:41:51 +0200},
  biburl       = {https://dblp.org/rec/conf/nips/WangPWYJ18.bib},
  bibsource    = {dblp computer science bibliography, https://dblp.org}
}

@inproceedings{ZhangWCHD18,
  author       = {Huan Zhang and
                  Tsui{-}Wei Weng and
                  Pin{-}Yu Chen and
                  Cho{-}Jui Hsieh and
                  Luca Daniel},
  //editor       = {Samy Bengio and
                  Hanna M. Wallach and
                  Hugo Larochelle and
                  Kristen Grauman and
                  Nicol{\`{o}} Cesa{-}Bianchi and
                  Roman Garnett},
  title        = {Efficient Neural Network Robustness Certification with General Activation Functions},
  booktitle    = {NeurIPS},
  pages        = {4944--4953},
  year         = {2018},
  //url          = {https://proceedings.neurips.cc/paper/2018/hash/d04863f100d59b3eb688a11f95b0ae60-Abstract.html},
  timestamp    = {Mon, 16 May 2022 15:41:51 +0200},
  biburl       = {https://dblp.org/rec/conf/nips/ZhangWCHD18.bib},
  bibsource    = {dblp computer science bibliography, https://dblp.org}
}

@inproceedings{WengZCSHDBD18,
  author       = {Tsui{-}Wei Weng and
                  Huan Zhang and
                  Hongge Chen and
                  Zhao Song and
                  Cho{-}Jui Hsieh and
                  Luca Daniel and
                  Duane S. Boning and
                  Inderjit S. Dhillon},
  //editor       = {Jennifer G. Dy and
                  Andreas Krause},
  title        = {Towards Fast Computation of Certified Robustness for {ReLU} Networks},
  booktitle    = {{ICML}},
  //series       = {Proceedings of Machine Learning Research},
  //volume       = {80},
  pages        = {5273--5282},
  //publisher    = {{PMLR}},
  year         = {2018},
  //url          = {http://proceedings.mlr.press/v80/weng18a.html},
  timestamp    = {Wed, 02 Dec 2020 16:43:27 +0100},
  biburl       = {https://dblp.org/rec/conf/icml/WengZCSHDBD18.bib},
  bibsource    = {dblp computer science bibliography, https://dblp.org}
}

@inproceedings{BotoevaKKLM20,
  author       = {Elena Botoeva and
                  Panagiotis Kouvaros and
                  Jan Kronqvist and
                  Alessio Lomuscio and
                  Ruth Misener},
  title        = {Efficient Verification of {ReLU}-Based Neural Networks via Dependency Analysis},
  booktitle    = {{AAAI}},
  pages        = {3291--3299},
  //publisher    = {{AAAI} Press},
  year         = {2020},
  //url          = {https://doi.org/10.1609/aaai.v34i04.5729},
  //doi          = {10.1609/AAAI.V34I04.5729},
  timestamp    = {Mon, 04 Sep 2023 16:50:23 +0200},
  biburl       = {https://dblp.org/rec/conf/aaai/BotoevaKKLM20.bib},
  bibsource    = {dblp computer science bibliography, https://dblp.org}
}

@inproceedings{KatzBDJK17,
  author       = {Guy Katz and
                  Clark W. Barrett and
                  David L. Dill and
                  Kyle Julian and
                  Mykel J. Kochenderfer},
  //editor       = {Rupak Majumdar and
                  Viktor Kuncak},
  title        = {{Reluplex}: An Efficient {SMT} Solver for Verifying Deep Neural Networks},
  booktitle    = {{CAV}},
  //series       = {Lecture Notes in Computer Science},
  //volume       = {10426},
  pages        = {97--117},
  //publisher    = {Springer},
  year         = {2017},
  //url          = {https://doi.org/10.1007/978-3-319-63387-9\_5},
  //doi          = {10.1007/978-3-319-63387-9\_5},
  timestamp    = {Wed, 25 Sep 2019 18:02:04 +0200},
  biburl       = {https://dblp.org/rec/conf/cav/KatzBDJK17.bib},
  bibsource    = {dblp computer science bibliography, https://dblp.org}
}

@inproceedings{KoLWDWL19,
  author       = {Ching{-}Yun Ko and
                  Zhaoyang Lyu and
                  Lily Weng and
                  Luca Daniel and
                  Ngai Wong and
                  Dahua Lin},
  //editor       = {Kamalika Chaudhuri and
                  Ruslan Salakhutdinov},
  title        = {{POPQORN}: Quantifying Robustness of Recurrent Neural Networks},
  booktitle    = {{ICML}},
  //series       = {Proceedings of Machine Learning Research},
  //volume       = {97},
  pages        = {3468--3477},
  //publisher    = {{PMLR}},
  year         = {2019},
  //url          = {http://proceedings.mlr.press/v97/ko19a.html},
  timestamp    = {Tue, 11 Jun 2019 15:37:38 +0200},
  biburl       = {https://dblp.org/rec/conf/icml/KoLWDWL19.bib},
  bibsource    = {dblp computer science bibliography, https://dblp.org}
}

@inproceedings{DuJSZLSFYB021,
  author       = {Tianyu Du and
                  Shouling Ji and
                  Lujia Shen and
                  Yao Zhang and
                  Jinfeng Li and
                  Jie Shi and
                  Chengfang Fang and
                  Jianwei Yin and
                  Raheem Beyah and
                  Ting Wang},
  //editor       = {Yongdae Kim and
                  Jong Kim and
                  Giovanni Vigna and
                  Elaine Shi},
  title        = {{Cert-RNN}: Towards Certifying the Robustness of Recurrent Neural Networks},
  booktitle    = {{CCS}},
  pages        = {516--534},
  //publisher    = {{ACM}},
  year         = {2021},
  //url          = {https://doi.org/10.1145/3460120.3484538},
  //doi          = {10.1145/3460120.3484538},
  timestamp    = {Sun, 19 Jan 2025 13:26:34 +0100},
  biburl       = {https://dblp.org/rec/conf/ccs/DuJSZLSFYB021.bib},
  bibsource    = {dblp computer science bibliography, https://dblp.org}
}

@inproceedings{AkintundeKLP19,
  author       = {Michael E. Akintunde and
                  Andreea Kevorchian and
                  Alessio Lomuscio and
                  Edoardo Pirovano},
  title        = {Verification of {RNN}-Based Neural Agent-Environment Systems},
  booktitle    = {{AAAI}},
  pages        = {6006--6013},
  //publisher    = {{AAAI} Press},
  year         = {2019},
  //url          = {https://doi.org/10.1609/aaai.v33i01.33016006},
  //doi          = {10.1609/AAAI.V33I01.33016006},
  timestamp    = {Mon, 04 Sep 2023 12:29:24 +0200},
  biburl       = {https://dblp.org/rec/conf/aaai/AkintundeKLP19.bib},
  bibsource    = {dblp computer science bibliography, https://dblp.org}
}

@inproceedings{ShiZCHH20,
  author       = {Zhouxing Shi and
                  Huan Zhang and
                  Kai{-}Wei Chang and
                  Minlie Huang and
                  Cho{-}Jui Hsieh},
  title        = {Robustness Verification for Transformers},
  booktitle    = {{ICLR}},
  //publisher    = {OpenReview.net},
  year         = {2020},
  //url          = {https://openreview.net/forum?id=BJxwPJHFwS},
  timestamp    = {Fri, 23 Oct 2020 16:16:10 +0200},
  biburl       = {https://dblp.org/rec/conf/iclr/ShiZCHH20.bib},
  bibsource    = {dblp computer science bibliography, https://dblp.org}
}

@inproceedings{ZugnerG19,
  author       = {Daniel Z{\"{u}}gner and
                  Stephan G{\"{u}}nnemann},
  //editor       = {Ankur Teredesai and
                  Vipin Kumar and
                  Ying Li and
                  R{\'{o}}mer Rosales and
                  Evimaria Terzi and
                  George Karypis},
  title        = {Certifiable Robustness and Robust Training for Graph Convolutional Networks},
  booktitle    = {{KDD}},
  pages        = {246--256},
  //publisher    = {{ACM}},
  year         = {2019},
  //url          = {https://doi.org/10.1145/3292500.3330905},
  //doi          = {10.1145/3292500.3330905},
  timestamp    = {Tue, 20 Aug 2024 07:54:44 +0200},
  biburl       = {https://dblp.org/rec/conf/kdd/ZugnerG19.bib},
  bibsource    = {dblp computer science bibliography, https://dblp.org}
}

@inproceedings{ZugnerG20,
  author       = {Daniel Z{\"{u}}gner and
                  Stephan G{\"{u}}nnemann},
  //editor       = {Rajesh Gupta and
                  Yan Liu and
                  Jiliang Tang and
                  B. Aditya Prakash},
  title        = {Certifiable Robustness of Graph Convolutional Networks under Structure Perturbations},
  booktitle    = {{KDD}},
  pages        = {1656--1665},
  //publisher    = {{ACM}},
  year         = {2020},
  //url          = {https://doi.org/10.1145/3394486.3403217},
  //doi          = {10.1145/3394486.3403217},
  timestamp    = {Tue, 09 Mar 2021 09:46:47 +0100},
  biburl       = {https://dblp.org/rec/conf/kdd/ZugnerG20.bib},
  bibsource    = {dblp computer science bibliography, https://dblp.org}
}

@inproceedings{JinSPZ20,
  author       = {Hongwei Jin and
                  Zhan Shi and
                  Venkata Jaya Shankar Ashish Peruri and
                  Xinhua Zhang},
  //editor       = {Hugo Larochelle and
                  Marc'Aurelio Ranzato and
                  Raia Hadsell and
                  Maria{-}Florina Balcan and
                  Hsuan{-}Tien Lin},
  title        = {Certified Robustness of Graph Convolution Networks for Graph Classification under Topological Attacks},
  booktitle    = {NeurIPS},
  year         = {2020},
  //url          = {https://proceedings.neurips.cc/paper/2020/hash/609a199881ca4ba9c95688235cd6ac5c-Abstract.html},
  timestamp    = {Tue, 19 Jan 2021 15:57:22 +0100},
  biburl       = {https://dblp.org/rec/conf/nips/JinSPZ20.bib},
  bibsource    = {dblp computer science bibliography, https://dblp.org}
}

@inproceedings{BojchevskiKG20,
  author       = {Aleksandar Bojchevski and
                  Johannes Klicpera and
                  Stephan G{\"{u}}nnemann},
  title        = {Efficient Robustness Certificates for Discrete Data: Sparsity-Aware Randomized Smoothing for Graphs, Images and More},
  booktitle    = {{ICML}},
  //series       = {Proceedings of Machine Learning Research},
  //volume       = {119},
  pages        = {1003--1013},
  //publisher    = {{PMLR}},
  year         = {2020},
  //url          = {http://proceedings.mlr.press/v119/bojchevski20a.html},
  timestamp    = {Tue, 15 Dec 2020 17:40:18 +0100},
  biburl       = {https://dblp.org/rec/conf/icml/BojchevskiKG20.bib},
  bibsource    = {dblp computer science bibliography, https://dblp.org}
}

@inproceedings{OsselinK023,
  author       = {Pierre Osselin and
                  Henry Kenlay and
                  Xiaowen Dong},
  //editor       = {Robin J. Evans and
                  Ilya Shpitser},
  title        = {Structure-aware robustness certificates for graph classification},
  booktitle    = {{UAI}},
  //series       = {Proceedings of Machine Learning Research},
  //volume       = {216},
  pages        = {1596--1605},
  //publisher    = {{PMLR}},
  year         = {2023},
  //url          = {https://proceedings.mlr.press/v216/osselin23a.html},
  timestamp    = {Mon, 28 Aug 2023 17:23:08 +0200},
  biburl       = {https://dblp.org/rec/conf/uai/OsselinK023.bib},
  bibsource    = {dblp computer science bibliography, https://dblp.org}
}

@inproceedings{AnZZLSHYLQ24,
  author       = {Dongdong An and
                  Hao Zhang and
                  Qin Zhao and
                  Jing Liu and
                  Jianqi Shi and
                  Yanhong Huang and
                  Yang Yang and
                  Xu Liu and
                  Shengchao Qin},
  //editor       = {Kazuhiro Ogata and
                  Dominique M{\'{e}}ry and
                  Meng Sun and
                  Shaoying Liu},
  title        = {Graph Convolutional Network Robustness Verification Algorithm Based on Dual Approximation},
  booktitle    = {{ICFEM}},
  //series       = {Lecture Notes in Computer Science},
  //volume       = {15394},
  pages        = {146--161},
  //publisher    = {Springer},
  year         = {2024},
  //url          = {https://doi.org/10.1007/978-981-96-0617-7\_9},
  //doi          = {10.1007/978-981-96-0617-7\_9},
  timestamp    = {Tue, 28 Jan 2025 09:07:02 +0100},
  biburl       = {https://dblp.org/rec/conf/icfem/AnZZLSHYLQ24.bib},
  bibsource    = {dblp computer science bibliography, https://dblp.org}
}

@inproceedings{TjengXT19,
  author       = {Vincent Tjeng and
                  Kai Yuanqing Xiao and
                  Russ Tedrake},
  title        = {Evaluating Robustness of Neural Networks with Mixed Integer Programming},
  booktitle    = {{ICLR}},
  //publisher    = {OpenReview.net},
  year         = {2019},
  //url          = {https://openreview.net/forum?id=HyGIdiRqtm},
  timestamp    = {Tue, 10 Aug 2021 17:46:21 +0200},
  biburl       = {https://dblp.org/rec/conf/iclr/TjengXT19.bib},
  bibsource    = {dblp computer science bibliography, https://dblp.org}
}

@inproceedings{BunelTTKM18,
  author       = {Rudy Bunel and
                  Ilker Turkaslan and
                  Philip H. S. Torr and
                  Pushmeet Kohli and
                  Pawan Kumar Mudigonda},
  //editor       = {Samy Bengio and
                  Hanna M. Wallach and
                  Hugo Larochelle and
                  Kristen Grauman and
                  Nicol{\`{o}} Cesa{-}Bianchi and
                  Roman Garnett},
  title        = {A Unified View of Piecewise Linear Neural Network Verification},
  booktitle    = {NeurIPS},
  pages        = {4795--4804},
  year         = {2018},
  //url          = {https://proceedings.neurips.cc/paper/2018/hash/be53d253d6bc3258a8160556dda3e9b2-Abstract.html},
  timestamp    = {Mon, 16 May 2022 15:41:51 +0200},
  biburl       = {https://dblp.org/rec/conf/nips/BunelTTKM18.bib},
  bibsource    = {dblp computer science bibliography, https://dblp.org}
}

@inproceedings{LuK20,
  author       = {Jingyue Lu and
                  M. Pawan Kumar},
  title        = {Neural Network Branching for Neural Network Verification},
  booktitle    = {{ICLR}},
  //publisher    = {OpenReview.net},
  year         = {2020},
  //url          = {https://openreview.net/forum?id=B1evfa4tPB},
  timestamp    = {Thu, 07 May 2020 17:11:48 +0200},
  biburl       = {https://dblp.org/rec/conf/iclr/LuK20.bib},
  bibsource    = {dblp computer science bibliography, https://dblp.org}
}

@inproceedings{ZugnerG19attack,
  author       = {Daniel Z{\"{u}}gner and
                  Stephan G{\"{u}}nnemann},
  title        = {Adversarial Attacks on Graph Neural Networks via Meta Learning},
  booktitle    = {{ICLR}},
  //publisher    = {OpenReview.net},
  year         = {2019},
  //url          = {https://openreview.net/forum?id=Bylnx209YX},
  timestamp    = {Thu, 25 Jul 2019 14:25:57 +0200},
  biburl       = {https://dblp.org/rec/conf/iclr/ZugnerG19.bib},
  bibsource    = {dblp computer science bibliography, https://dblp.org}
}

@inproceedings{XuC0CWHL19,
  author       = {Kaidi Xu and
                  Hongge Chen and
                  Sijia Liu and
                  Pin{-}Yu Chen and
                  Tsui{-}Wei Weng and
                  Mingyi Hong and
                  Xue Lin},
  //editor       = {Sarit Kraus},
  title        = {Topology Attack and Defense for Graph Neural Networks: An Optimization Perspective},
  booktitle    = {{IJCAI}},
  pages        = {3961--3967},
  //publisher    = {ijcai.org},
  year         = {2019},
  //url          = {https://doi.org/10.24963/ijcai.2019/550},
  //doi          = {10.24963/IJCAI.2019/550},
  timestamp    = {Tue, 15 Oct 2024 16:43:28 +0200},
  biburl       = {https://dblp.org/rec/conf/ijcai/XuC0CWHL19.bib},
  bibsource    = {dblp computer science bibliography, https://dblp.org}
}

@inproceedings{ChangRXHZC0H20,
  author       = {Heng Chang and
                  Yu Rong and
                  Tingyang Xu and
                  Wenbing Huang and
                  Honglei Zhang and
                  Peng Cui and
                  Wenwu Zhu and
                  Junzhou Huang},
  title        = {A Restricted Black-Box Adversarial Framework Towards Attacking Graph Embedding Models},
  booktitle    = {{AAAI}},
  pages        = {3389--3396},
  //publisher    = {{AAAI} Press},
  year         = {2020},
  //url          = {https://doi.org/10.1609/aaai.v34i04.5741},
  //doi          = {10.1609/AAAI.V34I04.5741},
  timestamp    = {Thu, 30 Jan 2025 17:05:02 +0100},
  biburl       = {https://dblp.org/rec/conf/aaai/ChangRXHZC0H20.bib},
  bibsource    = {dblp computer science bibliography, https://dblp.org}
}

@inproceedings{MuWL0XL21,
  author       = {Jiaming Mu and
                  Binghui Wang and
                  Qi Li and
                  Kun Sun and
                  Mingwei Xu and
                  Zhuotao Liu},
  //editor       = {Yongdae Kim and
                  Jong Kim and
                  Giovanni Vigna and
                  Elaine Shi},
  title        = {A Hard Label Black-box Adversarial Attack Against Graph Neural Networks},
  booktitle    = {{CCS}},
  pages        = {108--125},
  //publisher    = {{ACM}},
  year         = {2021},
  //url          = {https://doi.org/10.1145/3460120.3484796},
  //doi          = {10.1145/3460120.3484796},
  timestamp    = {Mon, 22 Nov 2021 09:23:07 +0100},
  biburl       = {https://dblp.org/rec/conf/ccs/MuWL0XL21.bib},
  bibsource    = {dblp computer science bibliography, https://dblp.org}
}

@article{LiXCXHZ23,
  author       = {Jintang Li and
                  Tao Xie and
                  Liang Chen and
                  Fenfang Xie and
                  Xiangnan He and
                  Zibin Zheng},
  title        = {Adversarial Attack on Large Scale Graph},
  journal      = {{IEEE} Trans. Knowl. Data Eng.},
  volume       = {35},
  number       = {1},
  pages        = {82--95},
  year         = {2023},
  //url          = {https://doi.org/10.1109/TKDE.2021.3078755},
  //doi          = {10.1109/TKDE.2021.3078755},
  timestamp    = {Sun, 15 Jan 2023 18:30:54 +0100},
  biburl       = {https://dblp.org/rec/journals/tkde/LiXCXHZ23.bib},
  bibsource    = {dblp computer science bibliography, https://dblp.org}
}

@inproceedings{ZouZDGKLT21,
  author       = {Xu Zou and
                  Qinkai Zheng and
                  Yuxiao Dong and
                  Xinyu Guan and
                  Evgeny Kharlamov and
                  Jialiang Lu and
                  Jie Tang},
  //editor       = {Feida Zhu and
                  Beng Chin Ooi and
                  Chunyan Miao},
  title        = {{TDGIA}: Effective Injection Attacks on Graph Neural Networks},
  booktitle    = {{KDD}},
  pages        = {2461--2471},
  //publisher    = {{ACM}},
  year         = {2021},
  //url          = {https://doi.org/10.1145/3447548.3467314},
  //doi          = {10.1145/3447548.3467314},
  timestamp    = {Tue, 13 Aug 2024 08:06:44 +0200},
  biburl       = {https://dblp.org/rec/conf/kdd/ZouZDGKLT21.bib},
  bibsource    = {dblp computer science bibliography, https://dblp.org}
}

@inproceedings{ZhangJWG21,
  author       = {Zaixi Zhang and
                  Jinyuan Jia and
                  Binghui Wang and
                  Neil Zhenqiang Gong},
  //editor       = {Jorge Lobo and
                  Roberto Di Pietro and
                  Omar Chowdhury and
                  Hongxin Hu},
  title        = {Backdoor Attacks to Graph Neural Networks},
  booktitle    = {{SACMAT}},
  pages        = {15--26},
  //publisher    = {{ACM}},
  year         = {2021},
  //url          = {https://doi.org/10.1145/3450569.3463560},
  //doi          = {10.1145/3450569.3463560},
  timestamp    = {Wed, 07 Aug 2024 15:41:07 +0200},
  biburl       = {https://dblp.org/rec/conf/sacmat/ZhangJWG21.bib},
  bibsource    = {dblp computer science bibliography, https://dblp.org}
}

@inproceedings{ZhouMWRV19,
  author       = {Kai Zhou and
                  Tomasz P. Michalak and
                  Marcin Waniek and
                  Talal Rahwan and
                  Yevgeniy Vorobeychik},
  //editor       = {Edith Elkind and
                  Manuela Veloso and
                  Noa Agmon and
                  Matthew E. Taylor},
  title        = {Attacking Similarity-Based Link Prediction in Social Networks},
  booktitle    = {{AAMAS}},
  pages        = {305--313},
  //publisher    = {International Foundation for Autonomous Agents and Multiagent Systems},
  year         = {2019},
  //url          = {http://dl.acm.org/citation.cfm?id=3331707},
  timestamp    = {Mon, 22 Jan 2024 12:10:32 +0100},
  biburl       = {https://dblp.org/rec/conf/atal/ZhouMWRV19.bib},
  bibsource    = {dblp computer science bibliography, https://dblp.org}
}

@article{ChenZCDX23,
  author       = {Jinyin Chen and
                  Jian Zhang and
                  Zhi Chen and
                  Min Du and
                  Qi Xuan},
  title        = {Time-Aware Gradient Attack on Dynamic Network Link Prediction},
  journal      = {{IEEE} Trans. Knowl. Data Eng.},
  volume       = {35},
  number       = {2},
  pages        = {2091--2102},
  year         = {2023},
  //url          = {https://doi.org/10.1109/TKDE.2021.3110580},
  //doi          = {10.1109/TKDE.2021.3110580},
  timestamp    = {Tue, 06 Aug 2024 08:09:47 +0200},
  biburl       = {https://dblp.org/rec/journals/tkde/ChenZCDX23.bib},
  bibsource    = {dblp computer science bibliography, https://dblp.org}
}

@inproceedings{Barcelo20,
  author       = {Pablo Barcel{\'{o}} and
                  Egor V. Kostylev and
                  Mika{\"{e}}l Monet and
                  Jorge P{\'{e}}rez and
                  Juan L. Reutter and
                  Juan Pablo Silva},
  title        = {The Logical Expressiveness of Graph Neural Networks},
  booktitle    = {{ICLR}},
  year         = {2020}
}

@inproceedings{BenediktLMT24,
  author       = {Michael Benedikt and
                  Chia{-}Hsuan Lu and
                  Boris Motik and
                  Tony Tan},
  //editor       = {Karl Bringmann and
                  Martin Grohe and
                  Gabriele Puppis and
                  Ola Svensson},
  title        = {Decidability of Graph Neural Networks via Logical Characterizations},
  booktitle    = {{ICALP}},
  //series       = {LIPIcs},
  //volume       = {297},
  pages        = {127:1--127:20},
  //publisher    = {Schloss Dagstuhl - Leibniz-Zentrum f{\"{u}}r Informatik},
  year         = {2024},
  //url          = {https://doi.org/10.4230/LIPIcs.ICALP.2024.127},
  //doi          = {10.4230/LIPICS.ICALP.2024.127},
  timestamp    = {Wed, 21 Aug 2024 22:46:00 +0200},
  biburl       = {https://dblp.org/rec/conf/icalp/BenediktLMT24.bib},
  bibsource    = {dblp computer science bibliography, https://dblp.org}
}

@article{Schonherr25,
  author       = {Moritz Sch{\"{o}}nherr and
                  Carsten Lutz},
  title        = {Logical Characterizations of {GNNs} with Mean Aggregation},
  journal      = {CoRR},
  volume       = {abs/2507.18145},
  year         = {2025},
  //url          = {https://doi.org/10.48550/arXiv.2507.18145},
  //doi          = {10.48550/ARXIV.2507.18145},
  eprinttype    = {arXiv},
  eprint       = {2507.18145},
  timestamp    = {Mon, 18 Aug 2025 22:03:50 +0200},
  biburl       = {https://dblp.org/rec/journals/corr/abs-2507-18145.bib},
  bibsource    = {dblp computer science bibliography, https://dblp.org}
}

@inproceedings{SalzerL23,
  author       = {Marco S{\"{a}}lzer and
                  Martin Lange},
  title        = {Fundamental Limits in Formal Verification of Message-Passing Neural Networks},
  booktitle    = {{ICLR}},
  //publisher    = {OpenReview.net},
  year         = {2023},
  //url          = {https://openreview.net/forum?id=WlbG820mRH-},
  timestamp    = {Wed, 24 Jul 2024 16:50:33 +0200},
  biburl       = {https://dblp.org/rec/conf/iclr/SalzerL23.bib},
  bibsource    = {dblp computer science bibliography, https://dblp.org}
}

@inproceedings{NunnSST24,
  author       = {Pierre Nunn and
                  Marco S{\"{a}}lzer and
                  Fran{\c{c}}ois Schwarzentruber and
                  Nicolas Troquard},
  title        = {A Logic for Reasoning about Aggregate-Combine Graph Neural Networks},
  booktitle    = {{IJCAI}},
  pages        = {3532--3540},
  //publisher    = {ijcai.org},
  year         = {2024},
  //url          = {https://www.ijcai.org/proceedings/2024/391},
  timestamp    = {Fri, 18 Oct 2024 20:53:42 +0200},
  biburl       = {https://dblp.org/rec/conf/ijcai/NunnSST24.bib},
  bibsource    = {dblp computer science bibliography, https://dblp.org}
}

@inproceedings{DaiSZLW19,
  author       = {Quanyu Dai and
                  Xiao Shen and
                  Liang Zhang and
                  Qiang Li and
                  Dan Wang},
  //editor       = {Ling Liu and
                  Ryen W. White and
                  Amin Mantrach and
                  Fabrizio Silvestri and
                  Julian J. McAuley and
                  Ricardo Baeza{-}Yates and
                  Leila Zia},
  title        = {Adversarial Training Methods for Network Embedding},
  booktitle    = {{WWW}},
  pages        = {329--339},
  //publisher    = {{ACM}},
  year         = {2019},
  //url          = {https://doi.org/10.1145/3308558.3313445},
  //doi          = {10.1145/3308558.3313445},
  timestamp    = {Sat, 30 Sep 2023 09:59:30 +0200},
  biburl       = {https://dblp.org/rec/conf/www/DaiSZLW19.bib},
  bibsource    = {dblp computer science bibliography, https://dblp.org}
}

@article{FengHTC21,
  author       = {Fuli Feng and
                  Xiangnan He and
                  Jie Tang and
                  Tat{-}Seng Chua},
  title        = {Graph Adversarial Training: Dynamically Regularizing Based on Graph Structure},
  journal      = {{IEEE} Trans. Knowl. Data Eng.},
  volume       = {33},
  number       = {6},
  pages        = {2493--2504},
  year         = {2021},
  //url          = {https://doi.org/10.1109/TKDE.2019.2957786},
  //doi          = {10.1109/TKDE.2019.2957786},
  timestamp    = {Tue, 01 Jun 2021 08:34:18 +0200},
  biburl       = {https://dblp.org/rec/journals/tkde/FengHTC21.bib},
  bibsource    = {dblp computer science bibliography, https://dblp.org}
}

@article{LiPCZLL23,
  author       = {Jintang Li and
                  Jiaying Peng and
                  Liang Chen and
                  Zibin Zheng and
                  Tingting Liang and
                  Qing Ling},
  title        = {Spectral Adversarial Training for Robust Graph Neural Network},
  journal      = {{IEEE} Trans. Knowl. Data Eng.},
  volume       = {35},
  number       = {9},
  pages        = {9240--9253},
  year         = {2023},
  //url          = {https://doi.org/10.1109/TKDE.2022.3222207},
  //doi          = {10.1109/TKDE.2022.3222207},
  timestamp    = {Mon, 30 Oct 2023 15:33:00 +0100},
  biburl       = {https://dblp.org/rec/journals/tkde/LiPCZLL23.bib},
  bibsource    = {dblp computer science bibliography, https://dblp.org}
}

@inproceedings{GoschG0CZG23,
  author       = {Lukas Gosch and
                  Simon Geisler and
                  Daniel Sturm and
                  Bertrand Charpentier and
                  Daniel Z{\"{u}}gner and
                  Stephan G{\"{u}}nnemann},
  //editor       = {Alice Oh and
                  Tristan Naumann and
                  Amir Globerson and
                  Kate Saenko and
                  Moritz Hardt and
                  Sergey Levine},
  title        = {Adversarial Training for Graph Neural Networks: Pitfalls, Solutions, and New Directions},
  booktitle    = {NeurIPS},
  year         = {2023},
  //url          = {http://papers.nips.cc/paper\_files/paper/2023/hash/b5a801e6bc4f4ffa3e6786518a324488-Abstract-Conference.html},
  timestamp    = {Fri, 01 Mar 2024 16:26:20 +0100},
  biburl       = {https://dblp.org/rec/conf/nips/GoschG0CZG23.bib},
  bibsource    = {dblp computer science bibliography, https://dblp.org}
}

@inproceedings{Wu0TDLZ19,
  author       = {Huijun Wu and
                  Chen Wang and
                  Yuriy Tyshetskiy and
                  Andrew Docherty and
                  Kai Lu and
                  Liming Zhu},
  //editor       = {Sarit Kraus},
  title        = {Adversarial Examples for Graph Data: Deep Insights into Attack and Defense},
  booktitle    = {{IJCAI}},
  pages        = {4816--4823},
  //publisher    = {ijcai.org},
  year         = {2019},
  //url          = {https://doi.org/10.24963/ijcai.2019/669},
  //doi          = {10.24963/IJCAI.2019/669},
  timestamp    = {Tue, 15 Oct 2024 16:43:28 +0200},
  biburl       = {https://dblp.org/rec/conf/ijcai/Wu0TDLZ19.bib},
  bibsource    = {dblp computer science bibliography, https://dblp.org}
}

@inproceedings{ZhangZ20,
  author       = {Xiang Zhang and
                  Marinka Zitnik},
  //editor       = {Hugo Larochelle and
                  Marc'Aurelio Ranzato and
                  Raia Hadsell and
                  Maria{-}Florina Balcan and
                  Hsuan{-}Tien Lin},
  title        = {{GNNGuard}: Defending Graph Neural Networks against Adversarial Attacks},
  booktitle    = {NeurIPS},
  year         = {2020},
  //url          = {https://proceedings.neurips.cc/paper/2020/hash/690d83983a63aa1818423fd6edd3bfdb-Abstract.html},
  timestamp    = {Tue, 09 Apr 2024 11:54:40 +0200},
  biburl       = {https://dblp.org/rec/conf/nips/ZhangZ20.bib},
  bibsource    = {dblp computer science bibliography, https://dblp.org}
}

@inproceedings{WangJCG21,
  author       = {Binghui Wang and
                  Jinyuan Jia and
                  Xiaoyu Cao and
                  Neil Zhenqiang Gong},
  //editor       = {Feida Zhu and
                  Beng Chin Ooi and
                  Chunyan Miao},
  title        = {Certified Robustness of Graph Neural Networks against Adversarial Structural Perturbation},
  booktitle    = {{KDD}},
  pages        = {1645--1653},
  //publisher    = {{ACM}},
  year         = {2021},
  //url          = {https://doi.org/10.1145/3447548.3467295},
  //doi          = {10.1145/3447548.3467295},
  timestamp    = {Wed, 07 Aug 2024 15:41:07 +0200},
  biburl       = {https://dblp.org/rec/conf/kdd/WangJCG21.bib},
  bibsource    = {dblp computer science bibliography, https://dblp.org}
}

@inproceedings{XiaYWJ24,
  author       = {Zaishuo Xia and
                  Han Yang and
                  Binghui Wang and
                  Jinyuan Jia},
  title        = {{GNNCert}: Deterministic Certification of Graph Neural Networks against Adversarial Perturbations},
  booktitle    = {{ICLR}},
  //publisher    = {OpenReview.net},
  year         = {2024},
  //url          = {https://openreview.net/forum?id=IGzaH538fz},
  timestamp    = {Wed, 07 Aug 2024 15:48:30 +0200},
  biburl       = {https://dblp.org/rec/conf/iclr/XiaYWJ24.bib},
  bibsource    = {dblp computer science bibliography, https://dblp.org}
}

@inproceedings{KipfW17,
  author       = {Thomas N. Kipf and
                  Max Welling},
  title        = {Semi-Supervised Classification with Graph Convolutional Networks},
  booktitle    = {{ICLR}},
  //publisher    = {OpenReview.net},
  year         = {2017},
  //url          = {https://openreview.net/forum?id=SJU4ayYgl},
  timestamp    = {Thu, 25 Jul 2019 14:25:55 +0200},
  biburl       = {https://dblp.org/rec/conf/iclr/KipfW17.bib},
  bibsource    = {dblp computer science bibliography, https://dblp.org}
}

@inproceedings{HamiltonYL17,
  author       = {William L. Hamilton and
                  Zhitao Ying and
                  Jure Leskovec},
  //editor       = {Isabelle Guyon and
                  Ulrike von Luxburg and
                  Samy Bengio and
                  Hanna M. Wallach and
                  Rob Fergus and
                  S. V. N. Vishwanathan and
                  Roman Garnett},
  title        = {Inductive Representation Learning on Large Graphs},
  booktitle    = {NeurIPS},
  pages        = {1024--1034},
  year         = {2017},
  //url          = {https://proceedings.neurips.cc/paper/2017/hash/5dd9db5e033da9c6fb5ba83c7a7ebea9-Abstract.html},
  timestamp    = {Thu, 21 Jan 2021 15:15:21 +0100},
  biburl       = {https://dblp.org/rec/conf/nips/HamiltonYL17.bib},
  bibsource    = {dblp computer science bibliography, https://dblp.org}
}

@article{BronsteinBCV21,
  author       = {Michael M. Bronstein and
                  Joan Bruna and
                  Taco Cohen and
                  Petar Velickovic},
  title        = {Geometric Deep Learning: Grids, Groups, Graphs, Geodesics, and Gauges},
  journal      = {CoRR},
  volume       = {abs/2104.13478},
  year         = {2021},
  //url          = {https://arxiv.org/abs/2104.13478},
  eprinttype    = {arXiv},
  eprint       = {2104.13478},
  timestamp    = {Tue, 04 May 2021 15:12:43 +0200},
  biburl       = {https://dblp.org/rec/journals/corr/abs-2104-13478.bib},
  bibsource    = {dblp computer science bibliography, https://dblp.org}
}

@article{FeyL19PyG,
  author       = {Matthias Fey and
                  Jan Eric Lenssen},
  title        = {Fast Graph Representation Learning with {PyTorch} {Geometric}},
  journal      = {CoRR},
  volume       = {abs/1903.02428},
  year         = {2019},
  //url          = {http://arxiv.org/abs/1903.02428},
  eprinttype    = {arXiv},
  eprint       = {1903.02428},
  timestamp    = {Sun, 31 Mar 2019 19:01:24 +0200},
  biburl       = {https://dblp.org/rec/journals/corr/abs-1903-02428.bib},
  bibsource    = {dblp computer science bibliography, https://dblp.org}
}

@inproceedings{HojnyZCM24,
  author       = {Christopher Hojny and
                  Shiqiang Zhang and
                  Juan S. Campos and
                  Ruth Misener},
  title        = {Verifying message-passing neural networks via topology-based bounds tightening},
  booktitle    = {{ICML}},
  //publisher    = {OpenReview.net},
  year         = {2024},
  //url          = {https://openreview.net/forum?id=nAoiUlz4Bf},
  timestamp    = {Mon, 02 Sep 2024 16:55:26 +0200},
  biburl       = {https://dblp.org/rec/conf/icml/HojnyZCM24.bib},
  bibsource    = {dblp computer science bibliography, https://dblp.org}
}

@inproceedings{ScholtenSGBG22,
  author       = {Yan Scholten and
                  Jan Schuchardt and
                  Simon Geisler and
                  Aleksandar Bojchevski and
                  Stephan G{\"{u}}nnemann},
  //editor       = {Sanmi Koyejo and
                  S. Mohamed and
                  A. Agarwal and
                  Danielle Belgrave and
                  K. Cho and
                  A. Oh},
  title        = {Randomized Message-Interception Smoothing: Gray-box Certificates for
                  Graph Neural Networks},
  booktitle    = {NeurIPS},
  year         = {2022},
  //url          = {http://papers.nips.cc/paper\_files/paper/2022/hash/d66d8164cfbf012cac2866edbb375035-Abstract-Conference.html},
  timestamp    = {Mon, 08 Jan 2024 16:31:36 +0100},
  biburl       = {https://dblp.org/rec/conf/nips/ScholtenSGBG22.bib},
  bibsource    = {dblp computer science bibliography, https://dblp.org}
}

@inproceedings{HaghshenasHN23,
  author       = {Seyed Hamed Haghshenas and
                  Abul Hasnat and
                  Mia Naeini},
  title        = {A Temporal Graph Neural Network for Cyber Attack Detection and Localization
                  in Smart Grids},
  booktitle    = {{ISGT}},
  pages        = {1--5},
  //publisher    = {{IEEE}},
  year         = {2023},
  //url          = {https://doi.org/10.1109/ISGT51731.2023.10066446},
  //doi          = {10.1109/ISGT51731.2023.10066446},
  timestamp    = {Tue, 28 Mar 2023 19:49:46 +0200},
  biburl       = {https://dblp.org/rec/conf/isgt/HaghshenasHN23.bib},
  bibsource    = {dblp computer science bibliography, https://dblp.org}
}

@inproceedings{McAuleyL13,
  author       = {Julian J. McAuley and
                  Jure Leskovec},
  //editor       = {Daniel Schwabe and
                  Virg{\'{\i}}lio A. F. Almeida and
                  Hartmut Glaser and
                  Ricardo Baeza{-}Yates and
                  Sue B. Moon},
  title        = {From amateurs to connoisseurs: Modeling the evolution of user expertise through online reviews},
  booktitle    = {{WWW}},
  pages        = {897--908},
  //publisher    = {International World Wide Web Conferences Steering Committee / {ACM}},
  year         = {2013},
  //url          = {https://doi.org/10.1145/2488388.2488466},
  //doi          = {10.1145/2488388.2488466},
  timestamp    = {Thu, 30 Jan 2025 09:01:37 +0100},
  biburl       = {https://dblp.org/rec/conf/www/McAuleyL13.bib},
  bibsource    = {dblp computer science bibliography, https://dblp.org}
}

@inproceedings{RayanaA15,
  author       = {Shebuti Rayana and
                  Leman Akoglu},
  //editor       = {Longbing Cao and
                  Chengqi Zhang and
                  Thorsten Joachims and
                  Geoffrey I. Webb and
                  Dragos D. Margineantu and
                  Graham Williams},
  title        = {Collective Opinion Spam Detection: Bridging Review Networks and Metadata},
  booktitle    = {{KDD}},
  pages        = {985--994},
  //publisher    = {{ACM}},
  year         = {2015},
  //url          = {https://doi.org/10.1145/2783258.2783370},
  //doi          = {10.1145/2783258.2783370},
  timestamp    = {Tue, 06 Nov 2018 16:59:36 +0100},
  biburl       = {https://dblp.org/rec/conf/kdd/RayanaA15.bib},
  bibsource    = {dblp computer science bibliography, https://dblp.org}
}

@article{SenNBGGE08,
  author       = {Prithviraj Sen and
                  Galileo Namata and
                  Mustafa Bilgic and
                  Lise Getoor and
                  Brian Gallagher and
                  Tina Eliassi{-}Rad},
  title        = {Collective Classification in Network Data},
  journal      = {{AI} Mag.},
  volume       = {29},
  number       = {3},
  pages        = {93--106},
  year         = {2008},
  //url          = {https://doi.org/10.1609/aimag.v29i3.2157},
  //doi          = {10.1609/AIMAG.V29I3.2157},
  timestamp    = {Tue, 16 Aug 2022 23:09:49 +0200},
  biburl       = {https://dblp.org/rec/journals/aim/SenNBGGE08.bib},
  bibsource    = {dblp computer science bibliography, https://dblp.org}
}

@inproceedings{BonaertDBV21,
  author       = {Gregory Bonaert and
                  Dimitar I. Dimitrov and
                  Maximilian Baader and
                  Martin T. Vechev},
  //editor       = {Stephen N. Freund and
                  Eran Yahav},
  title        = {Fast and precise certification of transformers},
  booktitle    = {{PLDI}},
  pages        = {466--481},
  //publisher    = {{ACM}},
  year         = {2021},
  //url          = {https://doi.org/10.1145/3453483.3454056},
  //doi          = {10.1145/3453483.3454056},
  timestamp    = {Mon, 03 Mar 2025 21:20:02 +0100},
  biburl       = {https://dblp.org/rec/conf/pldi/BonaertDBV21.bib},
  bibsource    = {dblp computer science bibliography, https://dblp.org}
}

@book{rossi2006handbook,
  title={Handbook of Constraint Programming},
  author={Rossi, Francesca and Van Beek, Peter and Walsh, Toby},
  year={2006},
  publisher={Elsevier}
}

@inproceedings{LaiZPZ24,
  author       = {Yuni Lai and
                  Yulin Zhu and
                  Bailin Pan and
                  Kai Zhou},
  title        = {Node-aware Bi-smoothing: Certified Robustness against Graph Injection Attacks},
  booktitle    = {{IEEE} {SP}},
  pages        = {2958--2976},
  //publisher    = {{IEEE}},
  year         = {2024},
  //url          = {https://doi.org/10.1109/SP54263.2024.00241},
  //doi          = {10.1109/SP54263.2024.00241},
  timestamp    = {Sat, 21 Sep 2024 14:24:56 +0200},
  biburl       = {https://dblp.org/rec/conf/sp/LaiZPZ24.bib},
  bibsource    = {dblp computer science bibliography, https://dblp.org}
}

@article{LadnerEA25,
  author       = {Tobias Ladner and
                  Michael Eichelbeck and
                  Matthias Althoff},
  title        = {Formal Verification of Graph Convolutional Networks with Uncertain
                  Node Features and Uncertain Graph Structure},
  journal      = {Trans. Mach. Learn. Res.},
  volume       = {2025},
  year         = {2025},
  //url          = {https://openreview.net/forum?id=B6y12Ot0cP},
  timestamp    = {Mon, 23 Jun 2025 16:40:29 +0200},
  biburl       = {https://dblp.org/rec/journals/tmlr/LadnerEA25.bib},
  bibsource    = {dblp computer science bibliography, https://dblp.org}
}

@inproceedings{SalmanY0HZ19,
  author       = {Hadi Salman and
                  Greg Yang and
                  Huan Zhang and
                  Cho{-}Jui Hsieh and
                  Pengchuan Zhang},
  //editor       = {Hanna M. Wallach and
                  Hugo Larochelle and
                  Alina Beygelzimer and
                  Florence d'Alch{\'{e}}{-}Buc and
                  Emily B. Fox and
                  Roman Garnett},
  title        = {A Convex Relaxation Barrier to Tight Robustness Verification of Neural Networks},
  booktitle    = {NeurIPS},
  pages        = {9832--9842},
  year         = {2019},
  //url          = {https://proceedings.neurips.cc/paper/2019/hash/246a3c5544feb054f3ea718f61adfa16-Abstract.html},
  timestamp    = {Mon, 16 May 2022 15:41:51 +0200},
  biburl       = {https://dblp.org/rec/conf/nips/SalmanY0HZ19.bib},
  bibsource    = {dblp computer science bibliography, https://dblp.org}
}

@inproceedings{XuS0WCHKLH20,
  author       = {Kaidi Xu and
                  Zhouxing Shi and
                  Huan Zhang and
                  Yihan Wang and
                  Kai{-}Wei Chang and
                  Minlie Huang and
                  Bhavya Kailkhura and
                  Xue Lin and
                  Cho{-}Jui Hsieh},
  //editor       = {Hugo Larochelle and
                  Marc'Aurelio Ranzato and
                  Raia Hadsell and
                  Maria{-}Florina Balcan and
                  Hsuan{-}Tien Lin},
  title        = {Automatic Perturbation Analysis for Scalable Certified Robustness and Beyond},
  booktitle    = {NeurIPS},
  year         = {2020},
  //url          = {https://proceedings.neurips.cc/paper/2020/hash/0cbc5671ae26f67871cb914d81ef8fc1-Abstract.html},
  timestamp    = {Tue, 11 Mar 2025 09:17:48 +0100},
  biburl       = {https://dblp.org/rec/conf/nips/XuS0WCHKLH20.bib},
  bibsource    = {dblp computer science bibliography, https://dblp.org}
}

@inproceedings{ZhangCFWSMTM23,
  author       = {Shiqiang Zhang and
                  Juan S. Campos and
                  Christian Feldmann and
                  David Walz and
                  Frederik Sandfort and
                  Miriam Mathea and
                  Calvin Tsay and
                  Ruth Misener},
  //editor       = {Alice Oh and
                  Tristan Naumann and
                  Amir Globerson and
                  Kate Saenko and
                  Moritz Hardt and
                  Sergey Levine},
  title        = {Optimizing over trained {GNNs} via symmetry breaking},
  booktitle    = {NeurIPS},
  year         = {2023},
  //url          = {http://papers.nips.cc/paper\_files/paper/2023/hash/8c8cd1b78cdae751265c88efc136e5bd-Abstract-Conference.html},
  timestamp    = {Fri, 01 Mar 2024 16:26:20 +0100},
  biburl       = {https://dblp.org/rec/conf/nips/ZhangCFWSMTM23.bib},
  bibsource    = {dblp computer science bibliography, https://dblp.org}
}

@inproceedings{YingHCEHL18,
  author       = {Rex Ying and
                  Ruining He and
                  Kaifeng Chen and
                  Pong Eksombatchai and
                  William L. Hamilton and
                  Jure Leskovec},
  //editor       = {Yike Guo and
                  Faisal Farooq},
  title        = {Graph Convolutional Neural Networks for Web-Scale Recommender Systems},
  booktitle    = {{KDD}},
  pages        = {974--983},
  //publisher    = {{ACM}},
  year         = {2018},
  //url          = {https://doi.org/10.1145/3219819.3219890},
  //doi          = {10.1145/3219819.3219890},
  timestamp    = {Sat, 09 Apr 2022 12:35:09 +0200},
  biburl       = {https://dblp.org/rec/conf/kdd/YingHCEHL18.bib},
  bibsource    = {dblp computer science bibliography, https://dblp.org}
}

@inproceedings{DehmamyBY19,
  author       = {Nima Dehmamy and
                  Albert{-}L{\'{a}}szl{\'{o}} Barab{\'{a}}si and
                  Rose Yu},
  //editor       = {Hanna M. Wallach and
                  Hugo Larochelle and
                  Alina Beygelzimer and
                  Florence d'Alch{\'{e}}{-}Buc and
                  Emily B. Fox and
                  Roman Garnett},
  title        = {Understanding the Representation Power of Graph Neural Networks in
                  Learning Graph Topology},
  booktitle    = {NeurIPS},
  pages        = {15387--15397},
  year         = {2019},
  //url          = {https://proceedings.neurips.cc/paper/2019/hash/73bf6c41e241e28b89d0fb9e0c82f9ce-Abstract.html},
  timestamp    = {Mon, 16 May 2022 15:41:51 +0200},
  biburl       = {https://dblp.org/rec/conf/nips/DehmamyBY19.bib},
  bibsource    = {dblp computer science bibliography, https://dblp.org}
}

@inproceedings{CorsoCBLV20,
  author       = {Gabriele Corso and
                  Luca Cavalleri and
                  Dominique Beaini and
                  Pietro Li{\`{o}} and
                  Petar Velickovic},
  //editor       = {Hugo Larochelle and
                  Marc'Aurelio Ranzato and
                  Raia Hadsell and
                  Maria{-}Florina Balcan and
                  Hsuan{-}Tien Lin},
  title        = {Principal Neighbourhood Aggregation for Graph Nets},
  booktitle    = {NeurIPS},
  year         = {2020},
  //url          = {https://proceedings.neurips.cc/paper/2020/hash/99cad265a1768cc2dd013f0e740300ae-Abstract.html},
  timestamp    = {Tue, 19 Jan 2021 15:57:42 +0100},
  biburl       = {https://dblp.org/rec/conf/nips/CorsoCBLV20.bib},
  bibsource    = {dblp computer science bibliography, https://dblp.org}
}

@inproceedings{RosenbluthTG23,
  author       = {Eran Rosenbluth and
                  Jan T{\"{o}}nshoff and
                  Martin Grohe},
  title        = {Some Might Say All You Need Is Sum},
  booktitle    = {{IJCAI}},
  pages        = {4172--4179},
  //publisher    = {ijcai.org},
  year         = {2023},
  //url          = {https://doi.org/10.24963/ijcai.2023/464},
  //doi          = {10.24963/IJCAI.2023/464},
  timestamp    = {Tue, 15 Oct 2024 16:43:28 +0200},
  biburl       = {https://dblp.org/rec/conf/ijcai/RosenbluthTG23.bib},
  bibsource    = {dblp computer science bibliography, https://dblp.org}
}

@inproceedings{geisler2021robustness,
  title={Robustness of Graph Neural Networks at Scale},
  author={Geisler, Simon and
          Schmidt, Tobias and
          {\c{S}}irin, Hakan and
          Z{\"u}gner, Daniel and
          Bojchevski, Aleksandar and
          G{\"u}nnemann, Stephan},
  booktitle={NeurIPS},
  //volume={34},
  pages={7637--7649},
  year={2021}
}

@inproceedings{morris2020tudataset,
  title={{TUDataset}: A collection of benchmark datasets for learning with graphs},
  author={Morris, Christopher and Kriege, Nils Morten and Bause, Franka and Kersting, Kristian and Mutzel, Petra and Neumann, Marion},
  booktitle={ICML 2020 Workshop on Graph Representation Learning and Beyond},
  year={2020}
}

@article{lu2025roblight,
  author       = {Chia{-}Hsuan Lu and
                  Tony Tan and
                  Michael Benedikt},
  title        = {Robustness Verification of Graph Neural Networks Via Lightweight Satisfiability
                  Testing},
  journal      = {CoRR},
  volume       = {abs/2510.18591},
  year         = {2025},
  //url          = {https://doi.org/10.48550/arXiv.2510.18591},
  //doi          = {10.48550/ARXIV.2510.18591},
  eprinttype    = {arXiv},
  eprint       = {2510.18591},
  timestamp    = {Sat, 15 Nov 2025 15:31:50 +0100},
  biburl       = {https://dblp.org/rec/journals/corr/abs-2510-18591.bib},
  bibsource    = {dblp computer science bibliography, https://dblp.org}
}

@inproceedings{xu2019gin,
  author       = {Keyulu Xu and
                  Weihua Hu and
                  Jure Leskovec and
                  Stefanie Jegelka},
  title        = {How Powerful are Graph Neural Networks?},
  booktitle    = {{ICLR}},
  //publisher    = {OpenReview.net},
  year         = {2019},
  //url          = {https://openreview.net/forum?id=ryGs6iA5Km},
  timestamp    = {Thu, 25 Jul 2019 13:03:15 +0200},
  biburl       = {https://dblp.org/rec/conf/iclr/XuHLJ19.bib},
  bibsource    = {dblp computer science bibliography, https://dblp.org}
}

@inproceedings{CheungGS17,
  author       = {Kevin K. H. Cheung and
                  Ambros M. Gleixner and
                  Daniel E. Steffy},
  //editor       = {Friedrich Eisenbrand and
                  Jochen K{\"{o}}nemann},
  title        = {Verifying Integer Programming Results},
  booktitle    = {{IPCO}},
  //series       = {Lecture Notes in Computer Science},
  //volume       = {10328},
  pages        = {148--160},
  //publisher    = {Springer},
  year         = {2017},
  //url          = {https://doi.org/10.1007/978-3-319-59250-3\_13},
  //doi          = {10.1007/978-3-319-59250-3\_13},
  timestamp    = {Tue, 14 May 2019 10:00:50 +0200},
  biburl       = {https://dblp.org/rec/conf/ipco/CheungGS17.bib},
  bibsource    = {dblp computer science bibliography, https://dblp.org}
}

@article{BarrettHS26,
  author       = {Clark W. Barrett and
                  Thomas A. Henzinger and
                  Sanjit A. Seshia},
  title        = {Certificates in {AI}: Learn but Verify},
  journal      = {Commun. {ACM}},
  volume       = {69},
  number       = {1},
  pages        = {66--75},
  year         = {2026},
  //url          = {https://doi.org/10.1145/3737447},
  //doi          = {10.1145/3737447},
  timestamp    = {Tue, 03 Feb 2026 08:25:15 +0100},
  biburl       = {https://dblp.org/rec/journals/cacm/BarrettHS26.bib},
  bibsource    = {dblp computer science bibliography, https://dblp.org}
}

@inproceedings{CohenRK19,
  author       = {Jeremy Cohen and
                  Elan Rosenfeld and
                  J. Zico Kolter},
  //editor       = {Kamalika Chaudhuri and
                  Ruslan Salakhutdinov},
  title        = {Certified Adversarial Robustness via Randomized Smoothing},
  booktitle    = {{ICML}},
  //series       = {Proceedings of Machine Learning Research},
  //volume       = {97},
  pages        = {1310--1320},
  //publisher    = {{PMLR}},
  year         = {2019},
  //url          = {http://proceedings.mlr.press/v97/cohen19c.html},
  timestamp    = {Mon, 07 Aug 2023 17:37:03 +0200},
  biburl       = {https://dblp.org/rec/conf/icml/CohenRK19.bib},
  bibsource    = {dblp computer science bibliography, https://dblp.org}
}

@inproceedings{WengCNSBOD19,
  author       = {Lily Weng and
                  Pin{-}Yu Chen and
                  Lam M. Nguyen and
                  Mark S. Squillante and
                  Akhilan Boopathy and
                  Ivan V. Oseledets and
                  Luca Daniel},
  //editor       = {Kamalika Chaudhuri and
                  Ruslan Salakhutdinov},
  title        = {{PROVEN}: Verifying Robustness of Neural Networks with a Probabilistic
                  Approach},
  booktitle    = {{ICML}},
  //series       = {Proceedings of Machine Learning Research},
  //volume       = {97},
  pages        = {6727--6736},
  //publisher    = {{PMLR}},
  year         = {2019},
  //url          = {http://proceedings.mlr.press/v97/weng19a.html},
  timestamp    = {Tue, 11 Jun 2019 15:37:38 +0200},
  biburl       = {https://dblp.org/rec/conf/icml/WengCNSBOD19.bib},
  bibsource    = {dblp computer science bibliography, https://dblp.org}
}

@article{MarzariCF25,
  author       = {Luca Marzari and
                  Ferdinando Cicalese and
                  Alessandro Farinelli},
  title        = {Probabilistically Tightened Linear Relaxation-based Perturbation Analysis
                  for Neural Network Verification},
  journal      = {J. Artif. Intell. Res.},
  volume       = {84},
  year         = {2025},
  //url          = {https://doi.org/10.1613/jair.1.20808},
  //doi          = {10.1613/JAIR.1.20808},
  timestamp    = {Sun, 01 Feb 2026 13:40:39 +0100},
  biburl       = {https://dblp.org/rec/journals/jair/MarzariCF25.bib},
  bibsource    = {dblp computer science bibliography, https://dblp.org}
}

@inproceedings{huang2026parameterized,
  title={Parameterized Abstract Interpretation for Transformer Verification},
  author={Huang, Pei and Wei, Dennis and Isac, Omri and Wu, Haoze and Wu, Min and Barrett, Clark},
  booktitle={AAAI},
  year={2026}
}

@inproceedings{MohammadinejadP21,
  author       = {Sara Mohammadinejad and
                  Brandon Paulsen and
                  Jyotirmoy V. Deshmukh and
                  Chao Wang},
  //editor       = {Catalin Dima and
                  Mahsa Shirmohammadi},
  title        = {{DiffRNN}: Differential Verification of Recurrent Neural Networks},
  booktitle    = {{FORMATS}},
  //series       = {Lecture Notes in Computer Science},
  //volume       = {12860},
  pages        = {117--134},
  //publisher    = {Springer},
  year         = {2021},
  //url          = {https://doi.org/10.1007/978-3-030-85037-1\_8},
  //doi          = {10.1007/978-3-030-85037-1\_8},
  timestamp    = {Tue, 17 Aug 2021 21:08:56 +0200},
  biburl       = {https://dblp.org/rec/conf/formats/MohammadinejadP21.bib},
  bibsource    = {dblp computer science bibliography, https://dblp.org}
}

@article{BanerjeeXS24,
  author       = {Debangshu Banerjee and
                  Changming Xu and
                  Gagandeep Singh},
  title        = {Input-Relational Verification of Deep Neural Networks},
  journal      = {Proc. {ACM} Program. Lang.},
  volume       = {8},
  number       = {{PLDI}},
  pages        = {1--27},
  year         = {2024},
  //url          = {https://doi.org/10.1145/3656377},
  //doi          = {10.1145/3656377},
  timestamp    = {Wed, 14 May 2025 16:00:49 +0200},
  biburl       = {https://dblp.org/rec/journals/pacmpl/BanerjeeXS24.bib},
  bibsource    = {dblp computer science bibliography, https://dblp.org}
}

\newpage
\appendix

\section{Variant for Graph Classification}
\label{appx:defgraphclass}

For graph-classification GNNs, the process to compute the embedding vectors is the same as for node-classification GNNs, as shown in Eq.~(\ref{eq:gnn}) in the main paper.
The difference arises from how the predicted class is obtained, namely for graph classification we have:
\begin{equation}
    \hat{c}_G = \argmax_{c \in C}{\,\left(\text{Softmax}\left( {\mathbf{W}_3 \cdot \sum_{v \in V}{\mathbf{h}_v^{(K)}}} + \mathbf{b}_2 \right) \left[c\right]\right)},
    \label{eq:gc_readout}
\end{equation}
where $\mathbf{W}_3$ is a learnable parameter matrix with dimension $m \times d_K$ and $\mathbf{b}_2$ is a learnable parameter vector.

A graph-classification GNN $f$ is called adversarially robust if and only if, for every perturbed graph $\tilde{G} \in \mathcal{Q}(G)$, it holds that $f(\tilde{G}) = \hat{c}_G$.

Our method adapts to graph classification by simply making slight modifications to the last part of the CSP encoding to align with the operation in Eq.~(\ref{eq:gc_readout}), which includes a summation of all the node embeddings and a linear transformation.
We remark that our bound tightening strategies are compatible with the modified encoding.
Moreover, our incremental solving-based verification can also be used with minor modification as shown in Algorithm~\ref{alg:incgnnverify}.
Although graph classification requires encoding of all the nodes in the graph, resulting in the same number of nodes in each GNN layer, our algorithm only encodes variables associated with the $(K-k)$-th to $K$-th GNN layers in $k$-th iteration, which can also reduces encoding costs and enhances efficiency.

\section{Greedy algorithm for mean aggregation function}
\label{appx:greedy}

We focus on the upper bounds and non-corner cases;
the lower bounds can be handled analogously.
Recall that for the mean aggregation function,
\begin{equation*}
    \overline{z}  =
    \max_{\substack{s_2 + s_3 \le s\\ 0 \le s_2 \le |X_2|\\ 0 \le s_3 \le |X_3|}}
    \overline{z}_{s_2, s_3},
\end{equation*}
where
\begin{equation*}
    \begin{aligned}
        \overline{z}_{s_2, s_3}
        \ =\ &
        \overline{w}_{s_2, s_3} / n_{s_2, s_3}, \\
        n_{s_2, s_3}
        \ =\ &
        |X_1| + |X_2| - s_2 + s_3, \\
        \overline{w}_{s_2, s_3}
        \ =\ &
        \sum_{x \in X_1} \overline{x} + 
        \sum_{1 \le i \le |X_2| - s_2} hi(X_2, i) +
        \sum_{1 \le i \le s_3} hi(X_3, i).
    \end{aligned}
\end{equation*}

For every $0 \le i \le |X_2|$ and $0 \le j \le |X_3|$,
$\overline{z}_{i, j + 1}$ can be rewritten as the weighted average of $\overline{z}_{i, j}$ and $hi(X_3, j+1)$,
\begin{equation*}
    \overline{z}_{i, j + 1}
    \ =\ 
    \frac{\left(|X_1| + |X_2| - i + j\right) \cdot \overline{z}_{i, j} + hi(X_3, j+1)}{|X_1| + |X_2| - i + j+1}.
\end{equation*}

We first prove that the sequence $\overline{z}_{i, 0}, \overline{z}_{i, 1}, \ldots, \overline{z}_{i, |X_3|}$
is \emph{unimodal with a possible plateau}.
It is sufficient to prove that the sequence satisfies the following two properties:

\medskip

    \begin{enumerate}
        \item For every $0 \le j \le |X_3| - 2$,
        if $\overline{z}_{i, j + 1} \le \overline{z}_{i, j}$,
        then $\overline{z}_{i, j + 2} \le \overline{z}_{i, j + 1}$.
        \item For every $0 \le j \le |X_3| - 2$,
        if $\overline{z}_{i, j + 1} < \overline{z}_{i, j}$,
        then $\overline{z}_{i, j + 2} < \overline{z}_{i, j + 1}$.
    \end{enumerate}

\begin{proof}
    We prove the two properties as follows:
    \begin{enumerate}
        \item By the property of weighted averages, we have
        \begin{equation*}
            hi(X_3, j + 1)\ \le\ \overline{z}_{i, j + 1}\ \le\ \overline{z}_{i, j}.
        \end{equation*}
        By definition of $hi(X_3, i)$, it follows that
        \begin{equation*}
            hi(X_3, j + 2)\ \le\ hi(X_3, j + 1).
        \end{equation*}
        Combining both inequalities gives
        \begin{equation*}
            hi(X_3, j + 2)
            \ \le\ 
            \overline{z}_{i, j + 1}.
        \end{equation*}
        The desired property follows since
        $\overline{z}_{i, j + 2}$ is the weighted average of $\overline{z}_{i, j + 1}$ and $hi(X_3, j + 2)$:
        \begin{equation*}
            hi(X_3, j + 2) \le \overline{z}_{i, j + 2} \le \overline{z}_{i, j + 1}.    
        \end{equation*}

        \item Since $\overline{z}_{i, j + 1} \neq \overline{z}_{i, j}$,
        it follows that $\overline{z}_{i, j + 1} \neq hi(X_3, j + 1)$.
        Therefore, the inequality between
        $hi(X_3, j + 1)$ and $\overline{z}_{i, j + 1}$ is strict,
        and
        the desired property follows by the same reasoning as above. \qed
    \end{enumerate}
\end{proof}
Next, note that $\overline{z}$ can be rewritten as
\begin{equation*}
    \overline{z}
    \ =\ 
    \max_{0 \le s_2 \le |X_2|}
    \left(
    \max_{0 \le s_3 \le \max\left(s - s_3, |X_3|\right)}
    \overline{z}_{s_2, s_3}
    \right).
\end{equation*}
Since the inner sequence is unimodal with a possible plateau,
the inner maximum can be found using ternary search in $O(\log s)$ time.
Hence, the total time complexity to compute $\overline{z}$ is $O(s \log s)$.

\section{Proof of the Tightness of Bounds}
\label{appx:tight_bounds}

Recall the problem: 

\medskip

\noindent
Given three finite sets of variables $X_1$, $X_2$, and $X_3$,
the upper and lower bounds for each variable,
and a non-negative integer $s$,
compute the upper and lower bounds for the variable
\begin{equation*}
    z := \mathbf{aggr}(X_1 \cup X'_2 \cup X'_3),
\end{equation*}
where 
$\mathbf{aggr} \in \left\{ \text{sum}, \text{max}, \text{mean} \right\}$,
$X_2' \subseteq X_2$, $X_3' \subseteq X_3$, and $|X_2 \backslash X_2'| + |X_3'| \le s$.

\subsubsection{Max aggregation.}
In this case,
the problem is straightforward through case-by-case analysis. The only subtlety is when $X_1$ is empty and $s \ge |X_2|$, in which case we may have $X'_2 = X'_3 = \emptyset$ and thus $X_1 \cup X'_2 \cup X'_3 = \emptyset$,
which implies that the corresponding upper and lower bounds are $0$.
Therefore, we must take 0 into consideration.

\subsubsection{Mean aggregation.}
In this case,
recall that we reduce the original problem to a combination of subproblems.
For non-negative integers $s_2$ and $s_3$, define the variable
\begin{equation*}
    z_{s_2, s_3} := \mathbf{mean}(X_1 \cup X'_2 \cup X'_3),
\end{equation*}
where 
$X_2' \subseteq X_2$,
$X_3' \subseteq X_3$,
$|X_2 \backslash X_2'| = s_2$,
and $|X_3'| = s_3$.
The upper and lower bounds of the original problem $z$ are given by
\begin{equation}
\label{eq:mean1}
    \begin{aligned}
        \overline{z}  =
        &\max_{\substack{s_2 + s_3 \le s\\ 0 \le s_2 \le |X_2|\\ 0 \le s_3 \le |X_3|}}
        \overline{z}_{s_2, s_3}
        \quad \text{and}\quad
        \underline{z} =
        &\min_{\substack{s_2 + s_3 \le s\\ 0 \le s_2 \le |X_2|\\ 0 \le s_3 \le |X_3|}}
        \underline{z}_{s_2, s_3}.        
    \end{aligned}
\end{equation}
Note that Eq.~(\ref{eq:mean1}) covers all possible allocations of budgets $(s_2, s_3)$ such that $s_2 + s_3 \le s$.
The correctness of this reduction is obvious.

Next, we show that the upper and lower bounds of the variable $z_{s_2, s_3}$ can be computed efficiently.
Here we focus on the upper bound;
a similar argument applies to the lower bound.

Let $(s_2, s_3)$ be a fixed pair of non-negative integers satisfying $s_2 + s_3 \le s$,
$0 \le s_2 \le |X_2|$, and $0 \le s_3 \le |X_3|$.
Let 
\begin{equation*}
    \begin{aligned}
        n_{s_2, s_3}
        \ :=\ &|X_1| + |X'_2| + |X'_3| \\
        \ =\ &|X_1| + |X_2| - s_2 + s_3.
    \end{aligned}
\end{equation*}
Note that
\begin{equation*}
    \begin{aligned}
        z_{s_2, s_3}
        \ =\ &
        \mathbf{mean}(X_1 \cup X'_2 \cup X'_3) \\
        \ =\ &
        \frac{1}{n_{s_2, s_3}}\cdot \mathbf{sum}(X_1 \cup X'_2 \cup X'_3) \\
        \ =\ &
        \frac{1}{n_{s_2, s_3}}\cdot \left(\mathbf{sum}(X_1) + \mathbf{sum}(X'_2) + \mathbf{sum}(X'_3)\right).
    \end{aligned}
\end{equation*}
Because the values of $s_2$ and $s_3$ are fixed,
the selections of $X'_2 \subseteq X_2$ and $X'_3 \subseteq X_3$ are independent.
Thus, the upper bound of $z_{s_2, s_3}$ is given by the sum of upper bounds of $\mathbf{sum}(X_1)$, $\mathbf{sum}(X'_2)$, and $\mathbf{sum}(X'_3)$, divided by $n_{s_2, s_3}$ where 
$X_2' \subseteq X_2$,
$X_3' \subseteq X_3$,
$|X_2 \backslash X_2'| = s_2$,
and $|X_3'| = s_3$.

Observe that, for a set of variables $Y$,
because the sum function is monotonic w.r.t. each variable,
the upper bound of the sum of the set $Y$
is given by the sum of the upper bounds of the variables in $Y$.
Therefore, we have the following conclusions:
\begin{itemize}
    \item The upper bound of $\mathbf{sum}(X_1)$
    is $\sum_{x \in X_1}\overline{x}$.
    \item For $\mathbf{sum}(X'_2)$
    where $X_2' \subseteq X_2$ and $|X_2 \backslash X_2'| = s_2$.
    Since $|X'_2| = |X_2| - s_2$, the upper bound of $\mathbf{sum}(X'_2)$ is given by the sum of the $(|X_2| - s_2)$-largest upper bounds in $X_2$, that is, $\sum_{1 \le i \le |X_2| - s_2} hi(X_2, i)$.
    \item Similarly, since $X'_3 \subseteq X_3$ and $|X'_3| = s_3$, 
    the upper bound of $\mathbf{sum}(X'_3)$ is $\sum_{1 \le i \le s_3} hi(X_3, i)$.
\end{itemize}
Thus $\overline{z}_{s_2, s_3}$ is upper bounded by
\begin{equation*}
    \frac{1}{n_{s_2, s_3}}
    \left(
    \sum\limits_{x \in X_1} \overline{x} + 
    \sum\limits_{1 \le i \le |X_2| - s_2} hi(X_2, i) +
    \sum\limits_{1 \le i \le s_3} hi(X_3, i)
    \right).
\end{equation*}

Note that the bound is tight.
That is, there exist sets $X_2'\subseteq X_2$ and $X_3'\subseteq X_3$ with $|X_2 \backslash X_2'| = s_2$ and $|X_3'| = s_3$ that achieve this bound exactly.

\section{Proof of Theorem 2}
\label{appx:incremental}
Recall the theorem:

\medskip

\noindent
Given a GNN $f$,
an attributed directed graph $G$, and a perturbation space $\mathcal{Q}(G)$,
$f$ is adversarially robust {(for a node $t$, if it is a node-classification GNN)} if and only if Algorithm~\ref{alg:incgnnverify} returns \texttt{robust}.

\medskip

\begin{proof}
    For $1 \le k \le K$, let $\Phi_k := \varphi_{obj} \land \bigwedge_{1 \le i \le k} \varphi_i$.
    Recall that $\varphi_k$ is the encoding of the computation of the $k$-th layer of the GNN $f$.
    Following the intuition described in Section~\ref{sec:encoding},
    it is straightforward to show that $f$ is adversarially robust for $t$ 
    if and only if $\Phi_K$ is unsatisfiable.
    
    If Algorithm~\ref{alg:incgnnverify} returns \texttt{nonrobust},
    then $\Phi_K$ is satisfiable.
    Hence, $f$ is not adversarially robust (for node $t$).
    
    On the other hand, observe that for $1 \le k \le K$,
    \begin{equation*}
        \Phi_K = \Phi_k  \land \bigwedge_{k < i \le K} \varphi_i.
    \end{equation*}
    If $f$ is not adversarially robust (for node $t$),
    then $\Phi_K$ is satisfiable,
    which implies that $\Phi_k$ is also satisfiable for $1 \le k \le K$.
    Therefore, Algorithm~\ref{alg:incgnnverify} returns \texttt{nonrobust}.
    \qed
\end{proof}

\section{Model Implementation and Training}
\label{appx:training}

\subsubsection{Implementation.}
To evaluate the performance of verifiers, we implemented a batch of GraphSAGE models built using PyTorch and the SAGEConv module of PyTorch Geometric.
The aggregation function of SAGEConv was configured as {sum}, {max}, and {mean}, respectively, to build distinct models.
The models consist of 3 SAGEConv layers (i.e., $K = 3$), with input dimension being the dimension of attributes, all hidden layers having a dimension of 32, and output dimension being the number of classes.
For graph-classification GNNs, the final linear layer also has an input dimension of 32.
Except for the last layer, the output of each layer is passed through a ReLU activation function. Finally, a Softmax function is applied to the output of the last layer to obtain predictions for each class.

\subsubsection{Datasets.}
We trained models on 4 node classification datasets and 2 graph classification datasets.
Cora and CiteSeer are two standard graph benchmarks that have been employed for evaluating verification algorithms in a number of previous works \cite{BojchevskiG19,WangJCG21,ScholtenSGBG22,HojnyZCM24}.
Amazon and Yelp are two Internet fraud datasets, and we employed them to evaluate the performance of our algorithm on real-world high-stakes tasks.
Because Amazon and Yelp are heterogeneous graphs with three types of edges, we used only the U-P-U edges for Amazon and R-U-R edges for Yelp to ensure compatibility with our GNN architectures.
MUTAG and ENZYMES are two biochemical datasets to predict whether specific properties are present based on the molecular structure.
Detailed information of these datasets is shown in \tablename~\ref{tab:graphinfo}.

\begin{table}[thpb]
    \centering
    \caption{Statistical information of datasets. For graph-classification datasets, \#Nodes and \#Edges represent the average number of nodes and edges per graph, respectively.}
    \label{tab:graphinfo}
    \setlength{\tabcolsep}{5pt}
    \begin{tabular}{lrrrrr}
        \toprule
        Dataset & \#Graphs & \#Nodes & \#Edges & \#Attributes & \#Classes \\
        \midrule
        Cora & 1 & 2,708 & 5,429 & 1,433 & 7 \\
        CiteSeer & 1 & 3,312 & 4,715 & 3,703 & 6 \\
        Amazon & 1 & 11,944 & 351,216 & 25 & 2 \\
        Yelp & 1 & 45,954 & 98,630 & 32 & 2 \\
        \midrule
        MUTAG & 188 & 17.9 & 39.6 & 7 & 2 \\
        ENZYMES & 600 & 32.6 & 124.3 & 3 & 6\\
        \bottomrule
    \end{tabular}
\end{table}

\subsubsection{Training.}
For Cora and Citeseer, we randomly selected 30\% of the nodes as the training set, 20\% as the validation set, and the remaining 50\% as the test set.
For Amazon and Yelp, given the highly imbalanced distribution of the two classes, we randomly selected 350 (from Amazon) and 2,000 (from Yelp) nodes per class as the training set, 70 (from Amazon) and 400 (from Yelp) nodes per class as the validation set, and the remaining nodes as the test set.
For MUTAG and ENZYMES, we randomly selected
80\% of the graphs as the training set due to the small size of the dataset, 10\% as the validation set, and the remaining 10\% as the test set.
All models were trained for 400 epochs with learning rate 0.001 and weight decay $5 \times 10^{-5}$, and early stopping is applied if the validation loss exceeded the average of the last 10 epochs after training for more than 200 epochs.
The models have been properly trained, and the accuracy on test sets is summarised in \tablename~\ref{tab:trainingacc}.

\begin{table}[htbp]
    \centering
    \caption{Accuracy of the trained GNN models on test sets.}
    \label{tab:trainingacc}
    \setlength{\tabcolsep}{5pt}
    \begin{tabular}{lcccccc}
      \toprule
      $\mathbf{aggr}$ & Cora & CiteSeer & Amazon & Yelp & MUTAG & ENZYMES \\
      \midrule
      sum & 81.02\% & 71.68\% & 95.97\% & 79.12\% & 88.89\% & 68.33\%\\
      max & 80.35\% & 72.40\% & 96.19\% & 79.02\% & 77.78\% & 68.33\% \\
      mean & 81.39\% & 72.58\% & 97.55\% & 80.52\% & 88.89\% & 71.67\%\\
      \bottomrule
    \end{tabular}
\end{table}

\section{Comprehensive Experimental Results}
\label{appx:fullexp}

\subsection{Comparison with \textsc{SCIP-MPNN}}
As described in Section~\ref{sec:exp}.2, we compared the performance of \textsc{GNNev} and \textsc{SCIP-MPNN} on sum-aggregated {3-layer} GNNs.
\tablename~\ref{tab:fullcompare_cora}
shows the comprehensive comparison results under different global structural perturbation budgets $\Delta$.
The average runtime, the number of solved tasks, and the number of winning tasks (i.e., when the runtime of the current verifier is shorter than other verifier(s) on that task) for each verifier are reported for the set of all instances and the set of robust instances, respectively.
Across models under different budgets, \textsc{GNNev} consistently outperformed \textsc{SCIP-MPNN} in terms of the number of winning tasks, especially on the robust instances.

\begin{table*}[htbp]
\centering
\caption{Detailed comparison results of \textsc{GNNev} and \textsc{SCIP-MPNN} on node (Cora, CiteSeer) and graph (MUTAG, ENZYMES) classification for 3-layer GNNs and a range of perturbation budgets.
The \#Win column indicates the number of  
winning tasks (i.e.\ when the runtime of the current verifier is shorter than other verifier(s) on that task), and the larger number on the same set of instances is shown in \textbf{bold}.}
\label{tab:fullcompare_cora}
\resizebox{\textwidth}{!}{
\begin{tabular}{l|cccccc|cccccc}
\toprule
\multirow{3}{*}{$\Delta$} & \multicolumn{6}{c|}{\textsc{SCIP-MPNN}} & \multicolumn{6}{c}{\textsc{GNNev}} \\
 & \multicolumn{3}{c}{All instances} & \multicolumn{3}{c|}{Robust instances} & \multicolumn{3}{c}{All instances} & \multicolumn{3}{c}{Robust instances} \\
 \cmidrule(lr){2-4} \cmidrule(lr){5-7} \cmidrule(lr){8-10} \cmidrule(lr){11-13}
 & Time(s) & \#Solved & \#Win & Time(s) & \#Solved & \#Win & Time(s) & \#Solved & \#Win & Time(s) & \#Solved & \#Win \\
\midrule
\multicolumn{13}{l}{\textit{Cora}} \\
 1 & 13.84 & 2,708 & 73 & 13.06 & 2,544 & 59 & 5.06 & 2,708 & \textbf{2,635} & 4.72 & 2,544 & \textbf{2,485} \\
 2 & 13.86 & 2,708 & 58 & 12.15 & 2,404 & 45 & 5.12 & 2,708 & \textbf{2,650} & 4.45 & 2,404 & \textbf{2,359} \\
 5 & 13.79 & 2,708 & 132 & 10.32 & 2,259 & 104 & 6.92 & 2,708 & \textbf{2,576} & 5.39 & 2,259 & \textbf{2,155} \\
 10 & 13.87 & 2,706 & 162 & 10.25 & 2,249 & 136 & 8.10 & 2,708 & \textbf{2,546} & 6.53 & 2,251 & \textbf{2,115} \\
 50 & 13.65 & 2,706 & 178 & 10.08 & 2,249 & 148 & 7.79 & 2,708 & \textbf{2,530} & 6.30 & 2,251 & \textbf{2,103} \\
\midrule
\multicolumn{13}{l}{\textit{CiteSeer}} \\
 1 & 9.59 & 3,308 & 1,126 & 9.12 & 3,156 & 1,009 & 6.14 & 3,312 & \textbf{2,186} & 5.70 & 3,160 & \textbf{2,151} \\
 2 & 9.59 & 3,307 & 1,300 & 8.77 & 3,076 & 1,134 & 6.84 & 3,312 & \textbf{2,012} & 6.08 & 3,081 & \textbf{1,947} \\
 5 & 9.63 & 3,307 & 1,550 & 8.04 & 3,010 & 1,327 & 9.07 & 3,308 & \textbf{1,759} & 7.56 & 3,010 & \textbf{1,686} \\
 10 & 10.26 & 3,307 & 1,548 & 8.43 & 2,985 & 1,319 & 9.51 & 3,308 & \textbf{1,761} & 8.06 & 2,990 & \textbf{1,672} \\
 50 & 10.37 & 3,305 & 1,561 & 7.52 & 2,978 & 1,328 & 9.41 & 3,295 & \textbf{1,750} & 7.14 & 2,980 & \textbf{1,654} \\
\midrule
\multicolumn{13}{l}{\textit{MUTAG}} \\
 1 & 31.08 & 188 & 0 & 0.42 & 6 & 0 & 3.38 & 188 & \textbf{188} & 0.03 & 6 & \textbf{6} \\
 2 & 111.60 & 167 & 37 & 9.68 & 6 & 0 & 37.93 & 185 & \textbf{148} & 7.12 & 6 & \textbf{6} \\
 5 & 193.22 & 106 & 34 & 5.48 & 4 & 1 & 57.58 & 148 & \textbf{134} & 4.75 & 4 & \textbf{3} \\
 10 & 212.52 & 70 & 19 & 0.00 & 0 & \textbf{0} & 79.37 & 136 & \textbf{130} & 0.00 & 0 & \textbf{0} \\
 50 & 193.80 & 98 & 21 & 0.00 & 0 & \textbf{0} & 81.89 & 136 & \textbf{130} & 0.00 & 0 & \textbf{0} \\
\midrule
\multicolumn{13}{l}{\textit{ENZYMES}} \\
 1 & 117.70 & 572 & 10 & 113.95 & 556 & 10 & 36.33 & 600 & \textbf{590} & 34.54 & 582 & \textbf{572} \\
 2 & 121.51 & 491 & 226 & 92.09 & 398 & 183 & 106.15 & 495 & \textbf{298} & 77.81 & 399 & \textbf{241} \\
 5 & 255.15 & 35 & 28 & 74.34 & 2 & 1 & 85.17 & 152 & \textbf{142} & 22.14 & 43 & \textbf{43} \\
 10 & 277.49 & 20 & 11 & 11.27 & 1 & 0 & 60.26 & 210 & \textbf{202} & 4.67 & 9 & \textbf{9} \\
 50 & 291.15 & 20 & 5 & 0.55 & 1 & 0 & 56.80 & 544 & \textbf{539} & 0.03 & 2 & \textbf{2} \\
\bottomrule
\end{tabular}
}
\end{table*}

\subsection{Verification Results and Performance}

We evaluated the performance of \textsc{GNNev} on 3-layer GNNs.
Due to the large number of nodes in Amazon and Yelp, we sampled 1,000 nodes from each dataset to form the instances for verification.
Initially, we selected nodes, for which the size of 3-hop incoming neighbours set
was within the range $[2,200]$ and which were correctly predicted by the model.
Next, if the number of nodes in a class was less than 500, all of them were selected, and nodes in another class were uniformly sampled to make the instances up to 1,000; otherwise, 500 nodes were uniformly sampled from each class.
We built two kinds of fragile edge sets. For edge deletion, we set all existing edges as fragile, i.e., $F=E$. For edge addition, we sampled a set of non-edges in each graph as fragile. Specifically, we randomly selected one and three non-edge(s) pointing to each node to build $F$ for node- and graph-classification datasets, respectively. This ensures that there are several fragile edges for any verification task.
Note that \textsc{GNNev} is able to accept any given $F \subseteq V \times V$ and does not depend on a specific sampling method.

\begin{figure}[t]
    \centering

    \begin{subfigure}[b]{0.47\textwidth}
        \centering
        \includegraphics[trim={0.3cm 0cm 0.24cm 0cm}, clip, width=\linewidth]{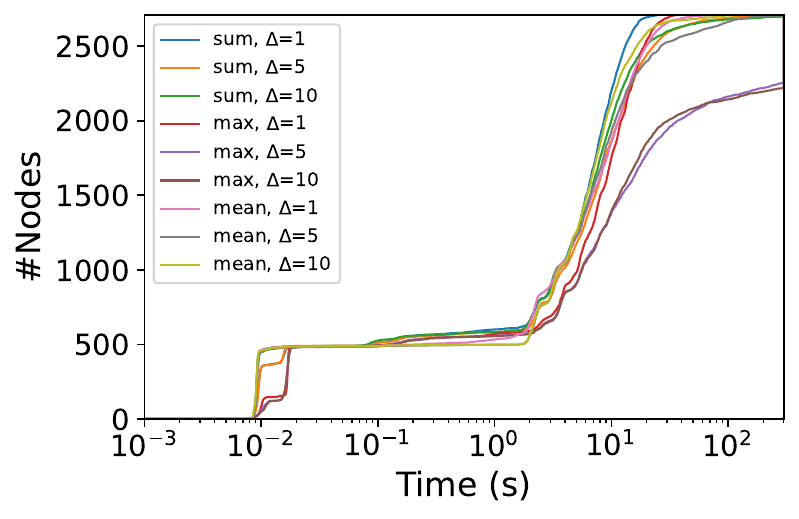}
        \caption{Cora}
        \label{fig:time_cora_delete}
    \end{subfigure}
    \hfill
    \begin{subfigure}[b]{0.47\textwidth}
        \centering
        \includegraphics[trim={0.3cm 0cm 0.24cm 0cm}, clip, width=\linewidth]{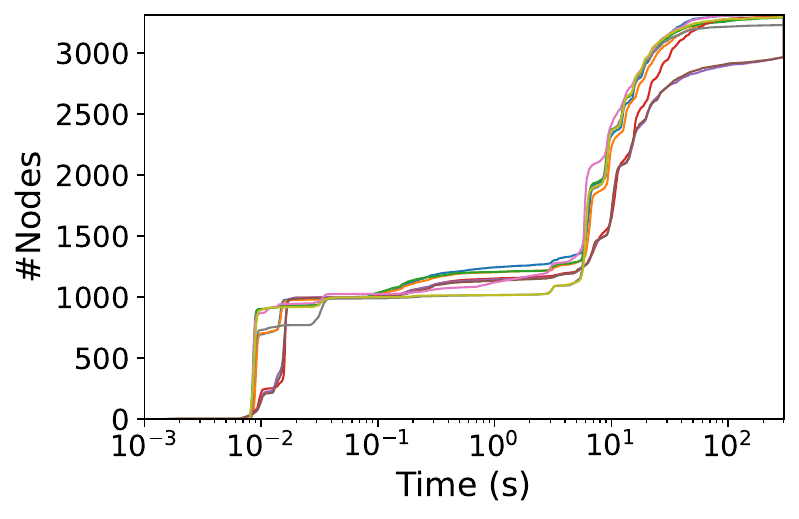}
        \caption{CiteSeer}
        \label{fig:time_citeseer_delete}
    \end{subfigure}
    
    \begin{subfigure}[b]{0.47\textwidth}
        \centering
        \includegraphics[trim={0.3cm 0cm 0.24cm 0cm}, clip, width=\linewidth]{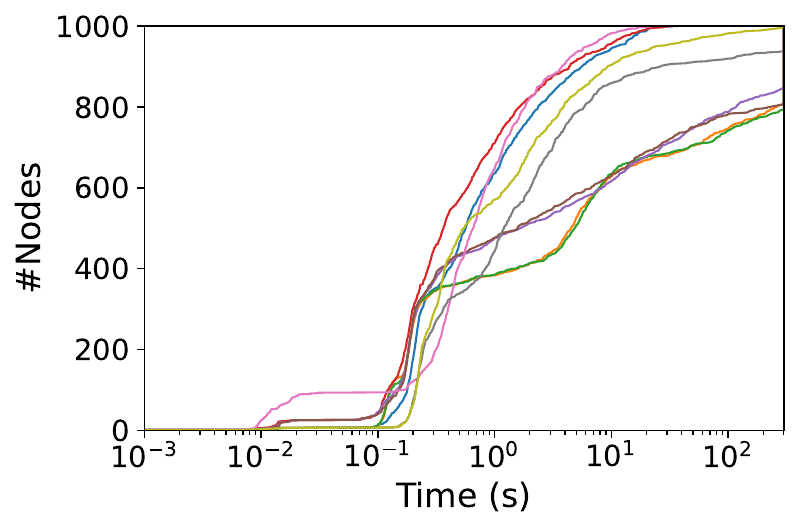}
        \caption{Amazon}
        \label{fig:time_amazon_delete}
    \end{subfigure}
    \hfill
    \begin{subfigure}[b]{0.47\textwidth}
        \centering
        \includegraphics[trim={0.3cm 0cm 0.24cm 0cm}, clip, width=\linewidth]{fig/time_Yelp_delete.pdf}
        \caption{Yelp}
        \label{fig:time_yelp_delete}
    \end{subfigure}

    \begin{subfigure}[b]{0.47\textwidth}
        \centering
        \includegraphics[trim={0.26cm 0cm 0.24cm 0cm}, clip, width=\linewidth]{fig/time_MUTAG_delete.pdf}
        \caption{MUTAG}
        \label{fig:time_mutag_delete}
    \end{subfigure}
    \hfill
    \begin{subfigure}[b]{0.47\textwidth}
        \centering
        \includegraphics[trim={0.26cm 0cm 0.24cm 0cm}, clip, width=\linewidth]{fig/time_ENZYMES_delete.pdf}
        \caption{ENZYMES}
        \label{fig:time_enzymes_delete}
    \end{subfigure}

    \caption{The number of tasks solved by \textsc{GNNev} {on node classification (a-d) and graph classification (e-f)} plotted against runtime under different aggregations and budgets. Only edge deletions are allowed ($F=E$). The verification objective is as defined in Section~\ref{sec:encoding}, Eq.~(\ref{eq:obj}).}
    \label{fig:time}
\end{figure}

\begin{figure}[t]
    \centering

    \begin{subfigure}[b]{0.47\linewidth}
        \centering
        \includegraphics[trim={0.3cm 0cm 0.24cm 0cm}, clip, width=\linewidth]{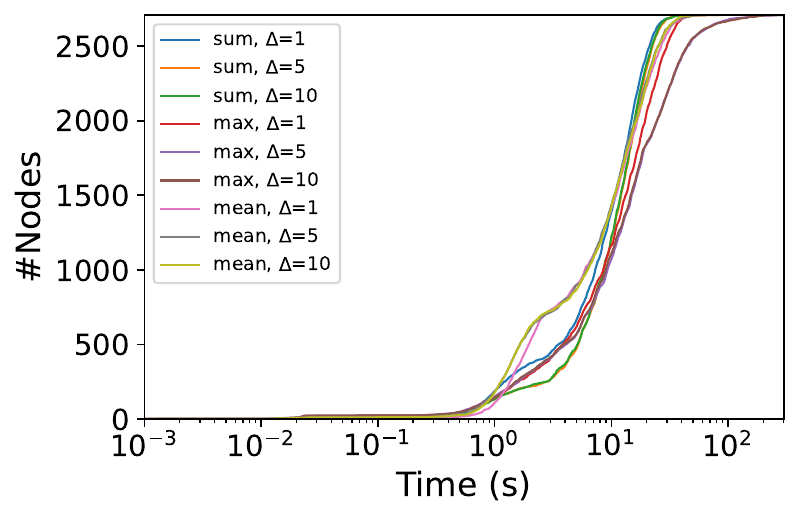}
        \caption{Cora}
        \label{fig:time_cora_add}
    \end{subfigure}
    \hfill
    \begin{subfigure}[b]{0.47\linewidth}
        \centering
        \includegraphics[trim={0.3cm 0cm 0.24cm 0cm}, clip, width=\linewidth]{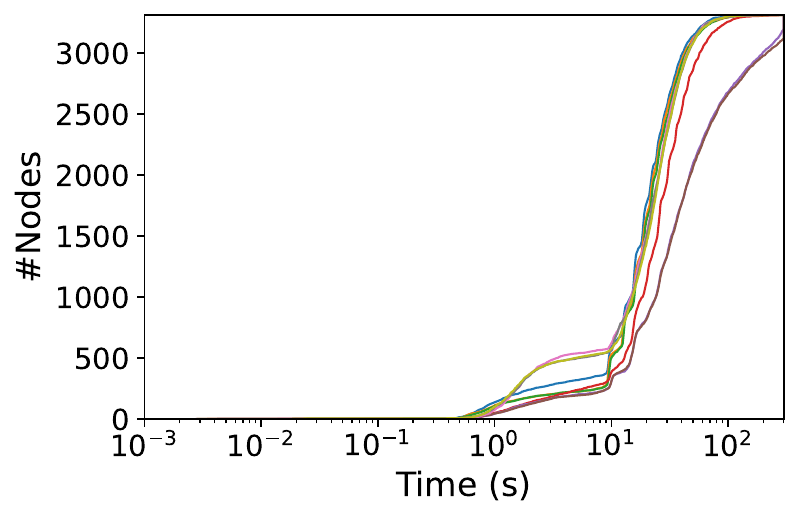}
        \caption{CiteSeer}
        \label{fig:time_citeseer_add}
    \end{subfigure}
    
    \begin{subfigure}[b]{0.47\linewidth}
        \centering
        \includegraphics[trim={0.3cm 0cm 0.24cm 0cm}, clip, width=\linewidth]{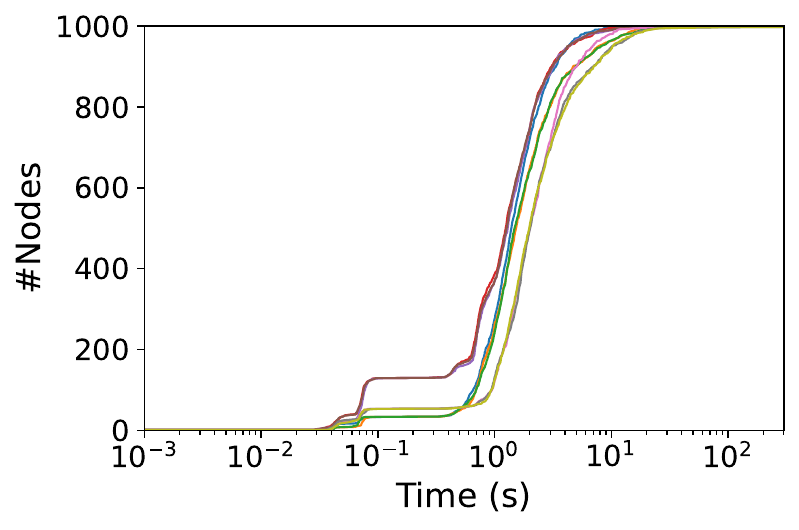}
        \caption{Amazon}
        \label{fig:time_amazon_add}
    \end{subfigure}
    \hfill
    \begin{subfigure}[b]{0.47\linewidth}
        \centering
        \includegraphics[trim={0.3cm 0cm 0.24cm 0cm}, clip, width=\linewidth]{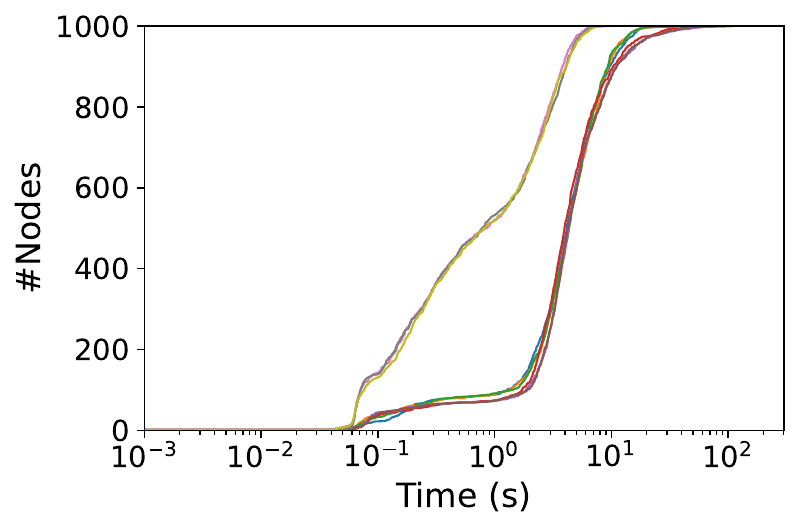}
        \caption{Yelp}
        \label{fig:time_yelp_add}
    \end{subfigure}

    \begin{subfigure}[b]{0.47\linewidth}
        \centering
        \includegraphics[trim={0.26cm 0cm 0.24cm 0cm}, clip, width=\linewidth]{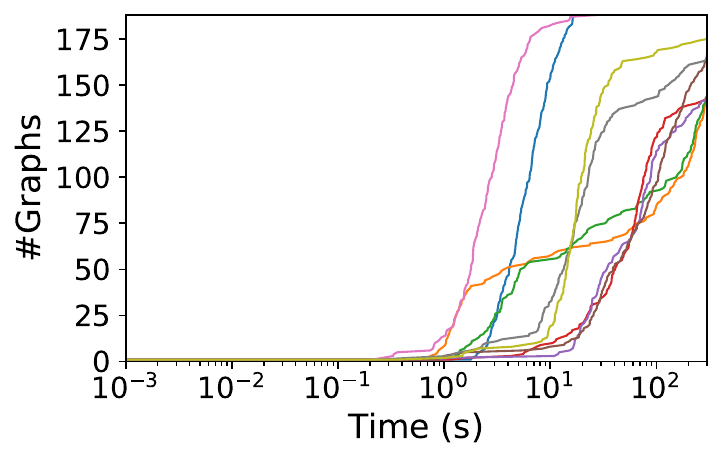}
        \caption{MUTAG}
        \label{fig:time_mutag_add}
    \end{subfigure}
    \hfill
    \begin{subfigure}[b]{0.47\linewidth}
        \centering
        \includegraphics[trim={0.26cm 0cm 0.24cm 0cm}, clip, width=\linewidth]{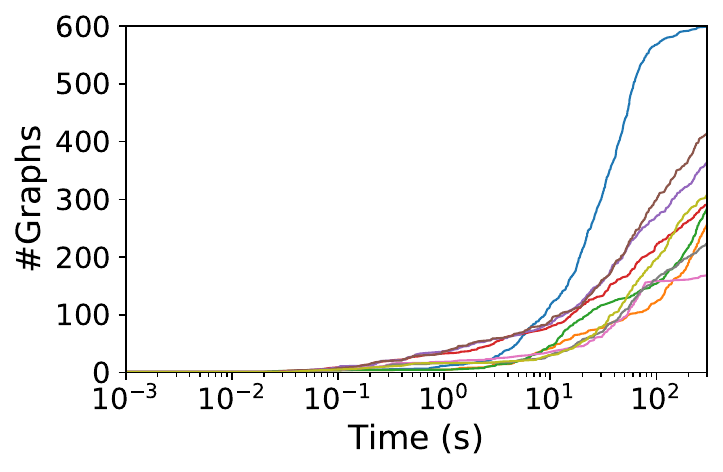}
        \caption{ENZYMES}
        \label{fig:time_enzymes_add}
    \end{subfigure}

    \caption{The number of tasks solved by \textsc{GNNev} {on node classification (a-d) and graph classification (e-f)} plotted against runtime under different aggregations and budgets. Only edge additions are allowed under predefined fragile non-edge sets as detailed in Appendix~\ref{appx:fullexp}.2.}
    \label{fig:time_add}
\end{figure}

\begin{figure*}[ht]
\centering
\setlength{\tabcolsep}{3pt}
\renewcommand{\arraystretch}{1.5}

\resizebox{\textwidth}{!}{
\begin{NiceTabular}{m{0.12\textwidth}ccc}
    & {$\mathbf{aggr}=\text{sum}$} & {$\mathbf{aggr}=\text{max}$} & {$\mathbf{aggr}=\text{mean}$} \\

    \Block[v-center]{}{Cora} &
    \begin{subfigure}{0.28\textwidth}
        \centering
        \includegraphics[width=\linewidth]{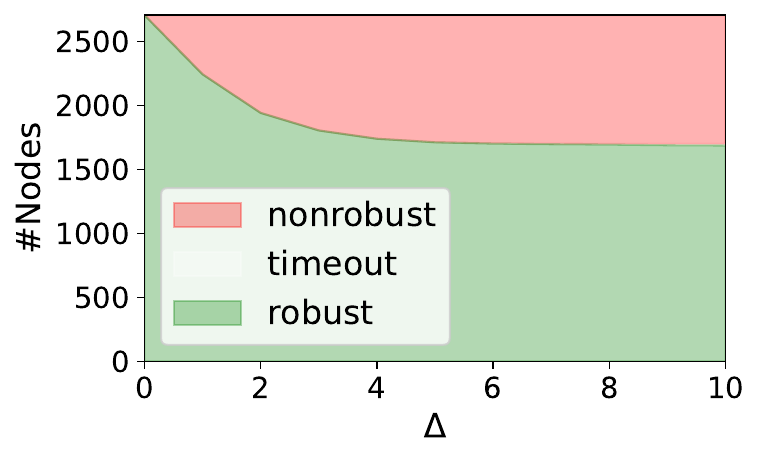}
    \end{subfigure} &
    \begin{subfigure}{0.28\textwidth}
        \centering
        \includegraphics[width=\linewidth]{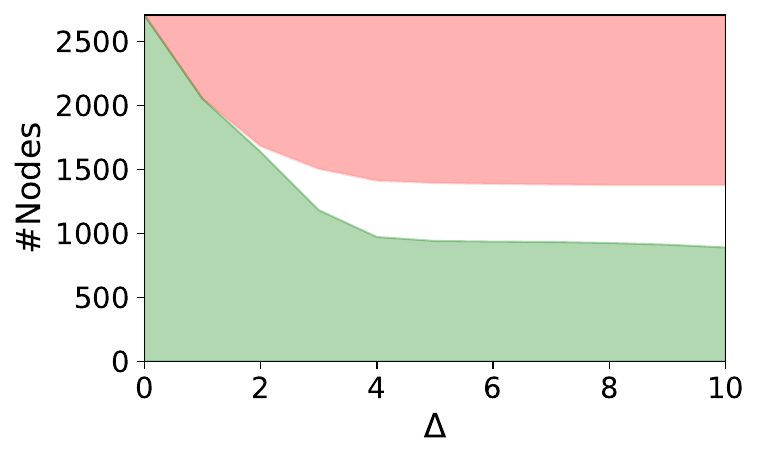}
    \end{subfigure} &
    \begin{subfigure}{0.28\textwidth}
        \centering
        \includegraphics[width=\linewidth]{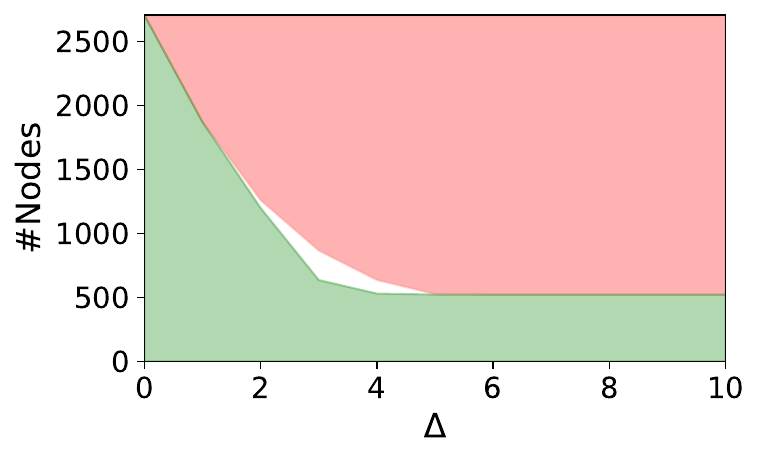}
    \end{subfigure} \\

    \Block[v-center]{}{CiteSeer} &
    \begin{subfigure}{0.28\textwidth}
        \centering
        \includegraphics[width=\linewidth]{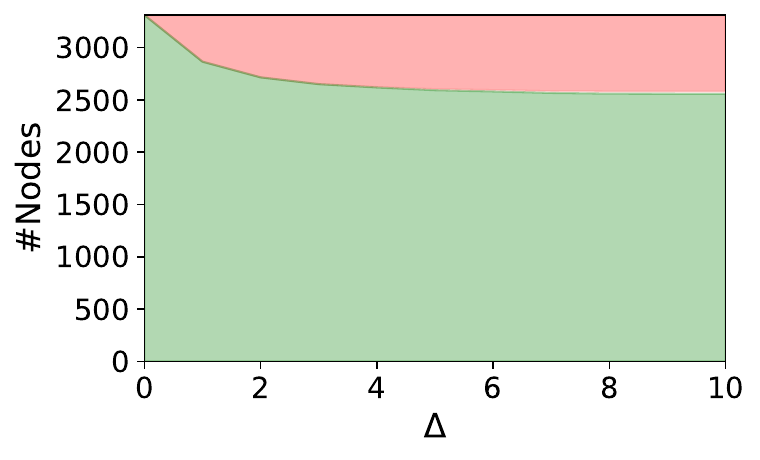}
    \end{subfigure} &
    \begin{subfigure}{0.28\textwidth}
        \centering
        \includegraphics[width=\linewidth]{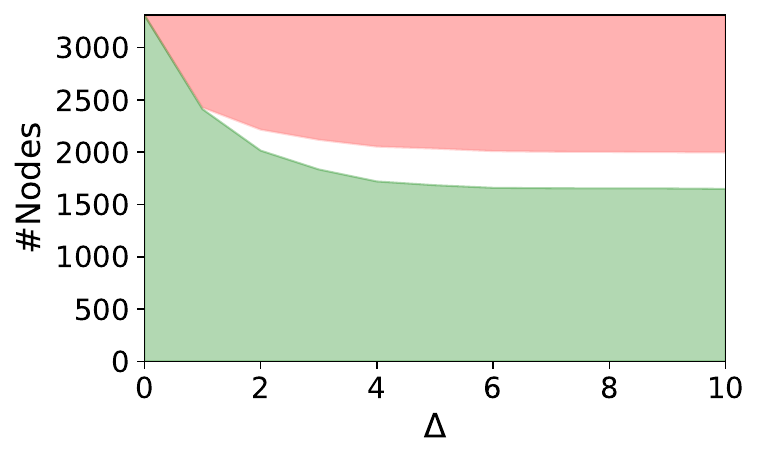}
    \end{subfigure} &
    \begin{subfigure}{0.28\textwidth}
        \centering
        \includegraphics[width=\linewidth]{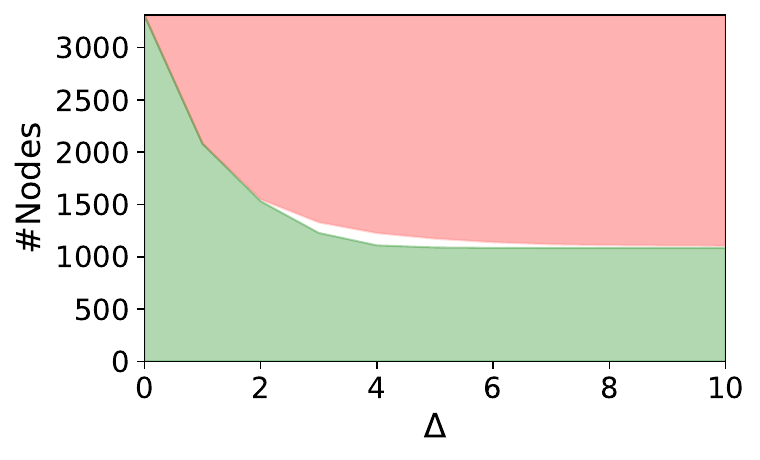}
    \end{subfigure} \\

    \Block[v-center]{}{Amazon} &
    \begin{subfigure}{0.28\textwidth}
        \centering
        \includegraphics[width=\linewidth]{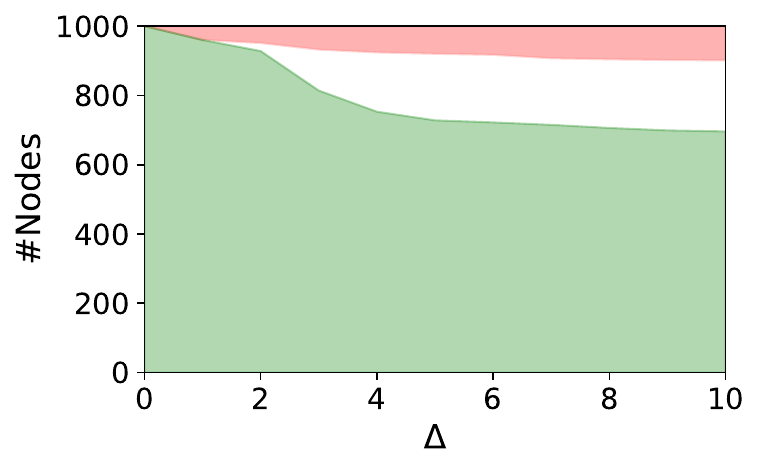}
    \end{subfigure} &
    \begin{subfigure}{0.28\textwidth}
        \centering
        \includegraphics[width=\linewidth]{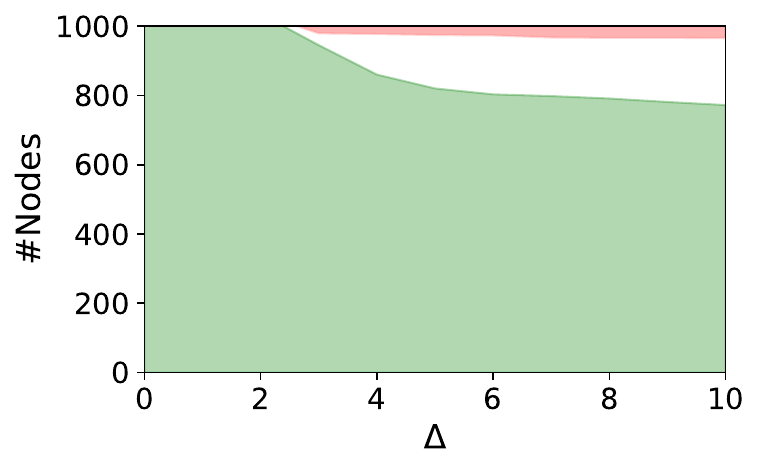}
    \end{subfigure} &
    \begin{subfigure}{0.28\textwidth}
        \centering
        \includegraphics[width=\linewidth]{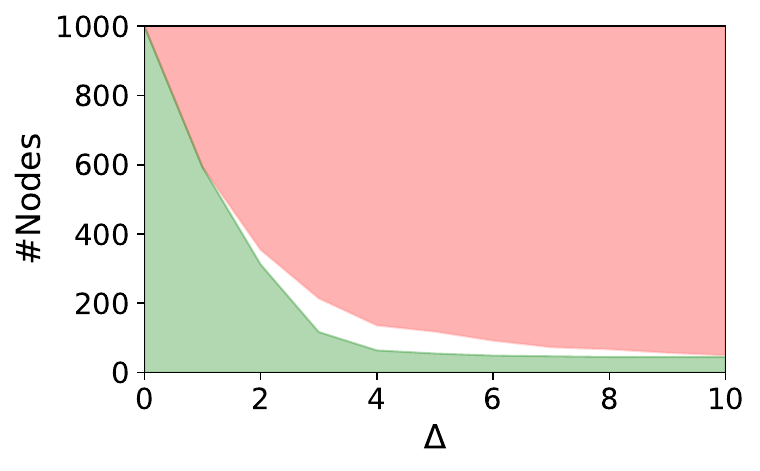}
    \end{subfigure} \\

    \Block[v-center]{}{Yelp} &
    \begin{subfigure}{0.28\textwidth}
        \centering
        \includegraphics[width=\linewidth]{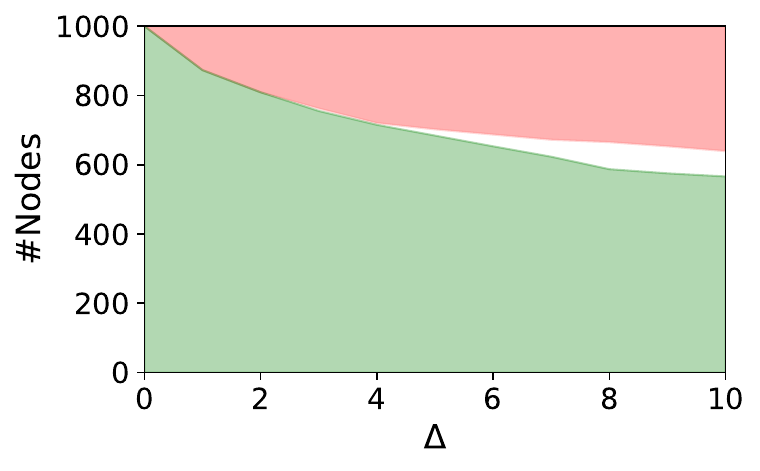}
    \end{subfigure} &
    \begin{subfigure}{0.28\textwidth}
        \centering
        \includegraphics[width=\linewidth]{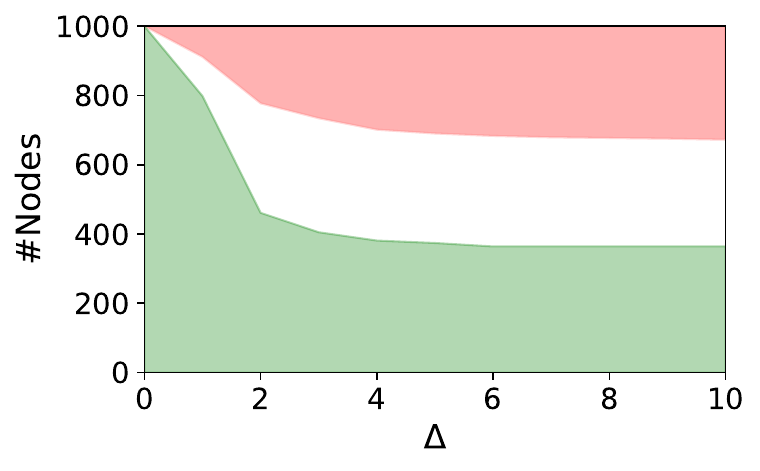}
    \end{subfigure} &
    \begin{subfigure}{0.28\textwidth}
        \centering
        \includegraphics[width=\linewidth]{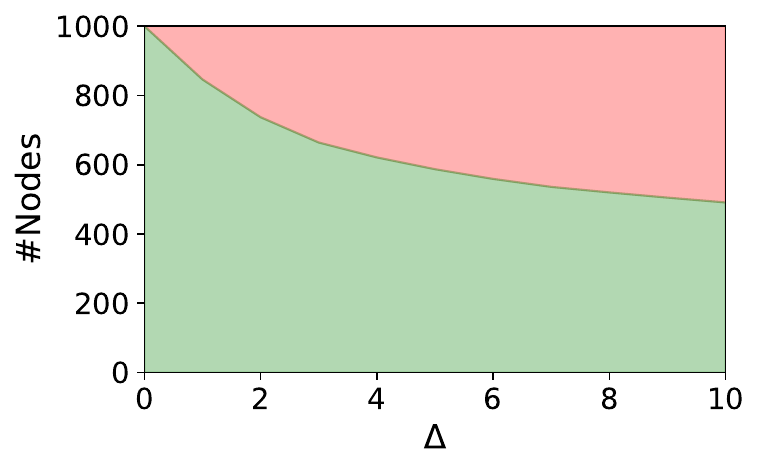}
    \end{subfigure} \\

    \Block[v-center]{}{MUTAG} &
    \begin{subfigure}{0.28\textwidth}
        \centering
        \includegraphics[width=\linewidth]{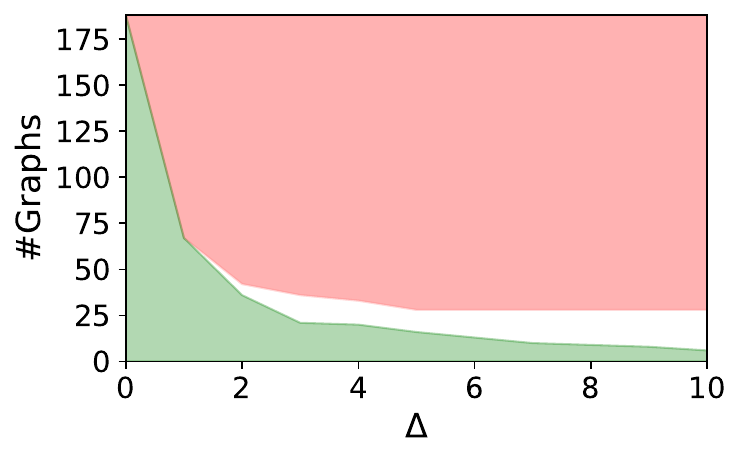}
    \end{subfigure} &
    \begin{subfigure}{0.28\textwidth}
        \centering
        \includegraphics[width=\linewidth]{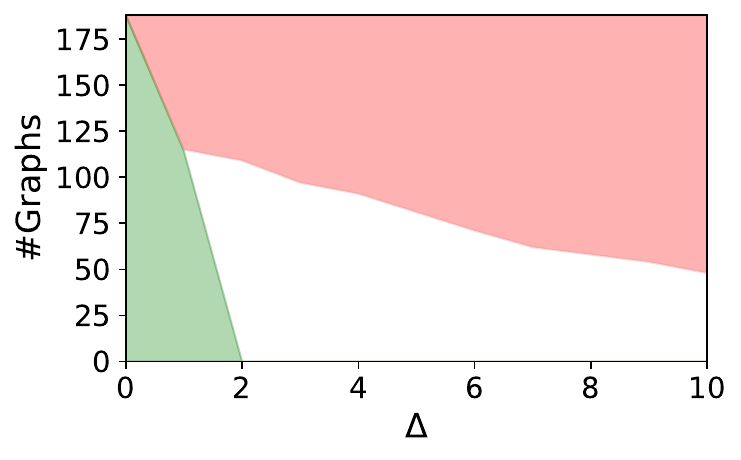}
    \end{subfigure} &
    \begin{subfigure}{0.28\textwidth}
        \centering
        \includegraphics[width=\linewidth]{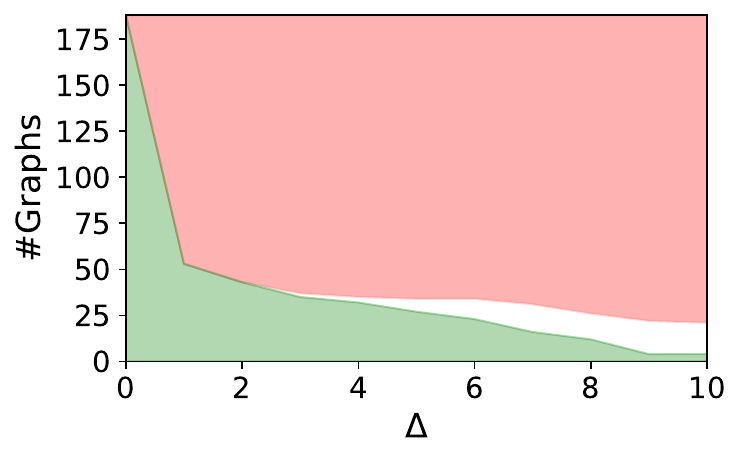}
    \end{subfigure} \\

    \Block[v-center]{}{ENZYMES} &
    \begin{subfigure}{0.28\textwidth}
        \centering
        \includegraphics[width=\linewidth]{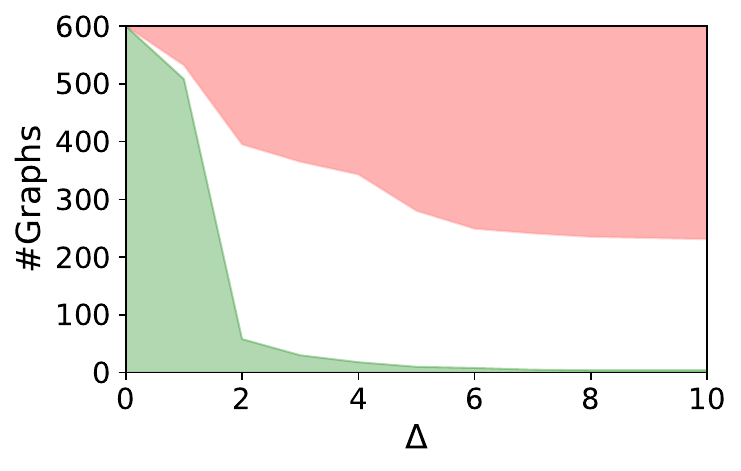}
    \end{subfigure} &
    \begin{subfigure}{0.28\textwidth}
        \centering
        \includegraphics[width=\linewidth]{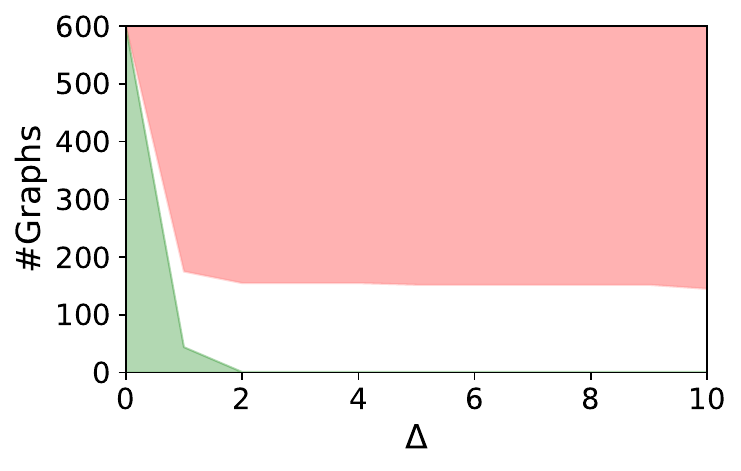}
    \end{subfigure} &
    \begin{subfigure}{0.28\textwidth}
        \centering
        \includegraphics[width=\linewidth]{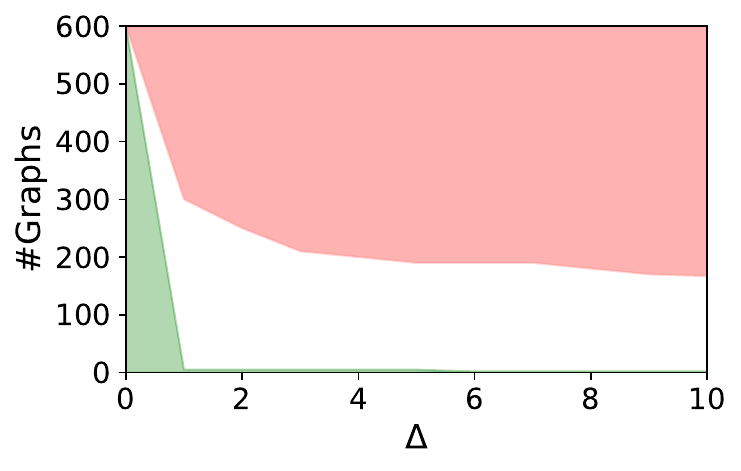}
    \end{subfigure}
\end{NiceTabular}
}
\caption{The evolution of the number of robust and nonrobust nodes verified by \textsc{GNNev} {on node classification (top four rows) and graph classification (bottom two rows)} as the global structural perturbation budget $\Delta$ increases. Only edge deletions are allowed, and all edges are set as fragile.}
\label{fig:evolution_delete}
\end{figure*}

\begin{figure*}[ht]
\centering
\setlength{\tabcolsep}{3pt}
\renewcommand{\arraystretch}{1.5}

\resizebox{\textwidth}{!}{
\begin{NiceTabular}{m{0.12\textwidth}ccc}
    & {$\mathbf{aggr}=\text{sum}$} & {$\mathbf{aggr}=\text{max}$} & {$\mathbf{aggr}=\text{mean}$} \\

    \Block[v-center]{}{Cora} &
    \begin{subfigure}{0.28\textwidth}
        \centering
        \includegraphics[width=\linewidth]{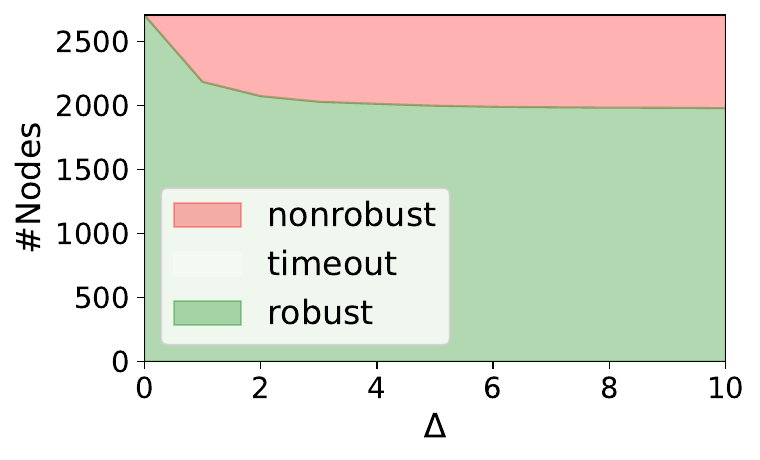}
    \end{subfigure} &
    \begin{subfigure}{0.28\textwidth}
        \centering
        \includegraphics[width=\linewidth]{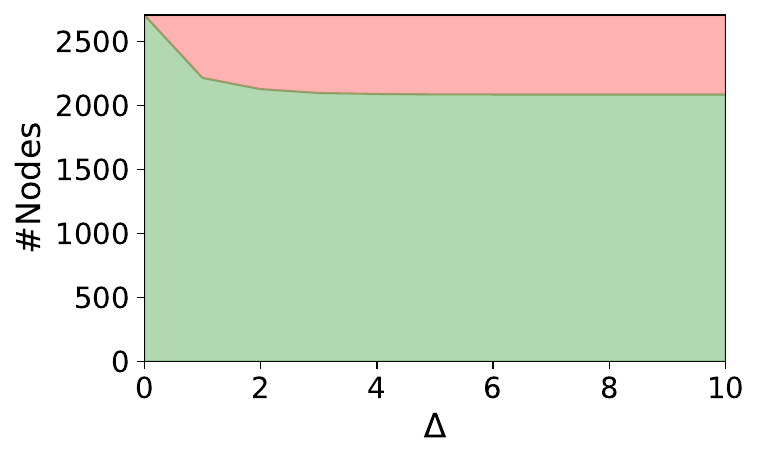}
    \end{subfigure} &
    \begin{subfigure}{0.28\textwidth}
        \centering
        \includegraphics[width=\linewidth]{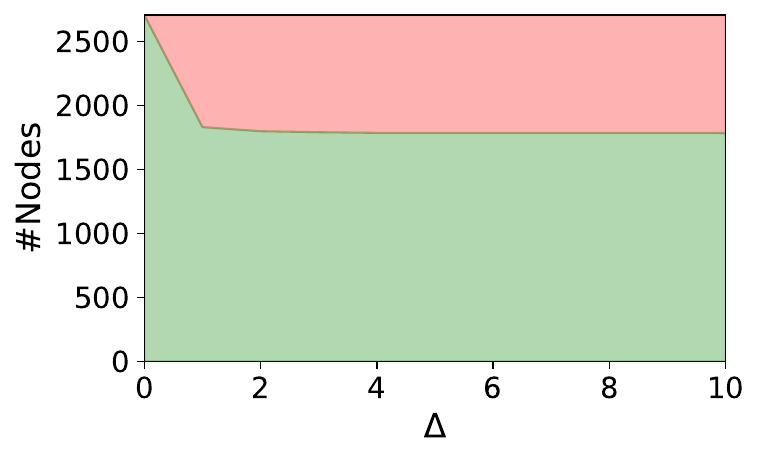}
    \end{subfigure} \\

    \Block[v-center]{}{CiteSeer} &
    \begin{subfigure}{0.28\textwidth}
        \centering
        \includegraphics[width=\linewidth]{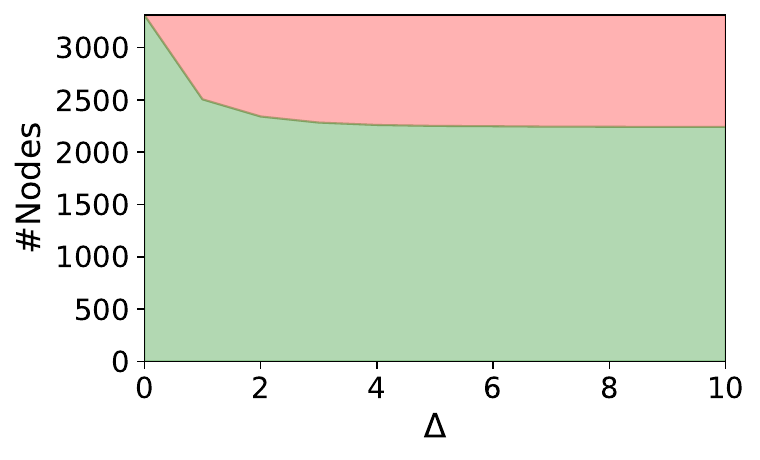}
    \end{subfigure} &
    \begin{subfigure}{0.28\textwidth}
        \centering
        \includegraphics[width=\linewidth]{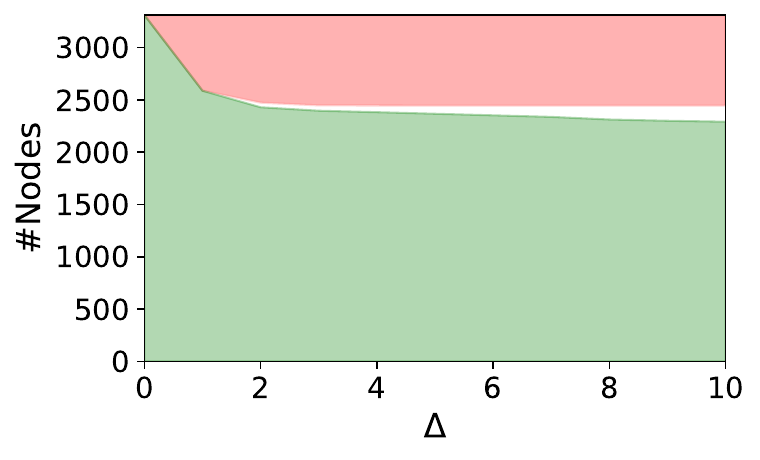}
    \end{subfigure} &
    \begin{subfigure}{0.28\textwidth}
        \centering
        \includegraphics[width=\linewidth]{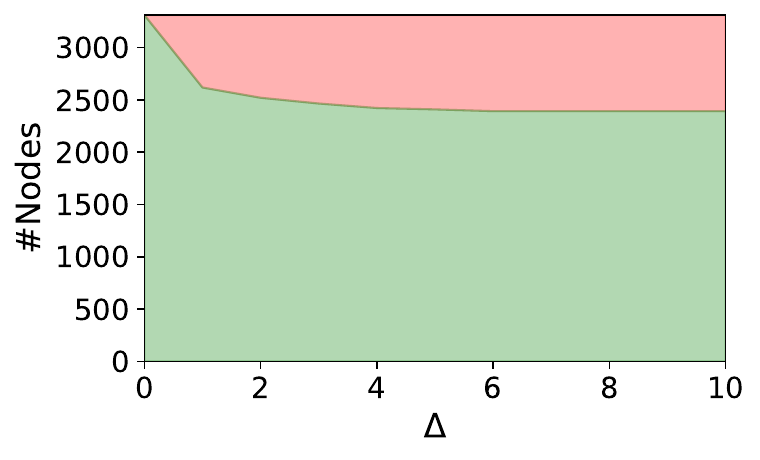}
    \end{subfigure} \\

    \Block[v-center]{}{Amazon} &
    \begin{subfigure}{0.28\textwidth}
        \centering
        \includegraphics[width=\linewidth]{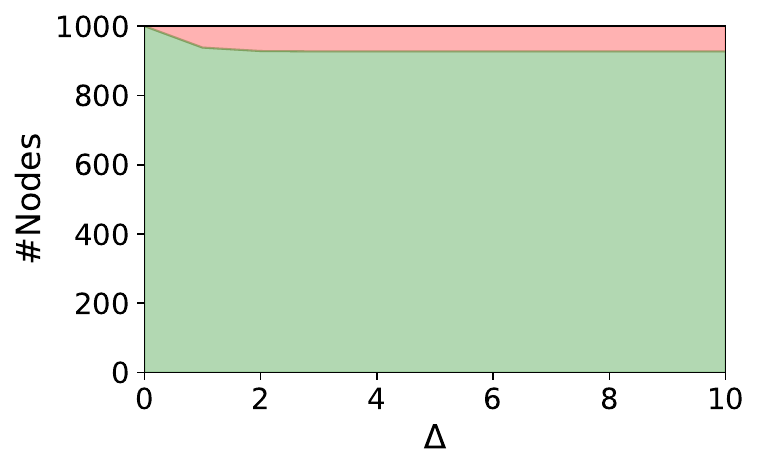}
    \end{subfigure} &
    \begin{subfigure}{0.28\textwidth}
        \centering
        \includegraphics[width=\linewidth]{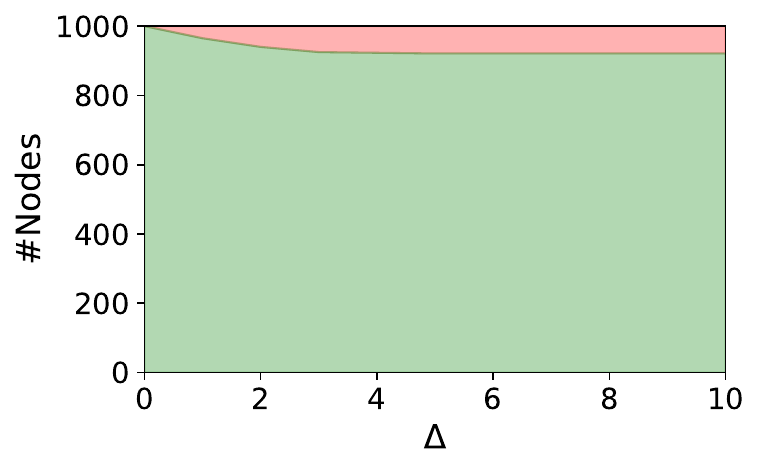}
    \end{subfigure} &
    \begin{subfigure}{0.28\textwidth}
        \centering
        \includegraphics[width=\linewidth]{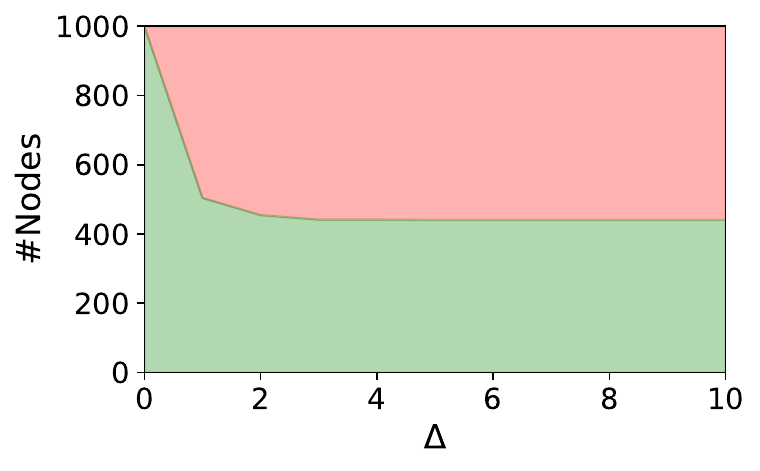}
    \end{subfigure} \\

    \Block[v-center]{}{Yelp} &
    \begin{subfigure}{0.28\textwidth}
        \centering
        \includegraphics[width=\linewidth]{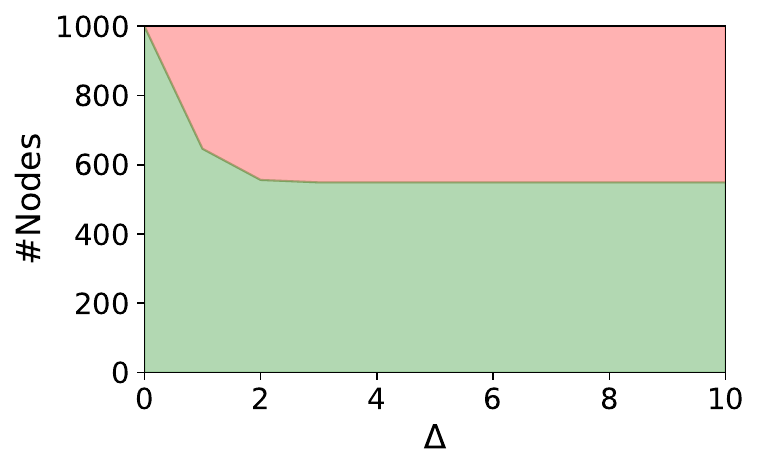}
    \end{subfigure} &
    \begin{subfigure}{0.28\textwidth}
        \centering
        \includegraphics[width=\linewidth]{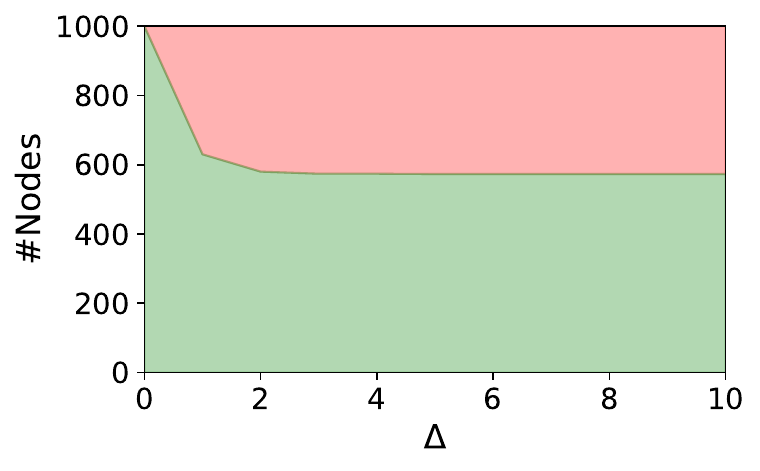}
    \end{subfigure} &
    \begin{subfigure}{0.28\textwidth}
        \centering
        \includegraphics[width=\linewidth]{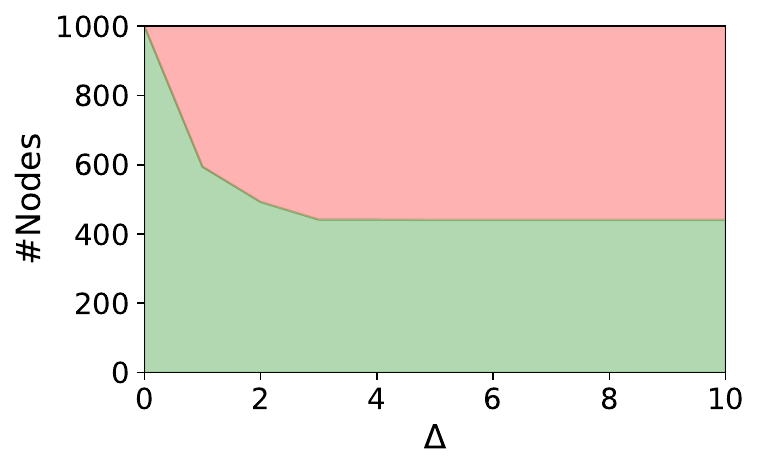}
    \end{subfigure} \\

    \Block[v-center]{}{MUTAG} &
    \begin{subfigure}{0.28\textwidth}
        \centering
        \includegraphics[width=\linewidth]{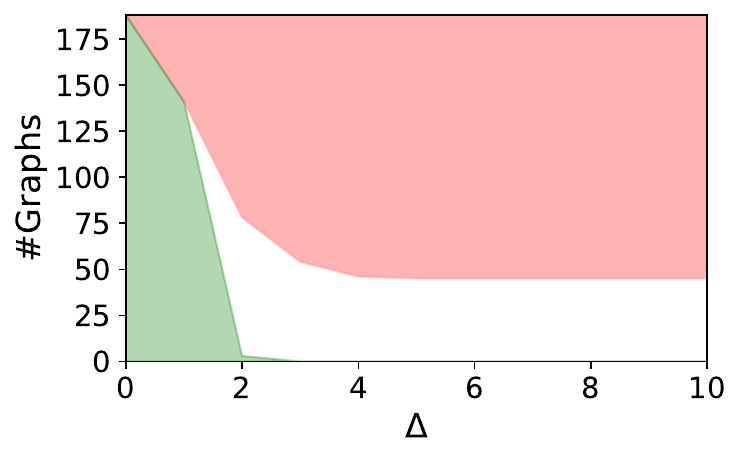}
    \end{subfigure} &
    \begin{subfigure}{0.28\textwidth}
        \centering
        \includegraphics[width=\linewidth]{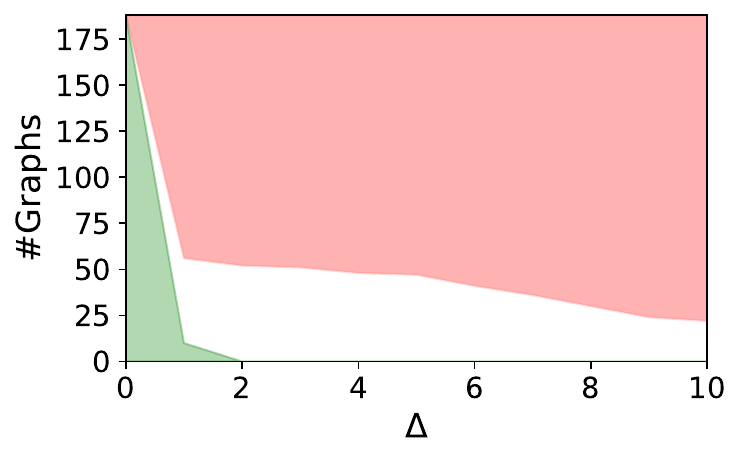}
    \end{subfigure} &
    \begin{subfigure}{0.28\textwidth}
        \centering
        \includegraphics[width=\linewidth]{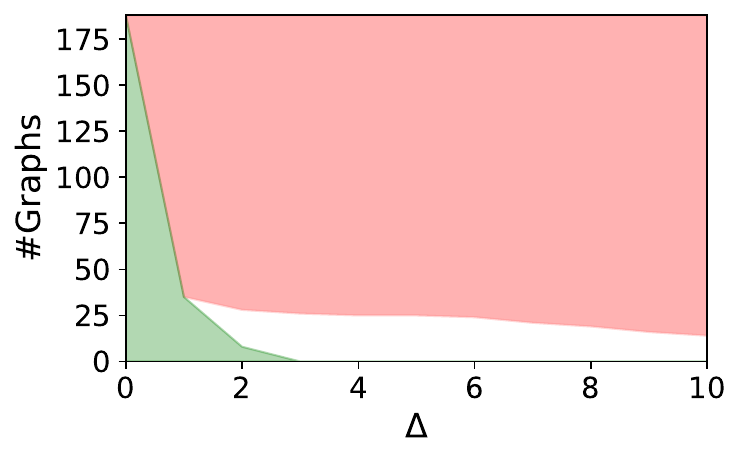}
    \end{subfigure} \\

    \Block[v-center]{}{ENZYMES} &
    \begin{subfigure}{0.28\textwidth}
        \centering
        \includegraphics[width=\linewidth]{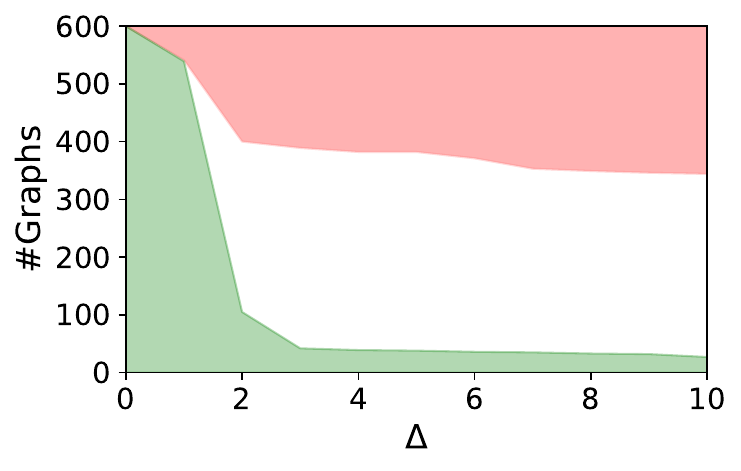}
    \end{subfigure} &
    \begin{subfigure}{0.28\textwidth}
        \centering
        \includegraphics[width=\linewidth]{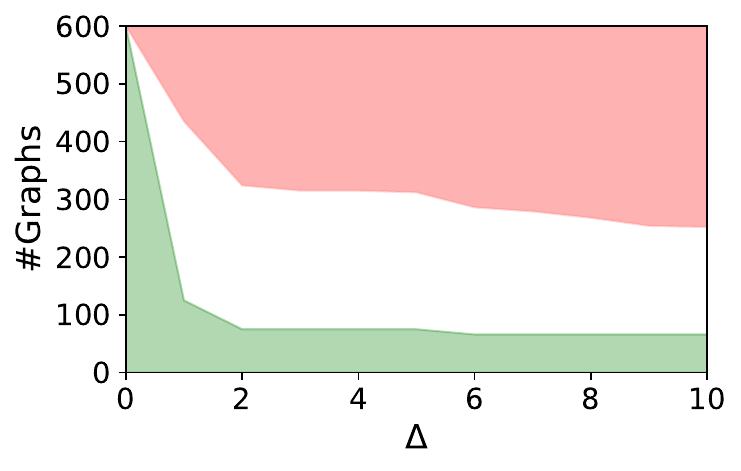}
    \end{subfigure} &
    \begin{subfigure}{0.28\textwidth}
        \centering
        \includegraphics[width=\linewidth]{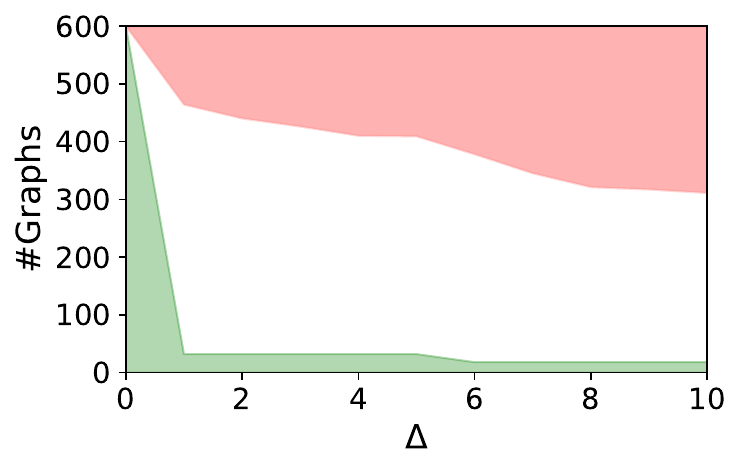}
    \end{subfigure}
\end{NiceTabular}
}
\caption{The evolution of the number of robust and nonrobust nodes verified by \textsc{GNNev} {on node classification (top four rows) and graph classification (bottom two rows)} as the global structural perturbation budget $\Delta$ increases. Only edge additions are allowed under predefined fragile non-edge sets as detailed in Appendix~\ref{appx:fullexp}.2.}
\label{fig:evolution_add}
\end{figure*}

\subsubsection{Results.}
\figurename~\ref{fig:time},\ref{fig:time_add} illustrate the runtime of \textsc{GNNev} under different aggregation functions and budgets for {structural perturbations (edge deletion and addition, respectively)}.
For edge addition, \textsc{GNNev} successfully solved all tasks within the time limit for Cora, Amazon, and Yelp (\figurename~\ref{fig:time_cora_add},\ref{fig:time_amazon_add},\ref{fig:time_yelp_add}).
For CiteSeer, there were 2 ($\Delta=1$), 78 ($\Delta=5$) and 152 ($\Delta=10$) tasks left unsolved with max aggregation, indicating that the difficulty of the problem increases with larger budgets (\figurename~\ref{fig:time_citeseer_add}).
Since the fragile edge set $F$ only included one non-edge for each node, the number of edges that can be perturbed in each task is much lower for edge addition, which makes the corresponding CSP easier to solve compared with edge deletion.
Note that setting $F$ to a large size can lead to a notable decrease in verification efficiency.

\figurename~\ref{fig:evolution_delete},\ref{fig:evolution_add} show the evolution of the number of robust and nonrobust nodes verified by \textsc{GNNev} as the structural perturbation budget increases.
We observed that, on node-classification datasets, the robustness of mean-aggregated GNNs is significantly lower compared to other GNNs, on Cora, CiteSeer and Amazon for edge deletion, and on Amazon for edge addition.
On graph-classification datasets, lack of robustness has been identified across aggregations.
This indicates that \textsc{GNNev} can reveal the robustness flaws of GNN models, which is beneficial for deploying these models in high-stakes applications.
Furthermore, for most nodes, the transition from robust to nonrobust occurred when $\Delta \le 3$, which also demonstrates the vulnerability of GNN models to simple adversarial attacks.

\begin{figure}[h]
    \centering
    
    \begin{subfigure}[b]{0.48\textwidth}
        \centering
        \includegraphics[trim={0cm 0cm 0cm 0cm}, clip, width=\linewidth]{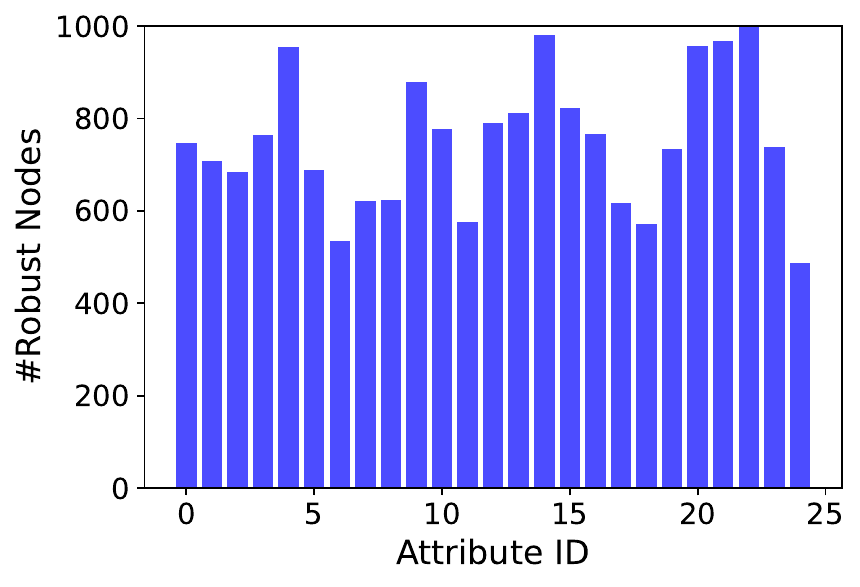}
        \caption{Amazon}
        \label{fig:attrpert_mean_amazon}
    \end{subfigure}
    \hfill
    \begin{subfigure}[b]{0.48\textwidth}
        \centering
        \includegraphics[trim={0cm 0cm 0cm 0cm}, clip, width=\linewidth]{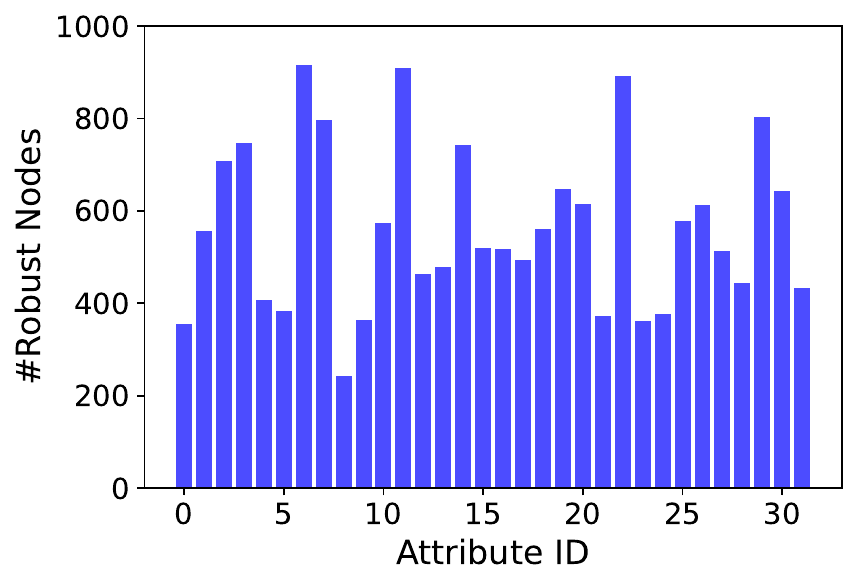}
        \caption{Yelp}
        \label{fig:attrpert_mean_yelp}
    \end{subfigure}
    \caption{The number of robust tasks solved by \textsc{GNNev} on real-world node-classification datasets for mean-aggregated GNNs. Only attribute perturbations are allowed on each single dimension under budgets as detailed in Appendix~\ref{appx:fullexp}.}
    \label{fig:attrperturb}
\end{figure}

\figurename~\ref{fig:attrperturb} shows the evaluation results of \textsc{GNNev} for attribute perturbations on two node-classification fraud datasets, Amazon and Yelp, given their rich attribute information.
Amazon has 25 attributes, including both continuous and discrete values. Yelp has 32 attributes, all of which are normalised to continuous values within $[0,1]$.
We performed an experiment, where we perturbed a single attribute for each run, for perturbation budget defined as the minimum to maximum values that have appeared in the data.
\textsc{GNNev} completed all runs within the time limit, demonstrating its efficiency. From the results on mean-aggregated GNNs, the models exhibited vulnerability on most attributes, especially on Yelp, where half of the nodes were successfully attacked by perturbing 13 out of 32 attributes.

\subsection{Ablation Studies}
We performed ablation experiments to evaluate the effectiveness of the proposed incremental solving algorithm.
A variant of \textsc{GNNev} was developed, where the incremental solving algorithm was replaced with a one-time solution of the complete encoding. We refer to this variant as \textsc{GNNev} w/o IS.
The experiments were conducted on {structural perturbations for} 3-layer GNNs, with perturbation budget $\Delta=2$ and fragile edge set $F=E$.

The results of the ablation experiments are shown in \tablename~\ref{tab:ablation}.
The number of winning tasks and the average runtime on Cora and CiteSeer are reported.
Compared to its variant without incremental solving, \textsc{GNNev} demonstrated lower average runtime and was more competitive on more tasks.
Specifically, on robust instances, the incremental solving algorithm lowers the runtime of up to 
82\% of tasks (on CiteSeer, mean aggregation).
These results indicate that the proposed incremental solving algorithm effectively enhances the practical performance of our exact verifier, 
offering the greatest advantage on robust instances, which is consistent with its design.

We have also shown the effectiveness of the proposed bound tightening strategies.
\tablename~\ref{tab:bound_stats} summarizes the mean and maximum gaps between the upper and lower bounds of the node embedding variables $y_{v,i}^{(k)}$ defined by Eq.~(\ref{eq:lin_trans}) across 4 datasets.
The experiments were conducted on 3-layer GNNs, with perturbation budget $\Delta=1$ and fragile edges set $F=E$.
Compared with the plain method as described in Section~\ref{sec:bound-tight}, our tightened bounds achieved lower mean and maximum gaps for all aggregations and layers $k$.
We remark that the sum aggregation function resulted in significantly larger gaps than mean and max, leading to performance drop of the solver.

\begin{table}[htbp]
    \centering
    \caption{Ablation experiments on the incremental solving algorithm of \textsc{GNNev}. The relatively better results are shown in \textbf{bold}.}
    \label{tab:ablation}
    \setlength{\tabcolsep}{6pt}
    \begin{tabular}{llrrrr}
        \toprule
        \multirow{2}{*}{Dataset} & \multirow{2}{*}{$\mathbf{aggr}$} & \multicolumn{2}{c}{\textsc{GNNev}} & \multicolumn{2}{c}{\textsc{GNNev} w/o IS} \\
        \cmidrule(lr){3-4} \cmidrule(lr){5-6}
        & & \multicolumn{1}{c}{\#Win} & \multicolumn{1}{c}{Time(s)} & \multicolumn{1}{c}{\#Win} & \multicolumn{1}{c}{Time(s)} \\
        \midrule
        \multicolumn{6}{l}{\textit{All instances}} \\
        \multirow{1}{*}{Cora} & sum & \textbf{1,743} & \textbf{7.84} & 965 & 9.21 \\
        & max & 1,313 & \textbf{19.81} & \textbf{1,383} & 20.57 \\
        & mean & \textbf{1,532} & \textbf{32.06} & 1,176 & 34.88 \\
        \multirow{1}{*}{CiteSeer} & sum & \textbf{2,491} & \textbf{10.09} & 821 & 13.35 \\
        & max & \textbf{1,989} & \textbf{28.47} & 1,222 & 29.82 \\
        & mean & \textbf{1,759} & \textbf{15.15} & 1,553 & 17.41 \\
        \midrule
        \multicolumn{6}{l}{\textit{Robust instances}} \\
        \multirow{1}{*}{Cora} & sum & \textbf{1,467} & \textbf{5.02} & 476 & 6.49 \\
        & max & \textbf{1,084} & \textbf{11.87} & 855 & 12.68 \\
        & mean & \textbf{931} & \textbf{20.33} & 269 & 23.03 \\
        \multirow{1}{*}{CiteSeer} & sum & \textbf{2,217} & \textbf{7.34} & 499 & 8.73 \\
        & max & \textbf{1,767} & \textbf{7.33} & 778 & 8.72 \\
        & mean & \textbf{1,257} & \textbf{7.75} & 273 & 10.09 \\
        \bottomrule
    \end{tabular}
\end{table}

\begin{table}[h]
    \centering
    \caption{Comparison of the gaps between the upper and lower bounds of node embedding variables for the tightened and plain bound propagation methods.}
    \label{tab:bound_stats}
    \setlength{\tabcolsep}{8pt}
    \begin{tabular}{llrrrr}
        \toprule
        \multirow{2}{*}{$\mathbf{aggr}$} & \multirow{2}{*}{$k$} & \multicolumn{2}{c}{Tightened Bounds} & \multicolumn{2}{c}{Plain Bounds} \\
        & & \multicolumn{1}{c}{Mean gap} & \multicolumn{1}{c}{Max gap} & \multicolumn{1}{c}{Mean gap} & \multicolumn{1}{c}{Max gap} \\
        \midrule
        \multicolumn{6}{l}{\textit{Cora}} \\
        \multirow{1}{*}{sum} & 1 & 0.58 & 2.66 &  0.67 & 3.42 \\
        & 2 & 7.46 & 38.26 &  9.29 & 50.58 \\
        & 3 & 91.94 & 472.87 &  104.66 & 566.26 \\
        \multirow{1}{*}{max} & 1 & 0.57 & 3.28 &  0.66 & 3.75 \\
        & 2 & 5.25 & 17.12 &  6.42 & 22.10 \\
        & 3 & 50.94 & 130.73 &  58.00 & 151.40 \\
        \multirow{1}{*}{mean} & 1 & 0.42 & 2.35 &  0.73 & 3.68 \\
        & 2 & 4.17 & 11.82 &  7.30 & 22.83 \\
        & 3 & 41.60 & 93.48 &  65.96 & 162.08 \\
        \midrule
        \multicolumn{6}{l}{\textit{CiteSeer}} \\
        \multirow{1}{*}{sum} & 1 & 0.46 & 6.13 &  0.55 & 14.36 \\
        & 2 & 4.99 & 97.94 &  6.40 & 189.29 \\
        & 3 & 51.57 & 1,002.33 &  59.86 & 1,354.72 \\
        \multirow{1}{*}{max} & 1 & 0.46 & 4.00 &  0.53 & 7.41 \\
        & 2 & 3.70 & 23.50 &  4.55 & 37.25 \\
        & 3 & 32.53 & 144.33 &  37.55 & 198.76 \\
        \multirow{1}{*}{mean} & 1 & 0.34 & 2.00 &  0.51 & 7.66 \\
        & 2 & 2.91 & 10.39 &  4.35 & 38.81 \\
        & 3 & 25.83 & 77.49 &  35.81 & 203.99 \\
        \midrule
        \multicolumn{6}{l}{\textit{Amazon}} \\
        \multirow{1}{*}{sum} & 1 & 46.29 & 993.37 &  144.50 & 21,387.24 \\
        & 2 & 391.64 & 14,691.80 &  562.75 & 25,208.20 \\
        & 3 & 2,008.59 & 37,071.13 &  2,678.37 & 60,287.40 \\
        \multirow{1}{*}{max} & 1 & 40.71 & 991.43 &  74.81 & 1,317.29 \\
        & 2 & 101.59 & 1,626.05 &  134.98 & 2,378.28 \\
        & 3 & 525.86 & 4,350.96 &  611.23 & 4,819.04 \\
        \multirow{1}{*}{mean} & 1 & 19.53 & 1,025.55 &  68.98 & 1,126.02 \\
        & 2 & 54.64 & 1,681.41 &  143.64 & 1,921.16 \\
        & 3 & 226.63 & 3,832.60 &  466.67 & 4,021.53 \\
        \midrule
        \multicolumn{6}{l}{\textit{Yelp}} \\
        \multirow{1}{*}{sum} & 1 & 1.80 & 6.76 &  11.16 & 182.42 \\
        & 2 & 28.76 & 387.72 &  391.38 & 17,145.60 \\
        & 3 & 711.55 & 34,765.66 &  7,880.43 & 993,569.36 \\
        \multirow{1}{*}{max} & 1 & 0.66 & 3.69 &  2.37 & 4.51 \\
        & 2 & 5.42 & 27.59 &  17.85 & 38.48 \\
        & 3 & 111.11 & 205.89 &  191.89 & 280.45 \\
        \multirow{1}{*}{mean} & 1 & 0.09 & 2.11 &  0.43 & 2.49 \\
        & 2 & 0.38 & 5.14 &  2.04 & 6.42 \\
        & 3 & 12.45 & 48.56 &  34.62 & 54.53 \\
        \bottomrule
    \end{tabular}
\end{table}

\end{document}